\documentclass[12pt,onecolumn,draftclsnofoot]{IEEEtran}
\usepackage{amsmath,amssymb,graphicx,cite}
\usepackage[noend,ruled]{algorithm2e}
\usepackage{color,graphics,epsfig}
\usepackage{tikz,pgfplots}
\usetikzlibrary{automata,fit,decorations.pathmorphing}
\usetikzlibrary{positioning,shapes.multipart}
\usepgflibrary{shapes.multipart}

\newcommand{\qB}{\ensuremath{\mathbf{q}}}

\newcommand{\wB}{\ensuremath{\mathbf{w}}}
\newcommand{\scrA}{\ensuremath{\mathcal{A}}}
\newcommand{\scrC}{\ensuremath{\mathcal{C}}}

\newcommand{\scrE}{\ensuremath{\mathcal{E}}}
\newcommand{\scrF}{\ensuremath{\mathcal{F}}}
\newcommand{\scrG}{\ensuremath{\mathcal{G}}}
\newcommand{\scrO}{\ensuremath{\mathcal{O}}}
\newcommand{\scrP}{\ensuremath{\mathcal{P}}}
\newcommand{\scrQ}{\ensuremath{\mathcal{Q}}}
\newcommand{\scrS}{\ensuremath{\mathcal{S}}}
\newcommand{\scrH}{\ensuremath{\mathcal{H}}}
\newcommand{\scrM}{\ensuremath{\mathcal{M}}}
\newcommand{\scrW}{\ensuremath{\mathcal{W}}}
\newcommand{\scrX}{\ensuremath{\mathcal{X}}}
\newcommand{\scrY}{\ensuremath{\mathcal{Y}}}
\newcommand{\rA}{\ensuremath{\rightarrow}}

\newcommand{\Bzero}{\ensuremath{\mathbf{0}}}
\newcommand{\Bone}{\ensuremath{\mathbf{1}}}

\newtheorem{theorem}{Theorem}
\newtheorem{definition}{Definition}

\newtheorem{lemma}{Lemma}
\newtheorem{property}{Property}

\begin{document}

\title{Discovering General partial orders in event streams}
\author{Avinash~Achar, Srivatsan~Laxman, V.~Raajay and  P.~S.~Sastry 
\thanks{Avinash~Achar is with Indian Institute of Science, Bangalore. Email: avinash@ee.iisc.ernet.in}
\thanks{S.~Laxman is with Microsoft Research Labs, Bangalore. Email: slaxman@microsoft.com}
\thanks{V.~Raajay is with Indian Institute of Science, Bangalore. Email:raajay@ee.iisc.ernet.in}
\thanks{P.~S.~Sastry is with Indian Institute of Science, Bangalore. Email: sastry@ee.iisc.ernet.in}}
\date{}
\maketitle

\begin{abstract}
Frequent episode discovery is a popular framework for pattern discovery in event streams.
An episode is a partially ordered set of nodes with each node associated with an event
type. Efficient (and separate) algorithms exist for episode discovery when the associated partial order is total 
(serial episode) and trivial (parallel episode).
In this paper, we propose efficient algorithms for discovering frequent episodes with general partial orders. These algorithms can be easily
specialized to discover serial or parallel episodes. Also, the algorithms are flexible enough to be specialized for mining in the space of
certain interesting subclasses of partial orders. 
We point out that there is an inherent combinatorial explosion in frequent partial order mining and 
 most importantly, frequency alone is not a sufficient measure of interestingness. We propose a new interestingness measure for general
 partial order episodes and a
 discovery 
 method based on this measure, for filtering out uninteresting partial orders.
Simulations demonstrate the effectiveness of our algorithms.
\end{abstract}

\section{Introduction}
\label{sec:intro}

Frequent episode discovery \cite{MTV97} is a popular framework for discovering temporal patterns in
symbolic time series data, with applications in several domains like manufacturing \cite{laxman06,USSL09}, telecommunication
\cite{MTV97}, WWW \cite{LTW08}, biology \cite{BCSZ06,PSU08}, finance
\cite{NF03}, intrusion detection \cite{LB00,WWT08}, text mining \cite{ITN04} etc.
The data in this framework is a single long time-ordered 
stream of events and each temporal pattern (called an episode) is essentially a small, partially ordered 
collection of nodes, with each node associated with a symbol (called event-type). The partial order
in the episode constrains the time-order in which events should appear in the data, in order for
the events to constitute an occurrence of the episode.
Patterns with a total order on their nodes are called {\em serial} episodes, while those with an empty
partial order are called {\em parallel} episodes \cite{MTV97}. The task  is to unearth all episodes whose frequency
in the data exceeds a user-defined threshold.

Currently, separate algorithms exist in the literature for discovering frequent serial and parallel
episodes in data streams \cite{MTV97,garriga03,laxman06,PSU08}, while no algorithms are available for the
case of episodes with general partial orders. Related work can be found in the
context of sequential patterns \cite{AS95,garriga05,PWLWWY06,MM00}
where the data consists of multiple sequences and the sequential pattern is a small partially ordered
collection of symbols. A sequential pattern is considered frequent 
if there are enough sequences (in the data) in  which the pattern   
occurs {\em atleast once}. By contrast, in frequent
episode discovery, we are looking for patterns that repeat often in 
a single long stream of events. This makes the computational task quite different from that in
sequential patterns. 

In this paper, we develop algorithms for discovering frequent episodes with general
partial order constraints over their nodes. We restrict our attention to a subclass of patterns
called {\em injective} episodes, where an event-type cannot appear more than once in a given episode.
This facilitates the design of efficient algorithms with no restriction
whatsoever on the partial orders of episodes. Further, our algorithms  
can handle the usual expiry time constraints for episode occurrences (which limit the time-spans of
valid occurrences to some user-defined maximum value).  Our algorithms can be easily specialized to
either discover only frequent serial episodes or only frequent parallel episodes.  Moreover, we can also specialize the method to focus the discovery process to certain classes  
of partial order episodes which satisfy what we call as the {\em maximal subepisode property}
(Serial episodes and parallel episodes are specific examples of classes that obey this property).

As we point out here, one of the difficulties in efficient discovery of general partial orders is that   
there is an inherent combinatorial explosion in the number of
frequent episodes of any given size. This is because, for any partial order episode with $n$ nodes, 
 there are an exponential
number of subepisodes, also of size $n$, all of which would occur at least as often as the episode.  
(Note that this problem does not arise in, e.g., frequent serial episode discovery  
because an $n$-node serial episode cannot have any $n$-node serial subepisode).  
Thus, frequency alone is insufficient as a measure of interestingness for  
episodes with general partial orders. To tackle this,   
 we propose a {\em new}
measure called {\em bidirectional evidence}, which captures some notion of entropy
of relative frequencies of pairs of events occurring in either order  
in the observed occurrences of an episode.
The mining procedure now requires a user-defined threshold on bidirectional evidence in
addition to the usual frequency threshold. We demonstrate the utility of our 
algorithms through extensive empirical studies.

The paper is organized as follows. In Sec.~\ref{sec:episode-formalism}, we briefly
review the frequent episodes formalism  and define injective episodes.
Sec.~\ref{sec:automata-properties} describes the finite state automata 
(and its associated properties) for tracking occurrences of injective episodes. Algorithms for counting frequencies 
of partial order episodes 
are described in Sec.~\ref{sec:counting-algos}. The candidate generation is described in Sec.~\ref{sec:cand-gen}. 
Sec.~\ref{sec:bidirectional-evidence} describes our new interestingness measure.
We present simulation results in Sec.~\ref{sec:simulation-results} and conclude in Sec.~\ref{sec:conclusions}.

\section{Episodes in event streams}
\label{sec:episode-formalism}

The data, referred to as an {\em event sequence}, is denoted by $\mathbb{D} = \langle
(E_1,t_1),$ $(E_2, t_2),$ $\ldots (E_n,t_n) \rangle$, where $n$ is
the number of events in the datastream. In each tuple $(E_i,t_i)$,
$E_i$ denotes the event type and $t_i$ the time of occurrence of the
event. The event types $E_i$, take values from a finite set, $\scrE$.
The sequence is ordered so that, $t_i \leq 
t_{i+1}$ for all $i = 1, 2, \ldots$. The following is an example 
sequence with 10 events:
\begin{eqnarray}
%\mathbb{D}_1&=&
%&&\langle (A,2),(B,3),(A,3),(A,7),(C,8),(B,9) \nonumber \\
%&& (D,11),(C,12),(A,13),(B,14),(C,15) \rangle
&&\langle (A,2),(B,3),(A,3),(A,7),(C,8),(B,9),(D,11),(C,12),(A,13),(B,14),(C,15) \rangle
\label{eq:example-sequence}
\end{eqnarray}

%A subsequence of $\mathbb{D}$ is defined as a subset of the events of $\mathbb{D}$ arranged in their time order. For example,
%$\langle (A,2),(A,7),(C,8)\rangle$ is a subsequence of sequence (\ref{eq:example-sequence}).
\begin{definition}
\label{def:episode}
\cite{MTV97} An $N$-node episode $\alpha$, is a tuple, $(V_\alpha,$ $<_\alpha,
g_\alpha)$, where $V_\alpha =\{v_1, v_2,\ldots, v_N\}$ denotes a collection of nodes,
$<_\alpha$ is a strict partial order\footnote{A strict partial order is a relation which is irreflexive, asymmetric and
transitive.} on $V_{\alpha}$ and $g_\alpha : V_\alpha \rA \scrE$ is a map that associates each node in the
episode with an event-type (out of the alphabet $\scrE$). 
\end{definition}
When $<_\alpha$ is a total order, $\alpha$ is referred to as a serial episode and when 
$<_\alpha$ is empty $\alpha$ is referred to as a parallel episode.
In general, episodes can be neither serial nor parallel. 
We denote episodes using a simple graphical notation. 
For example, consider a $3$-node episode
$\alpha = (V_\alpha,<_\alpha, g_\alpha)$, where $v_1<_\alpha v_2$ and $v_1<_\alpha v_3$, and with
$g_\alpha(v_1)=B$, $g_\alpha(v_2)=A$ and $g_\alpha(v_3)=C$. We denote this episode as 
 $(B \rA (A\,C))$, implying that $B$ is followed by $A$ and $C$ in any order.

%An occurence of episode $\alpha$ in an event sequence, $\mathbb{D}$, is a subsequence of
%$\mathbb{D}$ whose events appear in a time-order that respects $<_\alpha$. Stated formally it is as follows.
\begin{definition}
\cite{MTV97} Given a data stream, $\langle (E_1,t_1)$, $\ldots$, $(E_n,t_n)\rangle$ and an episode $\alpha=(V_{\alpha},<_\alpha
,g_\alpha)$, an occurrence of 
$\alpha$ is a map $h:V_\alpha \rA \{1,\ldots,n\}$ such that $g_\alpha(v) = E_{h(v)}$
for all $v \in V_\alpha$, and for all $v,w \in V_\alpha$ with $v <_\alpha w$ we have $t_{h(v)} < t_{h(w)}$. 
\end{definition}
For example, $\langle (B,3), (A,7), (C,8)\rangle$ and $\langle (B,9),(C,12),(A,13)\rangle$
constitute occurences of $(B \rA(A\,C))$ in the event sequence ($\ref{eq:example-sequence}$), while $\langle (B,3), (A,3),
(C,8)\rangle$ is not a valid occurence 
%since the events $(A,3)$ and $(B,3)$ violate the partial
%order of $(B\rA(A\,C))$
since $B$ does not occur {\em before} $A$.

Given any $N$-node episode, $\alpha$, it
is sometimes useful to represent an occurence, $h$, of $\alpha$  as a vector of integers
$[h(1),h(2)\ldots h(N)]$, where $h(i) < h(i+1), i=1,\ldots,(N-1)$. For example, in
sequence (\ref{eq:example-sequence}), the occurence corresponding to the subsequence $\langle(B,3),(A,7),(C,8)\rangle$
is associated with the vector $[ 2\ 4\ 5 ]$ (since $(B,3)$, $(A,7)$ and $(C,8)$ are the second,
fourth and fifth events in (\ref{eq:example-sequence}) respectively).

Consider an $N$-node episode, $\alpha$, and the set, $\scrH_\alpha$, of occurrences of $\alpha$ in 
event sequence $\mathbb{D}$. The occurrences in $\scrH_\alpha$ can be arranged according to 
the lexicographic ordering of the vectors, $[h(1),\ldots,h(N)]$, $h\in\scrH_\alpha$.
\begin{definition}
\cite{LSU07} The lexicographic order, $<_\star$, on the set, $\scrH_\alpha$ of occurrences of an
$N$-node episode, $\alpha$, in an event sequence, $\mathbb{D}$, can be defined as follows:
Given two different occurences $h_1$ and $h_2$ of $\alpha$ in $\mathbb{D}$, we have  $h_1<_\star
h_2$ iff the least $i$ for which $h_1(i)\neq h_2(i)$ is such that $h_1(i)< h_2(i)$.
\end{definition}

\begin{definition}
\cite{MTV97} Episode $\beta=(V_\beta,<_\beta, g_\beta)$
is said to be a {\em subepisode} of $\alpha=(V_\alpha,<_\alpha,g_\alpha)$ (denoted
$\beta \preceq \alpha$) if there exists a $1-1$ map $f_{\beta\alpha}\::\:V_\beta\rA V_\alpha$ such that
(i)~$g_\beta(v) = g_\alpha(f_{\beta\alpha}(v))$ for all $v \in V_\beta$, and (ii)~for all $v,w \in
V_\beta$ with $v<_\beta w$, we have $f_{\beta\alpha}(v)<_\alpha f_{\beta\alpha}(w)$ in $V_\alpha$.
\label{def:subepisode}
\end{definition}
In other words, for $\beta$ to be a subepisode of $\alpha$, all event-types of $\beta$ must also be
in $\alpha$, and the order among the event-types in $\beta$ must also hold in $\alpha$.
Thus, $(B\rA A)$, $(B\rA C)$ and $(AC)$ are the 2-node subepisodes of $(B\rA (AC))$. We note here that if $\beta
\preceq \alpha$, then every occurence of $\alpha$ contains an occurence of $\beta$.

Given an event sequence the datamining task here is to discover all frequent episodes, i.e., those episodes whose frequencies
exceed a given threshold. Frequency is some measure of how often an episode occurs in the data stream. The frequency of episodes can 
be defined in more than one way \cite{MTV97,LSU05}. In this paper, we consider
the non-overlapped occurrences-based frequency measure for episodes \cite{LSU05}.
Informally, two
occurrences of an episode are said to be non-overlapped if no event corresponding to one occurrence
appears in-between events of the other. The frequency of an episode is the size of the largest set
of non-overlapped occurrences for that episode in the given data stream.
\begin{definition}
\cite{LSU05} Consider a data stream (event sequence), $\mathbb{D}$, and an $N$-node episode, $\alpha$. Two occurences $h_1$ and $h_2$ of
$\alpha$ are said to
be non-overlapped in $\mathbb{D}$ if either $t_{h_1(N)} < t_{h_2(1)}$ or $t_{h_2(N)} < t_{h_1(1)}$. A set of
occurences is said to be non-overlapped if every pair of occurences in the set is non-overlapped.
The cardinality of the largest set of non-overlapped occurrences of $\alpha$ in $\mathbb{D}$ is
referred to as the {\bf non-overlapped frequency} of $\alpha$ in $\mathbb{D}$.

\label{def:no}
\label{def:nonoverlapped}
\end{definition}

\subsection{Injective Episodes}
\label{sec:injective-episodes}

In this paper, we consider a sub-class of episodes called {\em injective episodes}. An episode,
$\alpha=(V_\alpha,<_\alpha,g_\alpha)$ is said to be injective if the $g_\alpha$ is an injective
(or 1-1) map. For example, the episode $(B\rA (AC))$ is an injective episode, while $B\rA (AC)\rA B$
is not. 
%(since, in the latter example, the corresponding $g$ will map two nodes in the episode to the
%same event-type,  $B$). 
Thus, an injective episode, 
is simply a subset of event-types (out of the alphabet, $\scrE$) with a partial order defined
over it. This subset, which we will denote by $X^\alpha$, is same as the range of $g_\alpha$. The partial
order that is induced over $X^\alpha$ by $<_\alpha$ is denoted by $R^\alpha$. It is often
much simpler to view an injective episode, $\alpha$, in terms of the {\em partial order set},
$(X^\alpha,R^\alpha)$, that is associated with it. From now on, unless otherwise stated, when we say episode we mean an injective
episode.

In this paper, we will use either $(V_\alpha,<_\alpha,g_\alpha)$ or $(X^\alpha,R^\alpha)$ to denote
episode $\alpha$, depending on the context. Although $(X^\alpha,R^\alpha)$ is simpler,  in some contexts, e.g., when referring to episode 
occurrences,
the $(V_\alpha,<_\alpha,g_\alpha)$ notation comes in handy. However, there can be multiple $(V_\alpha,<_\alpha,g_\alpha)$ 
representations for the same underlying pattern under {\em Definition~\ref{def:episode}}. Consider, for example,
two $3$-node episodes, $\alpha_1=(V_{1},<_{\alpha_1},g_{\alpha_1})$ and $\alpha_2=(V_{2},<_{\alpha_2},g_{\alpha_2})$, 
defined as: (i)~$V_1=\{v_1,v_2,v_3\}$ with $v_1<_{\alpha_1} v_2$, $v_1<_{\alpha_1} v_3$ and
$g(v_1)=B$, $g(v_2)=A$, $g(v_3)=C$, and (ii)~$V_2=\{v_1,v_2,v_3\}$ with $v_2<_{\alpha_2} v_1$, $v_2<_{\alpha_2} v_3$ and $g(v_1)=A$, 
$g(v_2)=B$ and $g(v_3)=C$. Both $\alpha_1$ and $\alpha_2$ represent the same pattern, 
and they are 
%namely,
%$(B\rA(AC))$. Note that given an event sequence, any subsequence of it that constitutes an occurrence of 
%$\alpha_1$ will always constitute an occurrence of $\alpha_2$ and vice versa. Consequently,
%$\alpha_1$ and $\alpha_2$ are 
indistinguishable based on their occurrences, no matter what the given
data sequence is. (Notice that there is no such ambiguity in the $(X^\alpha,R^\alpha)$ representation). In
order to obtain a unique $(V_\alpha,<_\alpha,g_\alpha)$ representation for $\alpha$, 
we assume a lexicographic order over the alphabet, $\scrE$, 
and ensure that $(g_\alpha(v_1),\ldots, g_\alpha(v_N))$ is ordered as per this ordering. 
%and associate the nodes in an episode
%with event-types as per this lexicographic order. 
Note that this lexicographic order on
$\scrE$ is not related in anyway to the actual partial order, $\leq_\alpha$.
% that events in an occurrence of $\alpha$ must obey; 
The lexicographic ordering over $\scrE$ is only required to ensure
a unique representation of injective episodes in the $(V_\alpha,<_\alpha,g_\alpha)$ notation. 
%This way, $g_\alpha$ is always
%such that the sequence of event-types, $(g_\alpha(v_1),\ldots, g_\alpha(v_N))$, obeys the lexicographic ordering
%on $\scrE$. 
Referring to the earlier example involving $\alpha_1$ and $\alpha_2$, we will use $\alpha_2$ to denote the pattern $(B\rA(AC))$.  

%\begin{definition}
%\label{def:injective-episode}
%An $N$-node episode, $\alpha=(V_\alpha,<_\alpha,g_\alpha)$, is said to be an {\em injective episode}
%if $g_\alpha$ is 1-1. Further, $g_\alpha$ is such that the sequence of event-types,
%$(g_\alpha(v_1),\ldots, g_\alpha(v_N))$, obeys the lexicographic ordering on $\scrE$.
%An alternative representation for $\alpha$, referred to as the {\em partial order
%set} for $\alpha$, is given by the pair, $(X^\alpha,R^\alpha)$, where $X^\alpha$ denotes the range of $g_\alpha$ and $R^\alpha$ denotes
%the (strict) partial order relation induced on $X^\alpha$ by $<_\alpha$.
%
%\label{def:injective-episodes}
%\end{definition}

\begin{table*}
\caption{Some example episodes}
\label{table:example}
\begin{center}
\begin{tabular}{|c|c|c|}
\hline
Episode & Graphical Notation & $X^\alpha,R^\alpha$\\
\hline
$V=\{v_1,v_2,v_3\}$ & $(C\rA B\rA A)$ & $X^\alpha=\{A,B,C\}$ \\
 $g(v_1)=A,g(v_2)=B,g(v_3)=C$ &  & $R^\alpha=\{(C,B),(B,A),(C,A)\}$\\
 $<_\alpha=\{(v_2,v_1),(v_3,v_1)(v_3,v_2)\}$ & &\\
\hline
$V=\{v_1,v_2,v_3\}$ & $(A\, B\, C)$ & $X^\alpha=\{A,B,C\}$ \\
 $g(v_1)=A,g(v_2)=B,g(v_3)=C$ &  & $R^\alpha=\phi$\\
 $<_\alpha=\{\}$ & &\\
\hline
$V=\{v_1,v_2,v_3,v_4\}$ & $(A\,B)\rA (C\,D)$ & $X^\alpha=\{A,B,C,D\}$ \\
 $g(v_1)=A,g(v_2)=B,g(v_3)=C,g(v_4)=D$ &  & $R^\alpha=\{(A,C),(B,C),(A,D),(C,D)\}$\\
 $<_\alpha=\{(v_1,v_3),(v_2,v_3)(v_1,v_4),(v_2,v_4)\}$ & &\\
\hline
$V=\{v_1,v_2,v_3,v_4,v_5\}. \,g(v_1)=A,g(v_2)=B,$ & $(A\rA((B\rA(D\,E))\,C))$ & $X^\alpha=\{A,B,C,D,E\}$ \\
 $g(v_3)=C,g(v_4)=D,g(v_5)=E$ &  & $R^\alpha=\{(A,B),(A,C),(A,D),(A,E),$\\
 $<_\alpha=\{(v_1,v_2),(v_1,v_3),(v_1,v_4)(v_1,v_5),(v_2,v_4),(v_2,v_5)\}$ & &$(B,D),(B,E)\}$\\
\hline
\end{tabular}
\end{center}
\end{table*}
Finally, note that, if $\alpha$ and $\beta$ are injective episodes, and if $\beta \preceq \alpha$ (cf.~{\em
Definition~\ref{def:subepisode}}), then the associated partial order sets are related as follows:
$X_{\beta}\subseteq X_{\alpha}$ and $R_{\beta}\subseteq R_{\alpha}$. 
Some examples of injective
episodes, illustrating the different notations for episodes, is given in {\em Table
\ref{table:example}}.

\section{Finite State Automata for partial orders}
\label{sec:automata-properties}
\label{sec:fsa}

Finite State Automata (FSA) can be used to track occurrences of injective episodes under general partial
orders in a manner similar to the automata-based algorithms for parallel or serial episodes  \cite{LSU05,LSU07,MTV97}. In 
this section, we describe the basic construction of such automata. 

\begin{figure}
\center
\begin{tikzpicture}[>= stealth, node distance=1cm,auto]

\node[state,initial] (0)			{$\phi,AB$};
\node[state]         (1) [above right=of 0]     {$A,B$};
\node[state]         (2) [below right=of 0]     {$B,A$};
\node[state]         (3) [below right=of 1]     {$AB,C$};
\node[state,accepting]         (4) [right=of 3]     	{$ABC,\phi$};

\path[->,very thick] (0)	 edge			node		{$A$} (1)
       	 edge			node[swap]		{$B$} (2)
       	 edge [loop above]	node 		{$\scrE$\textbackslash$\{A,B\}$} ()			
         (1)	edge 			node 		{$B$} (3) 					
                edge [loop above]	node 		{$\scrE $\textbackslash$\{B\}$} ()
         (2)	edge 			node[swap] 		{$A$} (3)
                edge [loop below]	node 		{$\scrE $\textbackslash$\{A\}$} ()
         (3)	edge 			node 		{$C$} (4)
       	 edge [loop above] 	node 		{$\scrE $\textbackslash$\{C\}$} ()
	(4) edge [loop above] node 	{$\scrE$}	();
\end{tikzpicture}
\caption{Automaton for tracking occurrences of the episode $((A\,B)\rA C)$.}
\label{fig:automaton}
\end{figure}
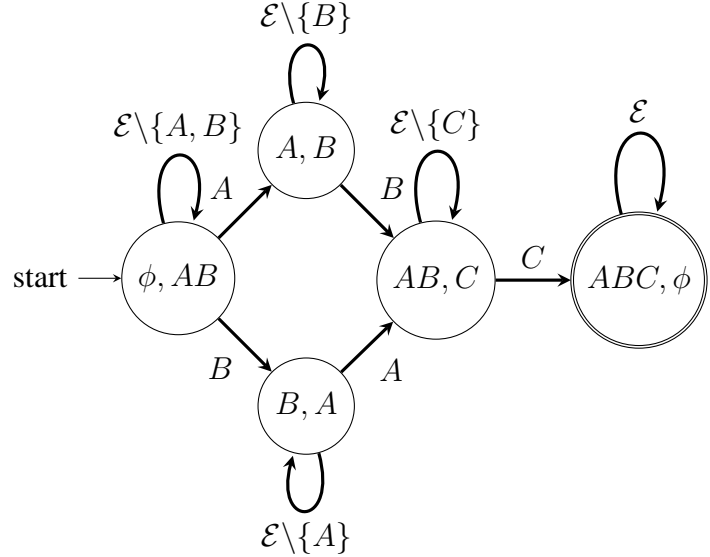

We first illustrate the automaton structure through an example. 
%Unlike serial episodes where we wait for only one event at a time, in case of (injective) partial
%order episodes (as in case of parallel episodes) we might in general need to wait for more than one event at any given
%time instant. For example, 
Consider  episode $(\alpha=(AB)\rightarrow C)$. Here, $X^{\alpha}=\{A,B,C\}$ and
$R^{\alpha}=\{(A,C),(B,C)\}$. The FSA used to track occurrences of this episode is shown in
Fig.~\ref{fig:automaton}. Each state, $i$, is associated with a pair of subsets of $X^\alpha$, namely,
$(\scrQ^\alpha_i,\scrW^\alpha_i)$;  $\scrQ^\alpha_i\subseteq X^\alpha$ denotes the event-types
that the automaton has already accepted by the time it arrives in {\em state i};
$\scrW^\alpha_i\subseteq X^\alpha$ denotes the event-types that the automaton in {\em state i} is
ready to accept. Initially, the automaton is in {\em state 0}, has not accepted any
events so far and is waiting for either of $A$ and $B$, i.e.,~$\scrQ^\alpha_0=\phi$ and $\scrW^\alpha_0=\{A,B\}$. 
%This is because, in
%$((AB)\rA C)$, events of type $A$ and $B$ can appear in any order, followed by a $C$. 
If we see a $B$ first, we accept it
and continue waiting for an $A$, i.e.,~the automaton transits to {\em state 2} with
$\scrQ^\alpha_2=\{B\}$, $\scrW^\alpha_2=\{A\}$. At this  point the automaton
is not yet ready to accept a $C$, which happens only after both $A$ and $B$ are encountered (in
whatever order). 
%Once the automaton reaches {\em state 2}, only after seeing an $A$ will it next
%wait for a $C$. 
If, instead of encountering a $B$, the automaton in {\em state 0} first encountered an $A$, then it
would transit into {\em state 1} (rather than {\em state 2}), where it would now wait for a $B$ to
appear (Thus, $\scrQ^\alpha_1=\{A\}$, $\scrW^\alpha_1=\{B\}$). 
Once both $A$ and $B$ appear in the data, the automaton will transit, either from {\em state
1} or {state 2}, and move into {\em state 3}, where it now waits for a $C$ ($\scrQ^\alpha_3=\{A,B\}$,
$\scrW^\alpha_3=\{C\}$). Finally, if the automaton now encounters a $C$ in the data stream, it will transit to the final state, namely, {\em
state 4} ($\scrQ^\alpha_4=\{A,B,C\}$, $\scrW^\alpha_4=\phi$) and recognize a full occurrence of the episode, $((AB)\rA C)$.

In any occurence of an  episode $\alpha$,  an event $E\in X^{\alpha}$ 
can occur only after all its parents in $R^{\alpha}$ have been seen. 
%For an injective episode, as a general strategy for tracking  its occurences, 
Hence, we initially  wait
for all those elements of $X^{\alpha}$ which are minimal elements of $R^{\alpha}$. Further, we start waiting for a 
non-minimal element, $E$,  of $R^{\alpha}$ immediately after all elements less than $E$
in $R^{\alpha}$ are seen. For each $E\in X^\alpha$, we refer to the subset of elements 
in $X^\alpha$ that are
less than $E$ (with respect to $R^\alpha$) as the {\em parents} of $E$ in episode, $\alpha$, and 
denote it  by $\pi_\alpha(E)$. We now define  $\scrA_{\alpha}$, the FSA to recognise occurences  of
$\alpha$.

%This strategy of computation can be 
%formalized using a finite state
%automaton described below for a general injective episode $\alpha$.

%Each state in $\scrA_{\alpha}$ is represented as $(\scrQ^{\alpha},\scrW^{\alpha})$, 
%a pair of sets, which are subsets of $X^{\alpha}$.
% $\scrQ^{\alpha}$ would denote the set of events already
%seen and  $\scrW^{\alpha}$ denotes the set of events the state is currently waiting for. As is evident from the %automaton of episode
%$(A B)\rA C$, not all subset pairs of $X^{\alpha}$ come up as states in $\scrA_\alpha$. Consider a state transition on %seeing an event
%$E$. This happens only after all parents of $E$ have already been accepted. So its intutively quite clear that for any %event $E$ in
%$\scrQ_\alpha$ of a state, its set of parents are also in $\scrQ_\alpha$. 
%We will characterize the state space of $\scrA_\alpha$ later in the section. 

%For each $E\in X^\alpha$, we refer to the subset of elements 
%in $X^\alpha$ that are
%less than $E$ (with respect to $R^\alpha$) as the {\em parents} of $E$ in episode, $\alpha$, and 
%denote it  by $\pi_\alpha(E)$. 

%We provide a construction of the Finite State Automaton, $\scrA_\alpha$,
%used to track the occurrences of an injective episode, $\alpha$ below.
%in {\em Definition~\ref{def:fsa}}  define $\scrA_{\alpha}$ in a 
%constructive 
%fashion as 
%follows.

\begin{definition}
\label{definition:automaton}
\label{def:fsa}

\em FSA $\scrA_\alpha$, used to track occurrences of
$\alpha$ in the data stream is defined as follows. Each state, $i$, in $\scrA_\alpha$, is
represented by a unique pair of subsets of $X^\alpha$, namely $(\scrQ^\alpha_i,\scrW^\alpha_i)$;
$\scrQ^\alpha_i\subseteq X^\alpha$ is the set of event-types that the automaton has accepted so far
and $\scrW^\alpha_i$ is the set of event-types that the automaton is
currently ready to accept. The initial state, namely, {\em state 0}, is
associated with the subsets pair, $(\scrQ^\alpha_0,\scrW^\alpha_0)$,
where $\scrQ^\alpha_0=\phi$ and $\scrW^\alpha_0$ is the collection of {\em least} elements in 
$X^\alpha$ with respect to $R^\alpha$. Let $i$ be the current state of $\scrA_\alpha$ and let the
next event in the data be of type, $E\in\scrE$. $\scrA_\alpha$ remains in {\em state $i$}  if $E\in
(X^\alpha\setminus\scrW^\alpha_i)$. If $E\in\scrW^\alpha_i$, then $\scrA_\alpha$ accepts $E$ and transits into state
$j$, with: 
\begin{eqnarray}
\scrQ^\alpha_j &=& \scrQ^\alpha_i \cup \{E\}\label{eq:Qalphaj}\\
\scrW^\alpha_j &=& \{E'\in (X^\alpha\setminus\scrQ^\alpha_j) \,:\,
\pi_\alpha(E')\subseteq\scrQ^\alpha_j\} \label{eq:Walphaj}
\end{eqnarray}
When $\scrQ^\alpha_j=X^\alpha$, (and hence $\scrW^\alpha_j=\phi$),  $j$ is the {\em final state} of $\scrA_\alpha$.
\end{definition}

It may be noted that not all possible tuples of $(\scrQ,\scrW)$, where $\scrQ\subseteq X_\alpha,
\scrW\subseteq X_\alpha$, constitute valid states of the automaton. For example in
Fig.~\ref{fig:automaton}, there can be no valid state corresponding to $\scrQ=\{A,C\}$ (since $C$
could not have been accepted without $B$ being accepted before it). %(See \cite{ALRS09} for more discussion).
We list below a few properties of the valid states of the automaton. All these are easily proved 
from the above definition. %\cite{techreport}. 

\begin{property}
For any state, $j$, of the automaton, $\scrA_\alpha$, the set, $\scrW^\alpha_j$, of event-types that
$\scrA_\alpha$ will wait for in state $j$ (as per Eq.~(\ref{eq:Walphaj}) in {\em Definition~\ref{def:fsa}}),
is exactly the set of {\em least} elements of $(X^\alpha\setminus\scrQ^\alpha_j)$ (with respect to the
partial order $R^\alpha$).
\label{property:Walternate}
\end{property}

\begin{proof}
If $E$ is a least element of $(X^\alpha\setminus\scrQ^\alpha_j)$, it implies that all parents of $E$
(if any) are outside $(X^\alpha\setminus\scrQ^\alpha_j)$. Hence, they must have already been accepted by $\scrA_\alpha$
(i.e.~we must have $\pi_\alpha(E)\subseteq\scrQ^\alpha_j$), and so, by (\ref{eq:Walphaj}), we have
$E\in\scrW^\alpha_j$. Conversely, every $E'$ that $\scrA_\alpha$ is waiting for
(according to (\ref{eq:Walphaj})) trivially belongs to $(X^\alpha\setminus\scrQ^\alpha_j)$ and since
(\ref{eq:Walphaj}) also prescribes that $\pi_\alpha(E')\subseteq\scrQ^\alpha_j$, we must have
$\pi_\alpha(E')\cap\scrW^\alpha_j=\phi$. Hence, such an $E'$ must be a least element of
$(X^\alpha\setminus\scrQ^\alpha_j)$.
\end{proof}

\begin{property}
\label{property:nonemptyW}
For any state, $j$, of automaton, $\scrA_\alpha$, if $(X^\alpha\setminus\scrQ^\alpha_j)$ is non-empty,
then $\scrW^\alpha_j$ is non-empty. 
Thus, the only state out of which $\scrA_\alpha$  makes
no state transitions no matter what the input sequence (i.e.~the only final state of $\scrA_\alpha$) 
is the one represented by the pair, $(X^\alpha,\phi)$.
\end{property}

\begin{proof}
If $(X^\alpha\setminus\scrQ^\alpha_j)$ is non-empty, then it must contain at least one {\em least}
element (with respect to $R^\alpha$) and from {\em Property~\ref{property:Walternate}}, this element
must be in $\scrW^\alpha_j$ (and hence, it must be non-empty).
\end{proof}

\begin{property}
Given the set, $\scrW^\alpha_j$, of event-types, that $\scrA_\alpha$ will wait
for in state, $j$, $j\neq 0$, the corresponding set %, $\scrQ^\alpha_j$, 
of event-types accepted by the time $\scrA_\alpha$ reaches state $j$, is given by
\begin{equation}
%\widetilde{\scrQ}^\alpha_j = 
\{E\in (X^\alpha\setminus\scrW^\alpha_j) \mathrm{\ s.t.\ }
\pi_\alpha(E)\cap\scrW^\alpha_j=\phi\}
\end{equation}
Thus, for any two distinct states, $i$ and $j$, of $\scrA_\alpha$, we
must have  both $\scrQ^\alpha_i\neq\scrQ^\alpha_j$ and $\scrW^\alpha_i\neq\scrW^\alpha_j$.
\label{property:WfixesQ}
\end{property}

\begin{proof}
If the automaton, $\scrA_\alpha$ has accepted an event of type $E$ (i.e.~if $E\in\scrQ^\alpha_j$ as
per {\em Definition~\ref{def:fsa}}) then all parents of $E$ (if any) should have been previously accepted by
$\scrA_\alpha$, and hence, we must have $E\in(X^\alpha\setminus\scrW^\alpha_j)$ and $\pi_\alpha(E)\cap\scrW^\alpha_j=\phi$.
To show the
other way, consider an $E\in(X^\alpha\setminus\scrW^\alpha_j)$ such that
$\pi_\alpha(E)\cap\scrW^\alpha_j=\phi$. Now, if $E\notin\scrQ^\alpha_j$ (i.e.~if $E$ has not yet
been accepted by $\scrA_\alpha$ as per {\em Definition~\ref{def:fsa}}), then
$\scrA_\alpha$ must wait for either $E$ or one (or more) of its parents, i.e.~either
$E\in\scrW^\alpha_j$ or $\pi_\alpha(E)\cap\scrW^\alpha_j\neq\phi$ (which contradicts our original
assumption for $E$). This completes the proof of {\em
Property~\ref{property:WfixesQ}}.

\end{proof}

The next two properties give an exact characterization of $\scrQ^\alpha$ and $\scrW^\alpha$ for an episode $\alpha$. They describe the
the kind of subsets(of $X^\alpha$) that actually come up as $\scrQ^\alpha$s and $\scrW^\alpha$s in $\scrA_\alpha$. 
\begin{property}
Let $\scrA_\alpha$ denote an automaton of episode, $\alpha$, as per construction. Given $\scrQ^\alpha \subseteq X^\alpha$, 
$\scrQ^\alpha$ is the set of event-types that $\scrA_\alpha$ has currently
accepted $\Longleftrightarrow$ $\forall E\in\scrQ^\alpha$, $\pi_\alpha(E)\subseteq\scrQ^\alpha$.
\label{property:Qclosed}
\end{property}

\begin{proof}
Since $\scrQ^\alpha_0$ was initially empty, for $E\in\scrQ^\alpha$, we must have had
$E\in\scrW^\alpha_i$ for some other (earlier) state, $i$. Now if $E\in\scrW^\alpha_i$, then either
(i)~if $\pi_\alpha(E)=\phi$, then $E$ must be a least element of $X^\alpha$ with respect to $R^\alpha$, or
(ii)~$\pi_\alpha(E)$ is non-empty, so that, by applying (\ref{eq:Walphaj}) for state $i$, we know $E$ must have been added
to $\scrW^\alpha_i$ only if $\pi_\alpha(E)\subseteq\scrQ^\alpha_i$.
But, from (\ref{eq:Qalphaj}), we know $\scrQ^\alpha_i\subseteq\scrQ^\alpha$. This implies
$\pi_\alpha(E)\subseteq\scrQ^\alpha$. 

Conversely, suppose $\scrQ^\alpha$ is such that $\forall E \in \scrQ^\alpha$, $\pi_\alpha(E)\subseteq\scrQ^\alpha$. Consider the
least element $E_1$ in $\scrQ^\alpha$(with respect to $R^\alpha$). $E_1$ has no parents in $\scrQ^\alpha$. By definition of 
$\scrQ^\alpha$,
$E_1$ has no parents outside $\scrQ^\alpha$. Hence, $E_1$ is also a least element of $X^\alpha$ which implies $E_1 \in \scrW^\alpha_0$. 
Hence
from state $0$, $\scrA_\alpha$ makes a transition(on seeing $E_1$) to a state $1$ with $\scrQ_1^\alpha=E_1$. Now consider a least 
element $E_2$ in
$\scrQ^\alpha \backslash \scrQ^\alpha_1$. One can verify on similar lines that $\pi_\alpha(E_2)\subseteq \scrQ^\alpha_1$. This
means(from (\ref{eq:Walphaj})) that $E_2\in \scrW_1^\alpha$. Hence, $\scrA_\alpha$ makes a transition(on seeing $E_2$) to a state $2$
with $\scrQ^\alpha_2= \scrQ^\alpha_1\cup\{E_2\}$. This process continues till a stage $k=|\scrQ^\alpha|$ at which $\scrA_\alpha$
actually enters a state $k$ where $\scrQ_k^\alpha=\scrQ^\alpha$.

\end{proof}

\begin{property}
Let $\scrA_\alpha$ denote an automaton of episode, $\alpha$, as per construction. Given $\scrW^\alpha \subseteq X^\alpha$, 
$\scrW^\alpha$ is the set of event-types that $\scrA_\alpha$ is currently waiting
for $\Longleftrightarrow$ $\forall E\in\scrW^\alpha$, $\pi_\alpha(E)\cap\scrW^\alpha=\phi$.
\label{property:Wclosed}
\end{property}

\begin{proof}
The forward direction is straightforward from (\ref{eq:Walphaj}). Conversely,
consider a $\scrW^\alpha$ such that $\forall E\in\scrW^\alpha$, $\pi_\alpha(E)\cap\scrW^\alpha=\phi$. Consider the following set.
\begin{equation}
\widetilde{\scrQ}^\alpha = \{E\in (X^\alpha\setminus\scrW^\alpha)\,:\, 
\pi_\alpha(E)\cap\scrW^\alpha=\phi\}
\end{equation}

Note that this set is exactly similar to the one defined in {\em Property~\ref{property:WfixesQ}}. We will first show that this
$\widetilde{\scrQ}^\alpha$ is such that $\forall E \in \widetilde{\scrQ}^\alpha, \pi_\alpha(E)\subseteq \widetilde{\scrQ}^\alpha$. If
this is true, then from {\em Property\ref{property:Qclosed}}, $\widetilde{\scrQ}^\alpha$ is a set of events that $\scrA_\alpha$ would  
have accepted at some stage. We next show that the $\scrW^\alpha$ that we started off with, is the set of event-types
$\scrA^\alpha$ would wait for, after having accepted the set of events $\widetilde{\scrQ}^\alpha$.

Consider an $E\in \widetilde{\scrQ}^\alpha$. Let $E'$ be a parent of $E$. We need to show that $E'\in \widetilde{\scrQ}^\alpha$ i.e.~
$E' \notin (X^\alpha\setminus \scrW^\alpha)$ and $\pi_\alpha(E')\cap\scrW^\alpha=\phi$. If
$E' \notin (X^\alpha\setminus \scrW^\alpha)$, then a parent of $E$ is in $\scrW^\alpha$ which contradicts $E \in
\widetilde{\scrQ}^\alpha$. If $\pi_\alpha(E)\cap\scrW^\alpha\neq\phi$, then $\exists$ an $E''\in \scrW^\alpha$ such that $(E'',E')\in
R^\alpha$. But we also have $(E',E) \in R^\alpha$. Hence, by trasitivity, $E''$ is a parent of $E$ in $\scrW^\alpha$, which
contradicts $E \in \widetilde{\scrQ}^\alpha$. 

We now show that $\scrW^\alpha$ is indeed the set of least elements in $\scrX^\alpha \setminus \widetilde{\scrQ}^\alpha$. Every
element in $Y=X^\alpha\setminus(\scrW^\alpha\cup \widetilde{\scrQ}^\alpha)$ should have a parent in $\scrW^\alpha$. Otherwise, $E$
must be in $\widetilde{\scrQ}^\alpha$. So no element in $Y$ is a least element of $\scrW^\alpha \cup Y=\scrX^\alpha \setminus
\widetilde{\scrQ}^\alpha$. Consider an $E\in \scrW^\alpha$. We need to show that $\forall E'\in \scrW^\alpha \cup Y$, $E'$ is not
lesser than $E$. Since no two elements of $\scrW^\alpha$ are related, no $E'$ from $\scrW^\alpha$ can be less than $E$. Suppose
there exists an $E'\in Y$ such that $(E',E)\in R^\alpha$. Since $E' \in Y$, it has a parent in $\scrW^\alpha$. By transitivity, $E\in
\scrW^\alpha$ has a parent in $\scrW^\alpha$, which contradicts the definition of $\scrW^\alpha$. Hence we have shown that
$\scrW^\alpha$ would be the set of events $\scrA^\alpha$ would wait for, after having accepted the set of events 
$\widetilde{\scrQ}^\alpha$.
\end{proof}

\begin{property}
\label{property:subset-reachability}
Consider two states, $i$ and $j$, of $\scrA_\alpha$ with sets of accepted states, $\scrQ^\alpha_i$
and $\scrQ^\alpha_j$, such that $\scrQ^\alpha_i \subsetneq \scrQ^\alpha_j$. Let $k=|\scrQ^\alpha_j \setminus \scrQ^\alpha_i|$. There
exists a sequence of events, $(E_1,\ldots,E_k)$, on which $\scrA_\alpha$, currently in
state $i$, will make $k$ state transitions, eventually arriving at a state $j$ with the set of
accepted events given by $\scrQ^\alpha_j$.
\end{property}

\begin{proof}
The proof is very similar to the converse argument in {\em Property \ref{property:Qclosed}}. For the sake of completeness, we give
the entire argument. Let $E_1 \in (\scrQ^\alpha_j\setminus\scrQ^\alpha_i)$ such that $E_1$ is a least element of
$(\scrQ^\alpha_j\setminus\scrQ^\alpha_i)$ (with respect to $R^\alpha$). From {\em
Property~\ref{property:Qclosed}} we know $\pi_\alpha(E_1)$ must belong to $\scrQ^\alpha_j
(\supsetneq \scrQ^\alpha_i)$. Since
$E_1$ is (by definition) the least element of $(\scrQ^\alpha_j\setminus\scrQ^\alpha_i)$, none of
$E_1$'s parents are in $(\scrQ^\alpha_j\setminus\scrQ^\alpha_i)$. So we must have
$\pi_\alpha(E_1) \subseteq \scrQ^\alpha_i$. This will ensure (from (\ref{eq:Walphaj})) that
$E_1\in\scrW^\alpha_i$, and so, $\scrA_\alpha$ in state $i$, on seeing $E_1$, will make a transition to
(say) state $i_1$, with $\scrQ^\alpha_{i_1} = \scrQ^\alpha_i \cup \{E_1\}$ and $\scrW^\alpha_{i_1} = 
\{E'\in (X^\alpha\setminus\scrQ^\alpha_{i_1}) \mathrm{\ s.t.\ }
\pi_\alpha(E')\subseteq\scrQ^\alpha_{i_1}\}$. Next we consider $E_2$, a least element in $(\scrQ^\alpha_j \setminus \scrQ^\alpha_{i_1})$,
and repeating the same argument as for $E_1$ above, we can see that $\scrA_\alpha$ will now transit 
into state $i_2$, with $\scrQ^\alpha_{i_2} = \scrQ^\alpha_{i_1} \cup \{E_2\}$ and $\scrW^\alpha_{i_2} = 
\{E'\in (X^\alpha\setminus\scrQ^\alpha_{i_2}) \mathrm{\ s.t.\ }
\pi_\alpha(E')\subseteq\scrQ^\alpha_{i_2}\}$. Thus, for $l=2,\ldots,k$, we can construct
$\scrQ^\alpha_{i_l}$ by adding the least element of $(\scrQ^\alpha_j\setminus\scrQ^\alpha_{l-1})$,
$E_l$, to $\scrQ^\alpha_{l-1}$.
At each step, the number of accepted elements increments by 1, so that after accepting $k$
events in this manner, $\scrA_\alpha$ will arrive in state $i_k$ with $\scrQ^\alpha_{i_k}=\scrQ^\alpha_j$
(since, $|\scrQ^\alpha_{i_k}|=|\scrQ^\alpha_j|$ and $\scrQ^\alpha_{i_k} \subseteq \scrQ^\alpha_j$).

\end{proof}

 \section{Counting Algorithms} 
\label{sec:counting-algos}
The data mining task in the frequent episode paradigm is to extract all episodes whose frequency exceeds a user-defined threshold. 
%The automata-based episode mining algorithms for serial 
%and parallel episodes use an apriori-based levelwise procedure for unearthing
%frequent patterns. The algorithms essentially perform
%a breadth-first search of the pattern space which is exponential in size.  
Like current algorithms for frequent serial/parallel episode discovery \cite{MTV97,LSU05}, 
we use an Apriori-style level-wise
procedure for mining frequent episodes with general partial orders. Each level has two steps, namely, {\em candidate generation} and {\em frequency counting}.
At level, $l$, candidate generation step combines frequent episodes of size $(l-1)$ to construct
candidates of size $l$. It exploits the simple but powerful fact that if a pattern is 
frequent then (certain kinds of) its subpatterns are also frequent.  
The frequency counting step computes frequencies of all episodes in the 
 candidates set and returns the set of
frequent $l$-size episodes. Sec.~\ref{sec:cand-gen} provides a detailed explanation of the candidate
generation. In this section, we present an algorithm for obtaining the frequencies (or counting the number of non-overlapped
occurences) of a set of general
injective episodes of a given size.

For counting the number of non-overlapped occurences of a set of serial episodes, \cite{LSU07}
proposed an algorithm using only one automaton per episode.  
This algorithm can be generalized as explained below, to count
non-overlapped occurences of a set of injective episodes (with general partial orders) 
by using the more general FSA of {\em Definition~\ref{def:fsa}}. 
We initialize the automaton (associated with the episode) in its start state.
The automaton would make transitions as prescribed by {\em Definition~\ref{def:fsa}}. 
We traverse the data stream and let the automaton transit 
to its next state as soon as a relevant event-type appears in the data stream. When the automaton 
reaches its final state, we increment the frequency of the episode and reset the automaton to its 
start state so that it would track the next occurence. 
 Since we have a set of candidates, we would have one such automaton for each 
episode. For each event in the data stream, we look at all automata waiting for that event and 
effect appropriate state transitions for all automata. Such an algorithm would count the 
non-overlapped occurrences of all candidate episodes through one pass over the data.  
%An automaton 
%is initialized for every candidate being counted and 
%is allowed to make transitions as prescribed by {\em Definition~\ref{def:fsa}}.
%Once it reaches its final state, frequency is incremented and a new automaton is initialized,
%which would track the next non-overlapped occurrence.
Consider this counting scheme for episode $\beta=(AB)\rA(CD)$ in the following data stream:
\begin{eqnarray}
&& \langle (A,1),(B,2),(A,3)(D,4),(E,5),(C,6),(D,7),\nonumber \\
&& (A,8),(B,9),(B,10),(C,12),(D,14) \rangle
\label{eq:algo-stream-ex}
\end{eqnarray}
The above  method tracks occurences $h_1=\langle(A,1), (B,2), (D,4),  (C,6)\rangle$ and 
$h_2=\langle(A,8), (B,9), (C,12),  (D,14)\rangle$ and returns frequency of 2 for this episode 
in this data stream. 

Though this algorithm is efficient (as it uses only one automaton per episode),
it cannot implement any temporal constraints on occurrences of episodes. One constraint 
that is often useful in applications is the {\em expiry time constraint} which is stated 
in terms of an upperbound on the {\em span} of an occurrence. 
The span of an occurence  is the largest difference between the times associated with
any two events in the occurrence. 
Under the expiry time constraint, the frequency of an episode is the maximum
number of non-overlapped occurrences such that span of each occurrence is less than a user-defined
threshold. (The window-width of \cite{MTV97} essentially implements a similar constraint). 
% of counting frequencies using {\em
%Algorithm~\ref{algo-no}} is that it does not directly enforce time constraints on the events
%constituting each of the occurrences. For example, in many scenarios it would be useful to ignore
%occurrences that are spread far apart. For example, we can define the notion of time-span of an
%occurrence, as the time difference between the first and last accepted events in the occurrence.
%To disregard occurrences that are spread far apart, the user can prescribe an {\em expiry time}
%constraint on the occurrences, and choose to ignore any occurrence whose time-span exceeds the
%expiry-time threshold. This way, the frequency measure for an episode, is the maximum number of
%non-overlapped occurrences that satisfy the given expiry time constraint. In this section, we
%present an algorithm for counting non-overlapped frequencies with an expiry time constraint, 
%for a set of (candidate) injective episodes under general partial orders.
An expiry time constraint is often useful because an occurence of a pattern constituted by events widely separated in time, may not really
indicate any underlying causative influences. Consider counting occurrences of $\beta$ in sequence (\ref{eq:algo-stream-ex}) with an expiry time
constraint of 4. The occurrences $h_1$ and $h_2$ of $\beta$ (that the algorithm would track as  
 specified earlier) have spans 5 and 6 respectively.  Hence our algorithm can only assign frequency 
of zero under the expiry time constraint. However, the occurence $h_3 = \langle (B,2),$ $(A,3), (D,4) (C,6)\rangle$ of $\beta$ in sequence  (\ref{eq:algo-stream-ex})
has a span of 4 (satisfying the constraint). The reason why our algorithm can not track $h_3$ is 
 that the automaton makes a state transition as soon as the relevant event-type appears in the 
 data and thus it accepts the event $(A,1)$. We can count non-overlapped occurrences under an 
expiry time constraint if we allow more than one automaton per episode as explained below. 

Consider this example  again with a modified algorithm as follows. As earlier, 
our automaton will accept $(A,1)$ and transit into a state that waits for a $B$. Now, 
since the automaton moved out of the start state we immediately spawn another automaton for this 
 episode which is initialized in the start state. Now when we encounter $(B,2)$ the first automaton 
 will accept it and move into a state where it waits for a $C$ or a $D$. The second automaton, which 
is in the start state, will also accept $(B,2)$ and move to a state where it waits for a $A$. (we would 
now intialize a third automaton for this episode because the second one moved out of the start 
 state). Now 
when we encounter $(A,3)$ the second automaton would accept it and move into a state of waiting for a 
$C$ or a $D$ which is the same state as the first one is in. From now on, both automata would 
 make identical transitions and hence we can retire the first automaton. This is because, the second 
automaton is initialized later and hence the span of the occurrence tracked by it would be smaller. 
(The third automaton would also accept $(A,3)$ and we will spawn a new fourth automaton in start 
 state). Now when we encounter $(D,4)$ and later $(C,6)$, the second automaton would reach the final 
state. Since the occurence tracked by this automaton satisfies the expiry time constraint, 
 we can now increment the frequency and then retire all other automata of this episode. We will 
also spawn a 
fresh automaton for this episode in the start state so that we can begin to track the next non-overlapped 
 occurence (if any) of this episode.  

We can now specify the general method for counting under expiry time constraint as follows. 
 Instead of spawning a new automaton only after
the existing one reaches its final state, we spawn a new automaton whenever 
 an existing automaton accepts its first event 
(i.e., when it transits out of its start state). Each of the automata makes a state transition as 
soon as a relevant event-type appears in the data stream. 
When counting like this, it is possible for two  automata to reach the same state. In
such cases, we drop the older one (retaining only the most recent automaton). This strategy
tracks, in a sense, the innermost occurence amongst a set of overlapping occurences that end together.
When any automaton (of an episode) reaches the final 
state, we check whether the span of the occurence tracked by this automaton satisfies 
 the expiry time constraint. If it does, we increment the frequency and 
 retire all the automata of that episode except for one automaton 
in the start state to track the next occurence. If the span of the occurrence tracked by the 
automaton that reached the final state does not satisfy the expiry time constraint, then we only 
retire that automaton. This is the algorithm that we use for counting the frequencies of episodes.  

%For example $h$ is the innermost occurence
%that ends at $(C,6)$. For example, on in (\ref{eq:algo-stream-ex}), two automata reach the same
%state (labelled by the pair of sets $(\{A,B\},\{C,D\}$) on seeing event $(A,3)$.
%The more recent automaton among them which tracked $(B,2)$ and $(A,3)$ is retained and this
%ultimately tracks the innermost occurence $h$. If the span of such an innermost occurence does not
%exceed the expiry-time constraint, $T_X$, frequency is incremented and all automata for $\beta$ are
%deleted except one in start state. This way, subsequent occurences tracked will be non-overlapped
%with the one just counted. On the other hand, if $h$ failed the expiry-time constraint,
%we only remove the current automaton  and continue looking for an innermost occurence satisfying $T_X$.

The pseudocode for counting non-overlapped occurrences of general injective episodes with an expiry-time
constraint is given in {\em Algorithm~\ref{algo:no-exp}}. 
%It essentially generalizes the serial episode non-overlapped counting algorithm with expiry
%constraints\cite{laxman06}. 
The inputs to the algorithm are: $\scrC_l$, the set of $l$-node
candidate episodes, $\mathbb{D}$, the event stream, $\scrE$, the set of event-types, 
  $\gamma$, the frequency
threshold,  and $T_X$, the expiry-time. The algorithm outputs the set, $\scrF_l$, of frequent episodes.  
 The event-types associated with an $l$-node episode,
$\alpha$, are stored in the $\alpha.g[]$ array -- for $i=1,\ldots,l$, $\alpha.g[i]$ is assigned the
value $g_\alpha(v_i)$.
We store the partial
order $<_\alpha$, associated with the episode as a  binary adjacency
matrix, $\alpha.e[][]$. The notation is: $\alpha.e[i][j]=1$ iff $v_i <_\alpha v_j$ 
(or equivalently, if $(\alpha.g[i],\alpha.g[j])\in R^\alpha$). 

%---------------------------------------------
%*********************************************
%---------------------------------------------

The main data structure is an array of lists, $waits()$, indexed by the set of event-types. The elements of
each list store the relevant information about all the automata that are waiting for a particular event-type
(and hence can make a state transition if that event-type appears in the data stream). The entries in the
list are of the form $(\alpha,\qB,\wB,j)$ where $\alpha$ is a candidate episode, $(\qB,\wB)$ is one of the
possible states of the automaton associated with $\alpha$ (cf.{\em ~Definition \ref{def:fsa}}) and $j$ is an integer. For an
event-type $E$, if $(\alpha,\qB,\wB,j)\in waits(E)$, it denotes that an automaton of the episode
$\alpha$ (with $\alpha.g[j]=E$) is currently in state $(\qB,\wB)$ and is waiting for an event-type $E$ to make
a state transition. Recall from {\em Definition \ref{def:fsa}} that each state of the automaton is specified by a 
pair of
subsets, $(\scrQ^\alpha,\scrW^\alpha)$, of the  set of event-types $X^\alpha$ of $\alpha$. In our
representation, $\qB$ and $\wB$ are $|X^\alpha|$-length binary vectors encoding the two sets
$(\scrQ^\alpha,\scrW^\alpha)$. Consider the earlier example episode $\beta = (A\,B)\rA(C\,D)$. For this, we
have $X^{\beta}=\{A,B,C,D\}$. Suppose this automaton has already accepted an $A$ and $B$ and is waiting for a
$C$ or $D$. So, its current state is $(\{A,B\},\{C,D\})$. This automaton would be listed both in $waits(C)$ and
$waits(D)$. We would have $(\beta, \qB, \wB, 3)\in waits(C)$ and $(\beta,\qB,\wB,4)\in waits(D)$ where
$\qB=[1\,1\,0\,0]$ and $\wB=[0\,0\,1\,1]$. Thus, in general, for an automaton in state
$(\scrQ^\alpha,\scrW^\alpha)$, there would be $|\scrW^\alpha|$ tuples in the different waits lists with the
tuples differing only in the fourth position. As we traverse the data, if the next event is of event-type
$E$, then we acces all the automata waiting for $E$ through $waits(E)$ and effect state transitions. Knowing
the current state of the automaton, we can compute its next state after accepting $E$ because
we have the partial order of the episode stored in $\alpha.e$ array. Since, as explained above, an automaton
can be listed in multiple $waits()$ lists (because it can be waiting for a set of event-types), we have to
ensure that the state transition is   properly reflected in all $waits()$ lists. 

In addition to the $\alpha.g$ and $\alpha.e$ arrays, the other pieces of information that we store with an
episode $\alpha$ are$:$ $\alpha.freq$, $\alpha.init$ and $\alpha.\wB_{start}$. The frequency of an episode is kept track of in $\alpha.freq$.
For each episode $\alpha$, $\alpha.init$ is a list that
keeps track of the times at which the various currently active automata of $\alpha$ made their transition out
of the start state. Each entry in this list is a pair $(\qB,t)$, indicating that an automaton
initialized (i.e., made its first state transition) at
time $t$ is currently in a state with the set of accepted events represented by $\qB$. Since a start-state automaton is yet to make
its first transition, there is no corresponding entry for this in $\alpha.init$. The information in $\alpha.init$ is necessary to 
properly take care of situations where an automaton transits into
a state already occupied by another automaton. It is also useful to check that the span of an occurence satisfies expiry time
constraint before incrementing frequency. $\alpha.\wB_{start}$ is a 
$|X^\alpha|$-length
binary vector encoding the set $\scrW_0^\alpha$, the set of all least elements of $X^\alpha$ (with respect to $R^\alpha$). In other words, it 
encodes the set of
all event-types for which an automaton for $\alpha$ would wait for in its start state. 
Since this information is needed everytime an automaton for the episode is to be initialized, it is useful to precompute it. 
%$bag$ is a general set of candidates which keep track
%of all those episodes for which, an associated automaton in start state has to be added after parsing the current event-type. A
%candidate episode is added to $bag$ either when (i) an associated automaton makes its first state transition on seeing the current
%event-type OR (ii)an automaton that has reached its final state has tracked an occurence satisfying $T_X$.

We now explain the working of {\em Algorithm~\ref{algo:no-exp}} 
%(for event streams with distinct time stamps)
 by referring 
to the line numbers in the pseudocode. Lines $4-12$
initialize all the $waits()$ lists by having one automaton for each candidate episode, waiting in its start state.
%From Defn. 5, it is easy to see that the start state of the automaton for $\alpha$ is represented by the
%pairs of binary vectors $(\mathbf{0},\wB)$ where $\mathbf{0}$ is a vector of all zeros and $\wB[i]=1$ iff
%$\alpha.e[j][i]=0$ $\forall\,j$. 
In the main data pass loop (lines $15-65$), we look at each item $(E_k,t_k)$,
$k=1,2\dots n$, in the event stream and modify the $waits()$ lists to affect state transitions of all automata
waiting for $E_k$.
This is done by accessing each tuple in $waits(E_k)$ list and processing it which is done in the loop starting on line $17$. This is
the main computation in the algorithm and we explain it below. For a tuple  $(\alpha,\qB_{cur},\wB_{cur},j)\in waits(E_k)$, we need to
affect a state transition (since we have seen $E_k$). The next state information for this automaton is denoted as
$\qB_{nxt}$, $\wB_{nxt}$ in the pseudocode. We compute $\qB_{nxt}$ by setting $j^{th}$ bit to one (line $20$). Recall that in the
start state we will have $\qB=\mathbf{0}$ (vector of all zeros). Hence if $\qB_{cur}=\mathbf{0}$, it means that this automaton is
making its first transition out of the start state and hence we add $(\qB_{nxt}, t_k)$ to $\alpha.init$ list in line $22$. 
(Recall
that $\alpha.init$ contains all active automaton for episode $\alpha$ and for each automaton we record its current state and the time
at which it made its transition out of the start state). 
Also, when $\qB_{cur}=\mathbf{0}$, since this automaton is now moving out of
its start state, we need a new automaton for $\alpha$ initialized in its start state. We do this by remembering  $\alpha$ in a temporary
memory called $bag$. (We accumulate all episodes for which new automata are to be initialized in the start state, in this temporary
memory called $bag$ while processing all tuples in $waits(E_k)$. Then, after processing all tuples in $waits(E_k)$, we initialize all
these automata in lines $58-62$). The final state of the automaton corresponds to $\qB$ becoming $\mathbf{1}$, a vector of all ones. If
$\qB_{nxt}\neq 1$, then this automaton, after the current state transition, is still an active automaton for $\alpha$ and hence we
need to update the $\alpha.init$ list by reflecting the new state of this automaton which is done in lines $25-28$. When $\qB_{nxt}\neq 1$,
to complete the computation of its next state, we need to find $\wB_{nxt}$. This automaton has now accepted its $j^{th}$ event type.
Hence, using the partial order information contained in $\alpha.e$ array, we need to find what all new event types it is ready to
accept. Using this, we can compute $\wB_{nxt}$ as done in lines $31$ to $37$. It is computed based on the children of $E_k$ in $R^{\alpha}$ as follows.
It is easy to verify based on {\em Definition \ref{def:fsa}} that $\scrW_{nxt}^{\alpha} = 
(\scrW_{cur}^{\alpha}\backslash E_k) 
\cup
\scrW'$, where 
\begin{displaymath}
\scrW' = \{\textrm{children } E' \,\textrm{ of }\, E_k \textrm{ in } R^{\alpha}\,:\,  \pi_{\alpha}(E)
\subseteq \scrQ^{\alpha}_{nxt} \}
\end{displaymath}
We then need to put this automaton in the $waits()$
list of all those event types that it can accept now. Also, we should modify state information in the $waits()$ lists of event types
corresponding to its previous state. This is done in lines $39-43$.  We point out that in this
process, the $waits()$ lists would end up having duplicate elements if there is already an automaton in the state $\qB_{nxt}$. 
If after the current state transition, the automaton came into a
state in which there is another automaton of this episode, then we have to remove the older automaton. Presence 
of an older automaton is
indicated by an entry $(\qB_{nxt},t')$ for some $t'$ in the $\alpha.init$ list. If $t'<t_{cur}$, where $t_{cur}$ is the time when the
current automaton made its first state transition, then we need to remove the older automaton, which is done in lines $45-47$. We would
also need to remove one of the duplicate elements in the appropriate $waits()$ lists as indicated in line $48-50$\footnote{The steps of automata
transition and check for an older automaton can be combined  and carried out more efficiently. For ease of explanation we have presented the 
two steps separately.}.
%explain the confusion.
If
$\qB_{nxt}=1$ (so that we have now reached the final state), then we need to check whether the span of the occurence tracked by this
automaton satisfies the expiry time constraint. We can compute the span because we know $t_{cur}$, the time at which this automaton
accepted the first event-type, from the entry for this automaton in $\alpha.init$ list. If the span of the occurence tracked is less
than expiry time, then we increment the frequency and remove all the other active automaton of this episode and then start a new
automaton in the start state (lines $52-57$). This completes the explaination of Algorithm~\ref{algo:no-exp}.

In the algorithm discussed above we are implicitly assuming that different events in the data stream have distinct time stamps. This
is because, in the data pass loop (starting on line $15$) an automaton can accept $E_{k+1}$ after accepting $E_k$ in the previous pass
through the loop. We now indicate how one can extend {\em Algorithm~\ref{algo:no-exp}} to handle data with multiple event types having 
the same
time-stamp. For such datastreams, event-types sharing the same time-stamp must be processed together. One needs to perform
unconditional state transitions of all the relevant automata, till all event-types occuring at a given time are parsed. The state
transition step needs a slight modification here compared to that of {\em Algorithm~\ref{algo:no-exp}}.
Consider an automaton for the episode
$(B \, C)\rA D$ waiting in its start state $(\scrQ_{cur}^{\alpha},\scrW_{cur}^{\alpha})$=$(\phi,\{B,C\})$. Suppose we have the event-types
$B$, $C$ and $D$ happening together at a time $t$. Let us denote the set of event-types occuring at time $t$ by $\scrS$. On processing 
$\scrS$, we would need
to accept both $B$ and $C$ but not $D$, though after accepting $B$ and $C$ it transits into a state where it waits for $D$. In general, 
an automaton waiting for a set of event-types $\scrW^\alpha_{cur}$ just before time $t$, should 
accept 
the set of events $\scrS \cap \scrW^\alpha_{cur}$ on seeing the set of event-types $S$ at time $t$. Accordingly,  for the next state,
$\scrQ^\alpha_{nxt}$= 
$\scrQ_{cur}^{\alpha} \cup (\scrS \cap \scrW^\alpha_{cur})$. $\scrW_{nxt}^\alpha$ can be computed from $\scrQ^\alpha_{nxt}$ as in 
{\em Definition \ref{def:fsa}}. Equivalently, we could do the same by processing event-type by event-type as in {\em
Algorithm~\ref{algo:no-exp}}, but such a strategy needs some extra caution. Suppose we had $C$ followed by $B$ and finally followed by $D$ in the event stream, but all with the same
associated time $t$. We parse $C$ and move to a state $(\{C\},\{B\})$. On parsing $B$, we move to a state $(\{B,C\},\{D\})$. Now, next
on processing $D$ if we accept it, we move to $(\{B,C,D\},\phi)$. But $<(C,t),(B,t),(D,t)>$ is not a valid occurence as $D$'s occurence
time must be strictly greater than that of $C$ and $B$. Hence even though we add $(\alpha,[1\,1\,0],[0\,0\,1],3)$ to $waits(D)$ after seeing
$(B,t)$, this potential transition cannot be active at time $t$. The important thing to note that is this element was freshly added to 
$waits()$ after we
started processing $S$. Hence, such potential transition information after adding to $waits()$ must be initially inactive, till, all 
event-types
at the current time are parsed. Such $waits()$ elements must be made active just before parsing event-types of the next time-instant.
After performing the state transitions pertaining to all event types at the current time instant, the rest of the steps are
essentially the same as in  {\em Algorithm~\ref{algo:no-exp}}.
 First, we perform the multiple automata check (there can be more than two automata in the same state now) and removal of all older automata 
 if
necessary. We follow this by the frequency incrementing step.  Since we increment frequency only after parsing all event-types at a given 
time, we need
to store the automata that reach the final state too during the state transition step. Finally, using the $bag$ list, we add 
automata
initialised in the start state, before processing the event-types occuring at the next time tick.

\begin{algorithm}
\tiny{
\caption{CountFrequencyExpiryTime($\scrC_l$, $\mathbb{D}$, $\gamma$, $\scrE$, $T_X$)}
\label{algo:no-exp}
\linesnumbered
\SetKw{KwAnd}{and}
\SetKwData{TRUE}{TRUE}
\SetKwData{FALSE}{FALSE}

\KwIn{Set $\scrC_l$ of candidate episodes, event stream  $\mathbb{D}= \langle (E_1,t_1), \ldots, (E_n,t_n) \rangle$, frequency 
threshold $\gamma$, set $\scrE$ of 
event types (alphabet), Expiry Time, $T_X$}
\KwOut{Set $\scrF$ of frequent episodes out of $\scrC_l$}

\BlankLine
$\scrF_l \leftarrow \phi$ and $bag \leftarrow \phi$ \;
\lForEach{event type $E \in \scrE$}{$waits[E] \leftarrow \phi$}\;

/* Initialization of the $waits()$ lists */

\ForEach{$\alpha \in \scrC_l$}{
	$\alpha.freq \leftarrow 0$  and $\alpha.\wB_{start} \leftarrow \Bzero$\;
	\For{$i \leftarrow 1$ \KwTo $|\alpha|$}{
		$j \leftarrow 1$ \;
		\lWhile{($j \leq |\alpha|$ \KwAnd $\alpha.e[j][i] = 0$)}{$j \leftarrow j+1$ }\;
		\lIf{($j = |\alpha| + 1$)}{$\alpha.\wB_{start}[i] \leftarrow 1$}\;
%		{Add $(\alpha,\Bzero,i)$ to $waits[\alpha.g[i]]$}\;
	}
	\For{$i \leftarrow 1$ \KwTo $|\alpha|$}{
		\If{$\alpha.\wB_{start}[i]=1$}{Add $(\alpha,\Bzero,\alpha.\wB_{start},i)$ to $waits[\alpha.g[i]]$\;}
		/* $\Bzero$ is a vector of all zeros */
	}

}

/* Database pass */

\For{$k \leftarrow 1$ \KwTo $n$}{
	/* $n$ is the number of events in the event stream */

	\ForEach{$(\alpha,\qB_{cur},\wB_{cur},j) \in waits[E_k]$}{
		/* $E_k$ - currently processed event-type in the event stream */

		/* Transit the current automaton to the next state */

		$\qB_{nxt} \leftarrow \qB_{cur}$ and $\qB_{nxt}[j]\leftarrow 1$\;
		
		\If{$\qB_{cur}=\Bzero$}{
		Add $(\qB_{nxt},t_k)$ to $\alpha.init$ and Add $\alpha$ to $bag$\;
		/* $t_k$ - time associated with the current event in event stream */

		}
		\Else{
			\If{$\qB_{nxt}\neq \Bone$}
			{
			/* $\Bone$ is a vector of all ones */

			Update $(\qB_{cur},t_{cur})$  in $\alpha.init$ to $(\qB_{nxt},t_{cur})$\;
			/* $t_{cur}$ would be the first state transition time of the current automaton */

			}
			
		}

		\If{($\qB_{nxt} \neq \Bone$ )}{
			$\wB_{nxt} \leftarrow \wB_{cur}$, $\wB_{nxt}[j] \leftarrow 0$ and $\wB_{temp} \leftarrow \wB_{nxt}$ \;
			\For{$i \leftarrow 1$ \KwTo $|\alpha|$}{
				\If{$\alpha.e[j][i] = 1$}{
					$flg \leftarrow $ \TRUE\;
					\For{($k' \leftarrow 1$; $k' \leq |\alpha|$ \KwAnd $flg  =$ \TRUE; 
					$k'\leftarrow k'+1$)}{
					 	\If{$\alpha.e[k'][i] = 1$ \KwAnd $\qB_{nxt}[k'] = 0$}{
						$flg \leftarrow $ \FALSE\;
						}
					}
					\lIf{$flg = $ \TRUE}{$\wB_{nxt}[i] \leftarrow 1$}\;
					
%						Add $(\alpha,\qB_{nxt},i)$ to $waits[\alpha.g[i]]$\;
%					}

				}
			}
			\For{$i \leftarrow 1$ \KwTo $|\alpha|$}{
				\If{$\wB_{temp}[i]=1$}
				{
					Replace $(\alpha,\qB_{cur}, \wB_{cur},i)$ from $waits[\alpha.g[i]]$ to $(\alpha,\qB_{nxt},
					\wB_{nxt},i)$\;
				}
				\If{$(\wB_{temp}[i]=0$ \KwAnd $\wB_{nxt}[i]=1)$}{
					Add $(\alpha,\qB_{nxt}, \wB_{nxt},i)$ to $waits[\alpha.g[i]]$\;
				}
			}
		
		}
		Remove $(\alpha,\qB_{cur}, \wB_{cur},j)$ from $waits[\alpha.g[j]]$\;
		/* Removing an older automaton if any in the next state */

		\If{($(\qB_{nxt},t')\in\alpha.init$ \KwAnd $t'<t_{cur}$)}{
				/* $t'$ is the first state transition time of an older automaton existing in state
				$\qB_{nxt}$ */

				Remove $(\qB_{nxt},t')$ from $\alpha.init$\;
				\For{$i \leftarrow 1$ \KwTo $|\alpha|$}{
				\If{$\wB_{nxt}[i]=1$}{
					Remove $(\alpha,\qB_{nxt}, \wB_{nxt},i)$ from $waits[\alpha.g[i]]$\;
				}
			}
				}
		/* Increment the frequency */

		\If{($\qB_{nxt} = \Bone$ \KwAnd $(t_k-t_{cur}) \leq T_X$)}{
			$\alpha.freq \leftarrow \alpha.freq + 1$ and Empty $\alpha.init$ list\;
			\For{$i \leftarrow 1$ \KwTo $|\alpha|$}{
				\ForEach{$(\alpha,\qB,\wB,i) \in waits[\alpha.g[i]]$}{
					Remove $(\alpha,\qB,\wB,i)$ from $waits[\alpha.g[i]]$ and Add $\alpha$ to $bag$\;
				}
			}
		}
	}
	/* Add automata initialized in the start state */

	\ForEach{$\alpha \in bag$}{
			\For{$i \leftarrow 1$ \KwTo $|\alpha|$}{
				\If{$\alpha.\wB_{start}[i]=1$}{Add $(\alpha,\Bzero,\alpha.\wB_{start},i)$ to $waits[\alpha.g[i]]$\;}
			%	/* $\Bzero$ is a vector of all zeros */
			}
		}
	Empty $bag$;

}

\lForEach{$\alpha \in \scrC_l$}{\lIf{$\alpha.freq > \gamma$}{Add $\alpha$ to $\scrF_l$}}\;

\Return{$\scrF_l$}
}
\end{algorithm}

\subsection{Space and time complexity of Algorithm~\ref{algo:no-exp}}
\label{sec:no-et-spacetime}
The number of automata that may be active (at any time) for each episode is central to the
space and time complexities of the {\em Algorithm~\ref{algo:no-exp}}.
The number of automata currently active for a given $l$-node episode, $\alpha$, is one more than the number of
elements in the $\alpha.init$ list. 
%This follows from the observation that every time a {\em fresh}
%automaton for $\alpha$ is initialized, i.e.~every time an automaton for $\alpha$ makes a transition out of state
%$0$ (line 17, {\em Algorithm~\ref{algo:no-exp}}), the time instant, $t_k$, of the event
%effecting that transition is recorded by adding an element, $(\qB_{nxt}, t_k)$, to the
%$\alpha.init$ list (where $\qB_{nxt}$ is a binary vector indicating which event-type the automaton
%is accepting to transit out of state $0$). Further, note that before adding $(\qB_{nxt},t_k)$ to the list, we remove
%from $\alpha.init$ any entry that looks like $(\qB_{nxt},t')$. In other words, when two automata for
%$\alpha$ (initialized at different times) reach the same state (i.e.~when they have the same
%$\qB$-vector) we retain only the automaton with more recent initialization time (lines 14-19, {\em
%Algorithm~\ref{algo:no-exp}}). 
%Thus, there can be no more than one entry in $\alpha.init$
%corresponding to every state of the automaton for $\alpha$). The size of the state space, however,
%can be as high as $2^l$, as is the case with $l$-node parallel episodes (i.e.~episodes with an empty partial order), since, if
%$\alpha$ is a parallel episode, every subset of $X^\alpha$ would correspond to a valid
%$\scrQ^\alpha$ (using {\em Property~\ref{property:Qclosed}} for parallel episodes). Does this now
%mean we can have $2^l$ entries in the $\alpha.init$ list? Fortunately, as we shall soon see, 
%this is not the case. In fact, in {\em Algoithm~\ref{algo:no-exp}} there can only be 
%(at most) $l$ entries in the $\alpha.init$ list.
We now show that there can be atmost $l$ entries in the $\alpha.init$ list of {\em Algorithm~\ref{algo:no-exp}}.
Recall that $(\qB_j, t_{i_j}) \in \alpha.init$ means that there is an automaton of episode $\alpha$ that is currently active which
made its transition out of the start state at time $t_{i_j}$ and is currently in state $\qB_j$. Suppose there are $m$ entries in 
$\alpha.init$ list, 
  namely,
$(\qB_1,t_{i_1}), \ldots,$ $(\qB_m,t_{i_m})$, with $t_{i_1} < t_{i_2} < \cdots < t_{i_m}$. Let 
$\{\scrQ^\alpha_1,\ldots,\scrQ^\alpha_m\}$ 
represent the 
corresponding sets of accepted event-types for these active automata. Consider
$k,l$ such that $1\leq k < l \leq m$. 
%The automata initialized at times $t_{i_k}$ and
%$t_{i_l}$ are referred to as the $k^\mathrm{th}$ and $l^\mathrm{th}$ automaton for $\alpha$,
%respectively, and we have $t_{i_k} < t_{i_l}$. 
The events in the data
stream that affected transitions in the $l^\mathrm{th}$ automaton (i.e. automaton which moved out of start state at $t_{i_l}$) would have 
also been {\em seen} by the
$k^\mathrm{th}$ automaton. If the $k^\mathrm{th}$ automaton has not already accepted previous events with the same
event-types, it will do so now on seeing the events which affect the transitions of the $l^{th}$ 
automaton. Hence, $\scrQ^\alpha_l\subsetneq \scrQ^\alpha_k$  for any $1\leq k < l \leq m$. Since
$\scrQ^\alpha\subseteq X^\alpha$  and $|X^\alpha|=l$, there are at most $l$ (distinct) telescoping
subsets of $X^\alpha$, and so, we must have $m\leq l$. 
%We are now ready to derive the space and time complexities for {\em Algorithm~\ref{algo:no-exp}}.

The time required for initialization
in {\em Algorithm~\ref{algo:no-exp}} is $\scrO(|\scrE|+|\scrC_l|l^2)$. This is because,  there are $|\scrE|$ $waits()$ lists to initialize 
and 
it takes $\scrO(l^2)$ time to find the least elements
for each of the $|\scrC_l|$ episodes.
%since there are $|\scrE|$
%$waits()$ lists and since we need to initialize one automaton for each episode in $\scrC_l$ by
%scanning the respective adjacency matrices. 
%Although we initialize only one automaton per episode,
%as we go down the data stream, the algorithm may simultaneously employ as many as $l$ automata for
%each episode in $\scrC_l$. 
For each of the $n$ events in the stream, the corresponding $waits()$ list contains
no more than $l|\scrC_l|$ elements as there can exist atmost $l$-automata per episode. The updates corresponding to each of these entries takes
$\scrO(l^2)$ time to find the new elements to be added to the $waits()$ lists. Thus, the time complexity of the data
pass is $\scrO(nl^3|\scrC_l|)$. 

For each automaton, we store its state information in the binary $l$-vectors $\qB$ and $\wB$. To be able to make 
$|\scrW|$
transitions from a given state, we maintain $|\scrW|$ elements in various $waits()$ lists with each element ready to accept one of the
event-types in $\scrW$. 
%(with the same entry in the first $3$ fields, but 
%with a different 
%entry in the $4^{th}$ field). 
Hence, for each automata we require $\scrO(l^2)$ space to store the state and its possible transitions. Since
there are $l$ such automata in the worst case,
the space complexity is $\scrO(l^3|C|)$.

\section{Candidate Generation}
\label{sec:cand-gen}

Recall that the episode discovery process employs a level-wise procedure where, each level
involves the two steps of candidate generation and frequency counting. 
In Sec.~\ref{sec:counting-algos}, we described the frequency counting algorithms. In this section, we describe 
the candidate generation
algorithm for injective episodes with general partial orders. 
The input to the candidate generation algorithm, 
at level $(l+1)$, is the set, $\scrF_l$,
of frequent episodes of size $l$. Under the  frequency  measure (based on non-overlapped occurences), we know that no
episode can be more frequent than any of its subepisodes. The candidate generation step
exploits this property, to construct the set, $\scrC_{l+1}$, of $(l+1)$-node candidate episodes.

%We first describe the candidate generation algorithm for mining partial orders with no 
%restrictions on its structure. Then  we indicate how the algorithm can be easily specialized to behave either as a frequent serial episode 
%miner 
%or
%a parallel episode miner. We further describe how the candidate generation scheme can be used to mine for partial orders with structural
%constraints based on the number of maximal paths and the length of the largest maximal path.

Recall (cf.~Sec.~\ref{sec:injective-episodes}) that it is simpler to view an injective episode 
$\alpha=(V_{\alpha},<_{\alpha},g_{\alpha})$,  
in terms of its associated {\em
partial order set}, $(X^{\alpha},R^{\alpha})$. 
%Note that $X^\alpha$ is the range of $g_\alpha$ and $R^{\alpha}$ is the partial
%order induced over $X^\alpha$ by $<_\alpha$. 
Each episode in $\scrF_l$ is represented by an $l$-element array of event-types, $\alpha.g$, and an
$l\times l$ matrix, $\alpha.e$, containing the adjacency matrix of the partial order. The array $\alpha.g$ exactly contains the elements of 
$X^{\alpha}$ 
sorted as per the lexicographic
ordering on the alphabet $\scrE$. We refer to $\alpha.g[i]=g_\alpha(v_i)$ as the $i^{th}$ node of $\alpha$. Note that the $i^{th}$ node of 
an episode has no relationship 
whatsoever with the
associated partial order $R^\alpha$. 
%Recall %from {\em Definition~\ref{def:injective-episodes}} 
%that for an episode $\alpha=(V_\alpha, <_\alpha,g_\alpha)$,
%with $V_\alpha=\{v_1,\ldots,v_l\}$, our notation is such that the sequence
%$(g_\alpha(v_1),\ldots,g_\alpha(v_l))$ obeys the lexicographic ordering on the alphabet $\scrE$. Hence we have, 
%$\alpha.g[i]=g_\alpha(v_i),\
%i=1,\ldots,l$. The partial order($R^\alpha$) information is stored in the binary adjacency matrix $\alpha.e$, where 
%$\alpha.e[i][j]=1$ if 
%$v_i<_\alpha v_j$, and $\alpha.e[i][j]=0$ otherwise.  This is because, $v_i<_\alpha v_j$ iff 
%$(g(v_i),g(v_j))\in R^\alpha$. 

 The principal task here is to generate all possible $(l+1)$-node candidates such that, each of their $l$-node 
 subepisodes are frequent. Each $(l+1)$-node candidate is generated by combining two suitable
$l$-node frequent episodes (out of $\scrF_l$). We first explain which pairs of episodes in $\scrF_l$ can be combined and then explain
how to combine them to get $(l+1)$-node episodes. Every pair of $l$-node
frequent episodes, $\alpha,\beta\in\scrF_l$, such that exactly the same $(l-1)$-node subepisode is obtained when their
respective last nodes are dropped, can be combined to obtain  one or more {\em  potential} $(l+1)$-node candidates.
Thus, the episodes $(C\rA A\rA B)$ and $(A\rA D\rA B)$ 
would be combined 
since the same
subepisode, namely $(A\rA B)$ is obtained by dropping the last nodes of $(C\rA A\rA B)$ and $(A\rA D\rA B)$, which are $C$ and $D$
respectively.
Episodes $(C\rA A\rA B)$ and  $((A B) \rA D)$ would not be combined (since different subepisodes,
namely $(A\rA B)$ and $(AB)$, are obtained on dropping the last nodes of $(C\rA A\rA B))$ and $((AB)\rA
D)$).
For every such
constructed candidate episode $\gamma$, if all its $l$-node
subepisodes are frequent, (i.e.~if they can all be found in $\scrF_l$) $\gamma$ is
declared a candidate episode and is added to the output set, $\scrC_{l+1}$. We can formalize this notion of which pairs of episodes
can be combined, as given below.

For an injective episode $\alpha$, 
let
$X^{\alpha}=\{x^{\alpha}_1,\ldots,x^{\alpha}_l\}$ denote the $l$ distinct event-types in
$\alpha$, indexed in lexicographic order. 
%In other words, %as per {\em
%s per the lexicographic order. 
%For each $\alpha_1$ belonging to $\scrF_l$, we combine it with all those
%$\alpha_2$s in $\scrF_l$, 
We combine two episodes $\alpha_1$ and $\alpha_2$ such that the following two conditions hold: (i)~$x^{\alpha_1}_i=x^{\alpha_2}_i,\ 
i=1,\ldots,(l-1)$, $x^{\alpha_1}_l\neq x^{\alpha_2}_l$ and
(ii)~$R^{\alpha_1}\arrowvert_{(X^{\alpha_1}\setminus\{x^{\alpha_1}_l\})}$
$=R^{\alpha_2}\arrowvert_{(X^{\alpha_2}\setminus\{x^{\alpha_2}_l\})}$ (i.e.~the restriction of
$R^{\alpha_1}$ to the first $(l-1)$ nodes of $\alpha_1$ is identical to the restriction of
$R^{\alpha_2}$ to the first $(l-1)$ nodes of $\alpha_2$). To ensure that the same pair of episodes are not picked up two times, we follow
the convention that $\alpha_1$ and $\alpha_2$ are such that $x_l^{\alpha_1}<x_l^{\alpha_2}$ under the lexicographic ordering. 
%(iii)$x^{\alpha_1}$ precedes $x^{\alpha_2}$ as per
%the lexicographic ordering on $\scrE$. 

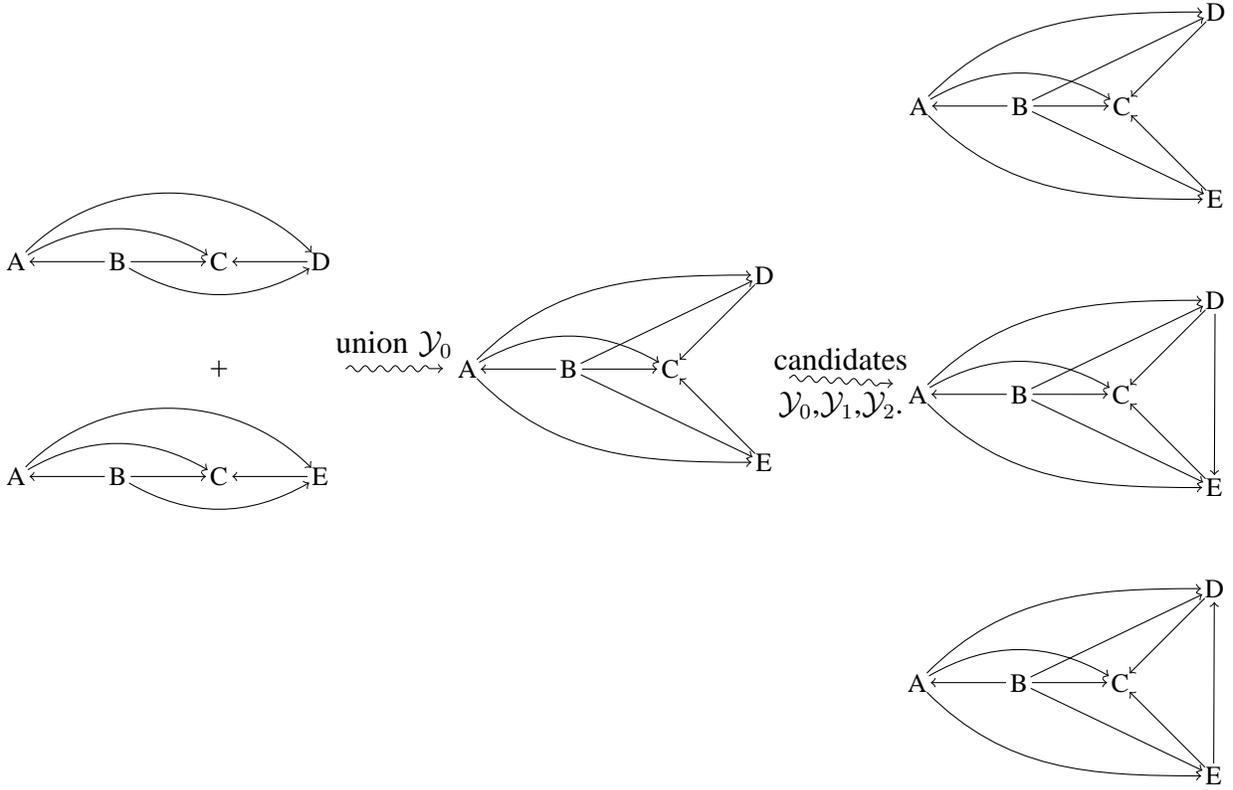
\begin{figure*}
\begin{center}
\pgfdeclarelayer{background}
\pgfsetlayers{background,main}
\begin{tikzpicture}
[ nodeevent/.style={circle,font=\small, inner sep=0pt, minimum size=3mm} ]

\begin{scope}[scale=0.3]
\node[]	 		(plus)				{+};
\node[nodeevent] 	(C1)	[above=of plus] 	{C};
\node[nodeevent]	(B1)	[left=of C1]		{B};
\node[nodeevent]	(D1)	[right=of C1]		{D};
\node[nodeevent]	(A1)	[left=of B1]		{A};
\node[nodeevent] 	(C2)	[below=of plus] 	{C};
\node[nodeevent]	(B2)	[left=of C2]		{B};
\node[nodeevent]	(E2)	[right=of C2]		{E};
\node[nodeevent]	(A2)	[left=of B2]		{A};
\draw[->] (B1) to 	(A1);
\draw[->] (B1) to 	(C1);
\draw[->] (D1) to 	(C1);
\draw[->] (A1) to [out=30,in=150] 	(C1);
\draw[->] (A1) to [out=45,in=135] 	(D1);
\draw[->] (B1) to [out=330,in=210] 	(D1);

\draw[->] (B2) to 	(A2);
\draw[->] (B2) to 	(C2);
\draw[->] (E2) to 	(C2);
\draw[->] (A2) to [out=30,in=150] 	(C2);
\draw[->] (A2) to [out=45,in=135] 	(E2);
\draw[->] (B2) to [out=330,in=210] 	(E2);
\end{scope}

\begin{scope}[xshift=6cm,scale=0.3]
%[ nodeevent/.style={circle, inner sep=0pt, minimum size=3mm} ]
\node[nodeevent]	 		(C3)					{C};
\node[nodeevent]	(D3)	[above right=of C3]		{D};
\node[nodeevent]	(E3)	[below right=of C3]		{E};
\node[nodeevent]	(B3)	[left=of C3]			{B};
\node[nodeevent]	(A3)	[left=of B3]			{A};

\draw[->] (B3) to 	(A3);
\draw[->] (B3) to 	(C3);
\draw[->] (D3) to 	(C3);
\draw[->] (E3) to 	(C3);
\draw[->] (A3) to [out=30,in=150] 	(C3);
\draw[->] (A3) to [out=-45,in=180] 	(E3);
\draw[->] (A3) to [out=45,in=180] 	(D3);
\draw[->] (B3) to 	(E3);
\draw[->] (B3) to 	(D3);
%\draw[->] (C3) to 	(E3);
\end{scope}

\begin{scope}[xshift=12cm,yshift=3.5cm,scale=0.3]
%[ nodeevent/.style={circle, inner sep=0pt, minimum size=3mm} ]
\node[nodeevent]	 		(C4)					{C};
\node[nodeevent]	(D4)	[above right=of C4]		{D};
\node[nodeevent]	(E4)	[below right=of C4]		{E};
\node[nodeevent]	(B4)	[left=of C4]			{B};
\node[nodeevent]	(A4)	[left=of B4]			{A};

\draw[->] (B4) to 	(A4);
\draw[->] (B4) to 	(C4);
\draw[->] (D4) to 	(C4);
\draw[->] (E4) to 	(C4);
\draw[->] (A4) to [out=30,in=150] 	(C4);
\draw[->] (A4) to [out=+45,in=180] 	(D4);
\draw[->] (A4) to [out=-45,in=180] 	(E4);
\draw[->] (B4) to 	(E4);
\draw[->] (B4) to 	(D4);
%\draw[->] (C3) to 	(E3);

\node[nodeevent]	(D5)	[below=of E4]				{D};
\node[nodeevent]	(C5)	[below left=of D5]		{C};
\node[nodeevent]	(E5)	[below right=of C5]		{E};
\node[nodeevent]	(B5)	[left=of C5]			{B};
\node[nodeevent]	(A5)	[left=of B5]			{A};

\draw[->] (B5) to 	(A5);
\draw[->] (B5) to 	(C5);
\draw[->] (D5) to 	(C5);
\draw[->] (E5) to 	(C5);
\draw[->] (A5) to [out=30,in=150] 	(C5);
\draw[->] (A5) to [out=+45,in=180] 	(D5);
\draw[->] (A5) to [out=-45,in=180] 	(E5);
\draw[->] (B5) to 	(E5);
\draw[->] (B5) to 	(D5);
\draw[->] (D5) to 	(E5);

\node[nodeevent]	(D6)	[below=of E5]				{D};
\node[nodeevent]	(C6)	[below left=of D6]		{C};
\node[nodeevent]	(E6)	[below right=of C6]		{E};
\node[nodeevent]	(B6)	[left=of C6]			{B};
\node[nodeevent]	(A6)	[left=of B6]			{A};

\draw[->] (B6) to 	(A6);
\draw[->] (B6) to 	(C6);
\draw[->] (D6) to 	(C6);
\draw[->] (E6) to 	(C6);
\draw[->] (A6) to [out=30,in=150] 	(C6);
\draw[->] (A6) to [out=+45,in=180] 	(D6);
\draw[->] (A6) to [out=-45,in=180] 	(E6);
\draw[->] (B6) to 	(E6);
\draw[->] (B6) to 	(D6);
\draw[->] (E6) to 	(D6);

\end{scope}

\begin{pgfonlayer}{background}
\node[fit=(A1)(B1)(C1)(D1)(A2)(B2)(C2)(E2)(plus)] (r1)	{};
\node[fit=(A3)(B3)(C3)(D3)(E3)]	(r2)	{};
\node[fit=(A4)(B4)(C4)(D4)(E4)(A5)(B5)(C5)(D5)(E5)(A6)(B6)(C6)(D6)(E6)]	(r3)	{};
\end{pgfonlayer}
\draw[->,decorate,decoration={snake, amplitude=.4mm, segment length=2mm, post length=1mm}]
(r1) -- (r2)
node [above,text width=3cm,text centered, midway]
{union $\scrY_0$};

\draw[->,decorate,decoration={snake, amplitude=.4mm, segment length=2mm, post length=1mm}]
(r2) -- (r3)
node [above,text width=3cm,text centered, midway]
{candidates }
node [below,text width=3cm,text centered, midway]
{$\scrY_0$,$\scrY_1$,$\scrY_2$.  };

\end{tikzpicture}
\end{center}
\caption{All combinations $\scrY_0$, $\scrY_1$ and $\scrY_2$ come up.}
\label{fig:Y0Y1Y2}
\end{figure*}

\begin{figure*}
\center
\begin{tikzpicture}
[ nodeevent/.style={circle,font=\small, inner sep=0pt, minimum size=3mm} ]
\pgfdeclarelayer{background}
\pgfsetlayers{background,main}
\begin{scope}[scale=0.3]
\node[]	 		(plus)				{+};
\node[nodeevent] 	(C1)	[above=of plus] 	{C};
\node[nodeevent]	(B1)	[left=of C1]		{B};
\node[nodeevent]	(D1)	[right=of C1]		{D};
\node[nodeevent]	(A1)	[left=of B1]		{A};
\node[nodeevent] 	(C2)	[below=of plus] 	{C};
\node[nodeevent]	(B2)	[left=of C2]		{B};
\node[nodeevent]	(E2)	[right=of C2]		{E};
\node[nodeevent]	(A2)	[left=of B2]		{A};
\draw[->] (B1) to 	(A1);
\draw[->] (B1) to 	(C1);
\draw[->] (D1) to 	(C1);
\draw[->] (A1) to [out=30,in=150] 	(C1);

\draw[->] (B2) to 	(A2);
\draw[->] (B2) to 	(C2);
%\draw[->] (C2) to 	(E2);
\draw[->] (A2) to [out=30,in=150] 	(C2);
\draw[->] (A2) to [out=45,in=135] 	(E2);
\draw[->] (B2) to [out=330,in=210] 	(E2);
\end{scope}

\begin{scope}[xshift=6cm,scale=0.3]
%[ nodeevent/.style={circle, inner sep=0pt, minimum size=3mm} ]
\node[nodeevent]	 		(C3)					{C};
\node[nodeevent]	(D3)	[above right=of C3]		{D};
\node[nodeevent]	(E3)	[below right=of C3]		{E};
\node[nodeevent]	(B3)	[left=of C3]			{B};
\node[nodeevent]	(A3)	[left=of B3]			{A};

\draw[->] (B3) to 	(A3);
\draw[->] (B3) to 	(C3);
\draw[->] (D3) to 	(C3);
\draw[->] (A3) to [out=30,in=150] 	(C3);
\draw[->] (A3) to [out=-45,in=180] 	(E3);
\draw[->] (B3) to 	(E3);
%\draw[->] (C3) to 	(E3);
\end{scope}

\begin{scope}[xshift=12cm,yshift=2cm,scale=0.3]
%[ nodeevent/.style={circle, inner sep=0pt, minimum size=3mm} ]
\node[nodeevent]	 		(C4)					{C};
\node[nodeevent]	(D4)	[above right=of C4]		{D};
\node[nodeevent]	(E4)	[below right=of C4]		{E};
\node[nodeevent]	(B4)	[left=of C4]			{B};
\node[nodeevent]	(A4)	[left=of B4]			{A};

\draw[->] (B4) to 	(A4);
\draw[->] (B4) to 	(C4);
\draw[->] (D4) to 	(C4);
\draw[->] (A4) to [out=30,in=150] 	(C4);
\draw[->] (A4) to [out=-45,in=180] 	(E4);
\draw[->] (B4) to 	(E4);
%\draw[->] (C3) to 	(E3);

\node[nodeevent]	(D5)	[below=of E4]				{D};
\node[nodeevent]	(C5)	[below left=of D5]		{C};
\node[nodeevent]	(E5)	[below right=of C5]		{E};
\node[nodeevent]	(B5)	[left=of C5]			{B};
\node[nodeevent]	(A5)	[left=of B5]			{A};

\draw[->] (B5) to 	(A5);
\draw[->] (B5) to 	(C5);
\draw[->] (D5) to 	(C5);
\draw[->] (A5) to [out=30,in=150] 	(C5);
\draw[->] (A5) to [out=-45,in=180] 	(E5);
\draw[->] (B5) to 	(E5);
\draw[->] (D5) to 	(E5);

\end{scope}

\begin{pgfonlayer}{background}
\node[fit=(A1)(B1)(C1)(D1)(A2)(B2)(C2)(E2)(plus)] (r1)	{};
\node[fit=(A3)(B3)(C3)(D3)(E3)]	(r2)	{};
\node[fit=(A4)(B4)(C4)(D4)(E4)(A5)(B5)(C5)(D5)(E5)]	(r3)	{};
\end{pgfonlayer}

\draw[->,decorate,decoration={snake, amplitude=.4mm, segment length=2mm, post length=1mm}]
(r1) -- (r2)
node [above,text width=3cm,text centered, midway]
{union $\scrY_0$};

\draw[->,decorate,decoration={snake, amplitude=.4mm, segment length=2mm, post length=1mm}]
(r2) to node[above,text width = 3cm, text centered] {candidates} (r3)
%node [above,text width=3cm,text centered, midway]
%{candidates }
node [below,text width=3cm,text centered, midway]
{$\scrY_0$, $\scrY_1$  };

\end{tikzpicture}

\caption{All nodes $A$, $B$ and $C$ prevent $\scrY_2$ from coming up.}
\label{fig:Y0Y1}
\end{figure*}
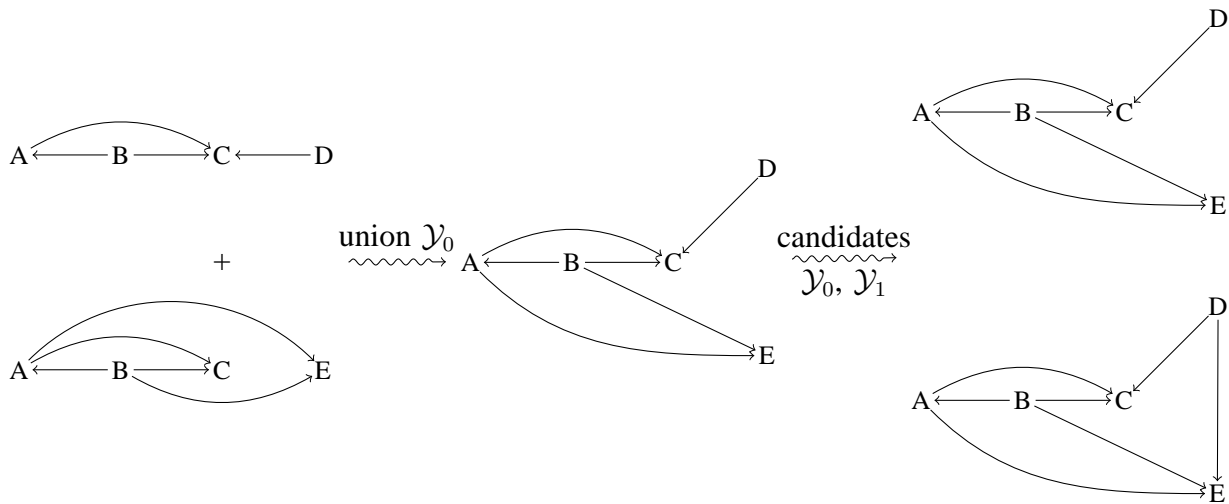

We first illustrate the process of constructing potential candidates through some
examples. Each pair of episodes $\alpha_1$ and $\alpha_2$, sharing the same $(l-1)$-node subepisode on
dropping their respective last nodes can lead to a maximum of three potential candidates, denoted by $\scrY_0$, $\scrY_1$ and $\scrY_2$. 
Consider the 
$\alpha_1$ 
and $\alpha_2$ of Fig.~\ref{fig:Y0Y1Y2}. We construct $\scrY_0$ as a simple
union of $\alpha_1$ and $\alpha_2$, i.e.~we set $X^{\scrY_0}=X^{\alpha_1}\cup X^{\alpha_2}$ and
$R^{\scrY_0}=R^{\alpha_1}\cup R^{\alpha_2}$. As it turns out, in this example, $R^{\scrY_0}$ is a valid partial
order over $X^{\scrY_0}$ (satisfying both anti-symmetry as well as transitive closure) and hence,
$\scrY_0$ is a valid injective episode (and a potential $5$-node candidate). There is no edge in
$\scrY_0$ between the last two nodes (i.e.~the nodes corresponding to event-types $D$ and $E$ 
respectively). By adding an edge from $D$ to $E$ we
get another valid partial order with the relation $R^{\scrY_0}\cup \{(D,E)\}$, and this corresponds to a second
injective candidate, $\scrY_1$, that we can construct using the $\alpha_1$ and $\alpha_2$ of
Fig.~\ref{fig:Y0Y1Y2}. Similarly, $R^{\scrY_0}\cup \{(E,D)\}$ corresponds to a valid partial order and 
this gives us a third potential candidate from the same $\alpha_1$ and $\alpha_2$. But not all
pairs of episodes can be combined in this manner to construct three different potential
candidates. For example, for the $\alpha_1$ and $\alpha_2$ of Fig.~\ref{fig:Y1only}, $\scrY_1$ is the only potential candidate. While
$(X^{\scrY_1},R^{\scrY_1})$ obeys transitive closure, $(X^{\scrY_0},R^{\scrY_0})$ 
is not transitively closed because $(D,C)$ and $(C,E)$ belong to $R^{\scrY_0}$, but $(D,E)$
does not. For the same reason $(X^{\scrY_2},R^{\scrY_2})$ is not transitively closed either. In the
example of Fig.~\ref{fig:Y0Y1}, $\scrY_0$ and $\scrY_1$ are potential candidates 
(but $\scrY_2$ is not a valid potential candidate because $(B,E)$ and $(E,D)$ are in $R^{\scrY_2}$,
while $(B,D)$ is not).

Thus, the general strategy for combining an episode $\alpha_1$ with a valid $\alpha_2$, satisfying the two
conditions mentioned before, is as follows. We attempt to construct an $(l+1)$-node candidate
from $\alpha_1$ and $\alpha_2$, by appending the last node of $\alpha_2$ to the last node of
$\alpha_1$. There are three possibilities to consider for combining $\alpha_1$ and $\alpha_2$:
\begin{eqnarray}
\label{eq:Y0}
X^{\scrY_0} = X^{\alpha_1}\cup X^{\alpha_2} &,& R^{\scrY_0} = R^{\alpha_1}\cup R^{\alpha_2}\\
\label{eq:Y1}
X^{\scrY_1} = X^{\alpha_1}\cup X^{\alpha_2} &,& R^{\scrY_1} = R^{\scrY_0} \cup 
\{(x^{\alpha_1}_l,x^{\alpha_2}_l)\}\\
\label{eq:Y2}
X^{\scrY_2} = X^{\alpha_1}\cup X^{\alpha_2} &,& R^{\scrY_0} = R^{\scrY_0} \cup 
\{(x^{\alpha_2}_l,x^{\alpha_1}_l)\}
\end{eqnarray}
In each case, if $R^{\scrY_j}$ is a valid partial order over $X^{\scrY_j}$, then
the $(l+1)$-node (injective) episode, $(X^{\scrY_j},R^{\scrY_j})$  is considered as a {\em potential} candidate. To verify the same, we 
need to check for the antisymmetry and transitive
closure of the above three possibilities. One can show that each $R^{\scrY_j}$ satisfies antisymmetry because $\alpha_1$ and
$\alpha_2$ share the same $(l-1)$ subepisode on dropping their last nodes.
To check for
transitive closure of $(X^{\scrY_j},R^{\scrY_j})$ we would need to ensure that for every triple $z_1,z_2,z_3\
\in\ X^{\scrY_j}$, if $(z_1,z_2)\in R^{\scrY_j}$ and $(z_2,z_3)\in R^{\scrY_j}$, then we must have
$(z_1,z_3)\in R^{\scrY_j}$. However, since $(R^{\alpha_1}\cup R^{\alpha_2}) \subseteq R^{\scrY_j}$
and since $(X^{\alpha_1},R^{\alpha_1})$ and $(X^{\alpha_2},R^{\alpha_2})$ are already known to be transitively closed,
we need to perform the transitivity closure check only for all size-3 subsets of $X^{\scrY_j}$ that are of the form
$\{x^{\alpha_1}_l, x^{\alpha_2}_l, x^{\alpha_1}_i\::\: 1\leq i\leq (l-1)\}$. Hence, the transitivity closure check is $\scrO(l)$. Finally,
if all the $l$-node subepisodes 
of $\scrY_j$ 
%,each obtained by dropping one of the $(l+1)$ nodes of $\scrY_j$, 
can be found in $\scrF_l$ then $\scrY_j$ is added to the final candidate list, $\scrC_{l+1}$, that
is output by the algorithm.
%example with three combinations

%example with one comb
\begin{figure*}
\center
\begin{tikzpicture}
[ nodeevent/.style={ circle, font=\small, inner sep=0pt, minimum size=3mm} ]
\pgfdeclarelayer{background}
\pgfsetlayers{background,main}
\begin{scope}[scale=0.3]
\node[]	 		(plus)				{+};
\node[nodeevent] 	(C1)	[above=of plus] 	{C};
\node[nodeevent]	(B1)	[left=of C1]		{B};
\node[nodeevent]	(D1)	[right=of C1]		{D};
\node[nodeevent]	(A1)	[left=of B1]		{A};
\node[nodeevent] 	(C2)	[below=of plus] 	{C};
\node[nodeevent]	(B2)	[left=of C2]		{B};
\node[nodeevent]	(E2)	[right=of C2]		{E};
\node[nodeevent]	(A2)	[left=of B2]		{A};
\draw[->] (B1) to 	(A1);
\draw[->] (B1) to 	(C1);
\draw[->] (D1) to 	(C1);
\draw[->] (A1) to [out=30,in=150] 	(C1);

\draw[->] (B2) to 	(A2);
\draw[->] (B2) to 	(C2);
\draw[->] (C2) to 	(E2);
\draw[->] (A2) to [out=30,in=150] 	(C2);
\draw[->] (A2) to [out=45,in=135] 	(E2);
\draw[->] (B2) to [out=330,in=210] 	(E2);
\end{scope}

\begin{scope}[xshift=6cm,scale=0.3]
%[ nodeevent/.style={circle, inner sep=0pt, minimum size=3mm} ]
\node[nodeevent]	 		(C3)					{C};
\node[nodeevent]	(D3)	[above right=of C3]		{D};
\node[nodeevent]	(E3)	[below right=of C3]		{E};
\node[nodeevent]	(B3)	[left=of C3]			{B};
\node[nodeevent]	(A3)	[left=of B3]			{A};

\draw[->] (B3) to 	(A3);
\draw[->] (B3) to 	(C3);
\draw[->] (D3) to 	(C3);
\draw[->] (A3) to [out=30,in=150] 	(C3);
\draw[->] (A3) to [out=-45,in=180] 	(E3);
\draw[->] (B3) to 	(E3);
\draw[->] (C3) to 	(E3);
\end{scope}

\begin{scope}[xshift=12cm,scale=0.3]
%[ nodeevent/.style={circle, inner sep=0pt, minimum size=3mm} ]
\node[nodeevent]	 		(C4)					{C};
\node[nodeevent]	(D4)	[above right=of C4]		{D};
\node[nodeevent]	(E4)	[below right=of C4]		{E};
\node[nodeevent]	(B4)	[left=of C4]			{B};
\node[nodeevent]	(A4)	[left=of B4]			{A};

\draw[->] (B4) to 	(A4);
\draw[->] (B4) to 	(C4);
\draw[->] (D4) to 	(C4);
\draw[->] (A4) to [out=30,in=150] 	(C4);
\draw[->] (A4) to [out=-45,in=180] 	(E4);
\draw[->] (B4) to 	(E4);
\draw[->] (C4) to 	(E4);
\draw[->] (D4) to 	(E4);
\end{scope}

\begin{pgfonlayer}{background}
\node[fit=(A1)(B1)(C1)(D1)(A2)(B2)(C2)(E2)(plus)] (r1)	{};
\node[fit=(A3)(B3)(C3)(D3)(E3)]	(r2)	{};
\node[fit=(A4)(B4)(C4)(D4)(E4)]	(r3)	{};
\end{pgfonlayer}

\draw[->,decorate,decoration={snake, amplitude=.4mm, segment length=2mm, post length=1mm}]
(r1) -- (r2)
node [above,text width=3cm,text centered, midway]
{union $\scrY_0$};

\draw[->,decorate,decoration={snake, amplitude=.4mm, segment length=2mm, post length=1mm}]
(r2) -- (r3)
node [above,text width=3cm,text centered, midway]
{candidates }
node [below,text width=3cm,text centered, midway]
{$\scrY_1$ only};

\end{tikzpicture}

\caption{Edges $(D,C)$ and $(C,E)$ prevent $\scrY_0$ and $\scrY_2$ from coming up as candidates}
\label{fig:Y1only}
\end{figure*}
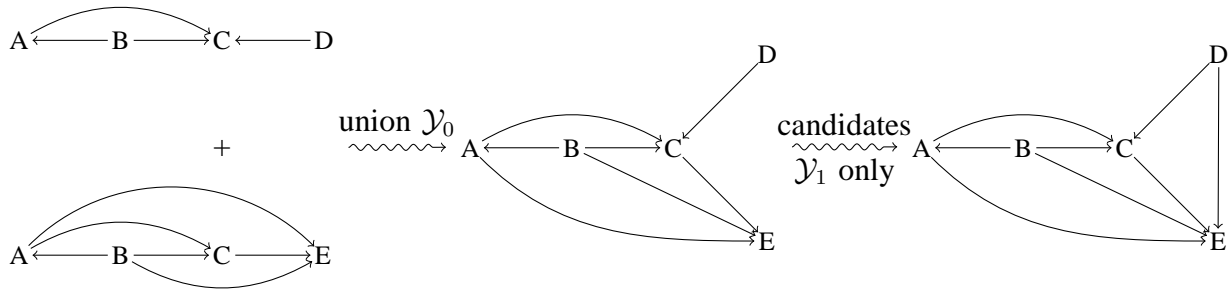

%example with two combinations

\begin{algorithm}
\small{
\caption{GenerateCandidates($\scrF_l$)}
\label{CandGen}
\label{algo-candidate-generation}
\linesnumbered
\SetKw{KwAnd}{and}
\SetKwFunction{GetPotentialCandidates}{GetPotentialCandidates}
\SetKwData{TRUE}{TRUE}
\SetKwData{FALSE}{FALSE}

\KwIn{Sorted array, $\scrF_l $, of frequent episodes of size $l$}
\KwOut{Sorted array, $\scrC_{l+1}$, of candidates of size $(l+1)$}

Initialize $\scrC_{l+1} \leftarrow \phi $ and $k\leftarrow 0$\;
\If{$l=1$}{
	\lFor{$h\leftarrow 1$ \KwTo $|\scrF_l|$}{$\scrF_l[h].blockstart \leftarrow 1$}\;
}

\For{$i\leftarrow 1$ \KwTo $|\scrF_l|$}{

	$currentblockstart \leftarrow k+1$\;
	\For{($ j \leftarrow i+1 $; $ \scrF_l[j].blockstart = \scrF_l[i].blockstart$; $ j \leftarrow j+1$)}{						
	\If{$\scrF_l[i].g[l] \neq \scrF_l[j].g[l]$}{
		$\scrP \leftarrow$ \GetPotentialCandidates{$\scrF_l[i]$, $\scrF_l[j]$}\;
			\ForEach{$\alpha \in \scrP$}{
				$flg \leftarrow$ \TRUE\;
				\For{($r \leftarrow 1$; $r < l$ \KwAnd $flg = $\TRUE; $r \leftarrow r+1$)}{
					\For{$x \leftarrow 1$ \KwTo $r-1$}{
						Set $\beta.g[x] = \alpha.g[x]$\;
						\lFor{$z\leftarrow 1$ \KwTo $r-1$}{$\beta.e[x][z] \leftarrow \alpha.e[x][z]$}\;
						\lFor{$z\leftarrow r$ \KwTo $l$}{$\beta.e[x][z] \leftarrow \alpha.e[x][z+1]$}\;
					}
					\For{$x \leftarrow r$ \KwTo $l$}{
							$\beta.g[x] \leftarrow \alpha.g[x+1]$\;
						\lFor{$z\leftarrow 1$ \KwTo $r-1$}{$\beta.e[x][z] \leftarrow \alpha.e[x+1][z]$}\;
						\lFor{$z\leftarrow r$ \KwTo $l$}{$\beta.e[x][z] \leftarrow \alpha.e[x+1][z+1]$}\;
					}
					\lIf{$\beta \notin \scrF_l$}{$flg \leftarrow$ \FALSE}\;
				}
				\If{$flg = $ \TRUE}{
						$k \leftarrow k+1$\;
						Add $\alpha$ to $\scrC_{l+1}$\;
						$\scrC_{l+1}[k].blockstart \leftarrow currentblockstart$\;
				}
			}
		}
	}
}
\Return{$\scrC_{l+1}$}
}
\end{algorithm}

Interestingly, one need not check whether all the $l$-node subepisodes of a potential $(l+1)$-node candidate 
$\scrY_j$
are in $\scrF_l$. The number of such sub-episodes can in general be very large. It is enough to check whether all 
the $l$-node subepisodes obtained by restricting $R^{\scrY_j}$ to an $l$-node subset of $X^{\scrY_j}$  are present in $\scrF_l$. 
%There are exactly $(l+1)$ such subepisodes. We explain in detail the reason for
%this  below.
%We know that if an $(l+1)$-node 
%injective episode is frequent
%in the non-overlapped sense, then all its $l$-node sub-episodes are also frequent. 
For example, consider a $3$-node episode
$(X^\alpha=\{A,B,C\},R^\alpha=\{(A,B),(A,C)\})$. Its $2$-node sub-episodes are  the serial episodes $A\rA B$ and  $A\rA C$, and parallel
episodes $(A\,B)$, $(A\,C)$ and $(B\,C)$. So in general, an $(l+1)$-node injective episode has more than $(l+1)$ $l$-node subepisodes. 
Let us
consider those $l$-node sub-episodes of $(X^\alpha,R^\alpha)$ which are obtained by restricting $R^\alpha$ to a $l$-subset of $X^\alpha$.
We can have $(l+1)$ such subepisodes. In this example, $A\rA B$, $A\rA C$ and $(B\,C)$ are the three $2$-node subepisodes of $\alpha$
obtained by restricting $R^\alpha$ to all the possible $2$-element subsets of $X^{\alpha}$.  Note that the remaining $2$-node
subepisodes of $A\rA (B\,C)$, namely $(A\,B)$ and $(A\,C)$, are also subepisodes of one or the other of these three $2$-node
subepisodes
. 
For any $N$-node episode $\alpha$, let us denote by $\scrM^\alpha_k$, the set of all $k$-node subepisodes ($k<N$), obtained by restriction of 
$R^\alpha$ 
to $k$-subsets
of $X^\alpha$.  We note the following. For every $k$-node subepisode $\gamma$ of $\alpha$, there exists a 
$\beta \in \scrM^\alpha_k$ such that
$\gamma$ is a subepisode of $\beta$. Also for every $\beta \in \scrM^\alpha_k$ there exists no other $\delta \in \scrM^\alpha_k$ such that
$\beta$ is a subepisode of $\delta$. Hence, $\scrM^\alpha_k$ is maximal for the set of all $k$-node subepisodes of $\alpha$. Therefore
in the rest of the paper, we refer to subepisodes obtained by dropping\footnote{We refer to a subepisode (of an episode $\alpha$)
obtained by restricting $R^{\alpha}$ to a strict subset of $X^\alpha$, as a subepisode obtained by dropping one or more nodes.} one or more nodes as maximal subepisodes. Hence, if all the  
maximal $l$-node subepisodes of a potential $(l+1)$-node candidate are frequent, then all its 
$l$-node subepisodes must also be frequent, which means it is enough to check if all the $l$-node maximal subepisodes of a potential
candidate are frequent. 
%We 
%would in general want to generate an $(l+1)$-node injective 
%episode in
%the candidate list, $\scrC_{l+1}$, iff all its $l$-node subepisodes are frequent. But it is enough if we
%check if 
%all its maximal $l$-node subepisodes are frequent. This is because all maximal $l$-node subepisodes being frequent implies that all
%$l$-node subepisodes are also frequent. 

For a given frequent episode $\alpha_1$, we now describe how one can efficiently search for all other combinable  frequent episodes of 
the same size.
%without any explicit search of the entire set of frequent episodes, at each
%level. 
At level $1$ (i.e. $l=1$), we ensure that $\scrF_1$ is ordered according to the lexicographic ordering on the set of event types
$\scrE$.  
Let $\scrF_l[i]$ 
denote the $i^{th}$ episode in
the collection, $\scrF_l$, of $l$-node frequent episodes. Suppose
$\scrF_1$ consists of the frequent episodes $A$, $C$ and $E$, then we have $\scrF_1[1]=A$, $\scrF_1[2]=C$ and
$\scrF_1[3]=E$. All the three $1$-node episodes share the same sub-episode $\phi$, on dropping their last
event. As per the candidate generation algorithm, any two $1$-node episodes are combined to form a parallel episode and
two serial episodes. Accordingly here, episode $A$ is combined with $C$ and $E$ to form $6$ candidates in
$\scrC_2$. Similarly, $C$ is combined with $E$ to add three more candidates to $\scrC_2$. Note that the first
$6$ candidates share the same $1$-node subepisode $A$ on dropping their last event. Also, the next three
candidates share a similar $1$-node subepisode $C$, on dropping their last event. The candidate generation
procedure adopted at each level here, is such that the episodes which share the same subepisode on dropping
their last events appear consecutively in the generated list of candidates, at each level. We refer to such a
maximal set of episodes as a {\em block}. In addition, we maintain the episodes in each block so that they are ordered
lexicographically with respect to the array of event types. Since, the block information aids us to efficiently decide the kind of
episodes to combine, at each level right from level one, we store the 
block
information. At level $1$, all nodes belong to a single block. For a given $\alpha_1 \in \scrF_l$, the
set of all valid episodes ($\alpha_2$) (satisfying the conditions explained before) with which $\alpha_1$ can be
combined, are all those episodes placed below $\alpha_1$ (except the ones which share the same set of event
types with $\alpha_1$) in the same block.  
%Continuing the example, suppose $(A\,C)$ and $(A\rA C)$ happen to
%be in $\scrF_2$ after counting at level $2$. Even though they share the same block, we would not combine such
%episodes, since to obtain a $3$-node injective episode we need $3$ distinct event-types from the two
%combining episodes. Thus the candidate generation algorithm skips over episodes(in the same block) if they
%have exactly the same sets of event-types(i.e. if $X^{\alpha_1}=X^{\alpha_2}$). 
All candidate episodes
obtained by
combining a given $\alpha_1$ with all permissible episodes ($\alpha_2$) below it in the same block of $\scrF_l$, will
give rise to a block of episodes in $C_{l+1}$, each of them having $\alpha_1$ as their common $l$-node
sub-episode on dropping their last nodes. Hence, the block information of $C_{l+1}$ can be naturally obtained
during its construction itself. 
Even though the episodes within each block are sorted in lexicographic
order of their respective arrays of event-types, we point out that
the full $\scrF_l$ doesn't obey the lexicographic ordering based on the arrays of event-types. For example, 
the episodes 
$((AB)\rA C))$ and $(A\rA
(BC))$ both have the same array of event-types, but would appear in different blocks
(with, for example, an episode like $((AB)\rA D)$ appearing in the same block as $((AB)\rA C)$, while $(A\rA(BC))$, since it belongs to a 
different block, may appear later in $\scrF_l$).

The pseudocode for the candidate generation procedure, {\tt GenerateCandidates()}, is listed in {\em
Algorithm~\ref{algo-candidate-generation}}. The input to {\em
Algorithm~\ref{algo-candidate-generation}} is a collection, $\scrF_l$, of $l$-node frequent
episodes (where, $\scrF_l[i]$ is used to denote the $i^\mathrm{th}$ episode in the collection). The
episodes in $\scrF_l$ are organized in blocks, and episodes within each block appear in
lexicographic order with respect to the array of event types. 
%A maximal set of episodes which share the same $(l-1)$ 
%prefix 
%and
%the same partial order restricted to this prefix set, is referred to as a block. 
We use an array $\scrF_l.blockstart$  to store the block information of every episode.\footnote{a similar array for storing block 
information is used for parallel and serial episode
candidate generation in \cite{MTV97} } $\scrF_l.blockstart[i]$
will hold a value $k$ such that $\scrF_l[k]$ is the first element of the block to which $\scrF_l[i]$ belongs to. The output of the 
algorithm is the collection, $\scrC_{l+1}$, of candidate episodes of size
$(l+1)$.  Initially, $\scrC_{l+1}$ is empty and, if $l=1$, all ($1$-node) episodes are assigned to
the same block (lines 1-3, {\em Algorithm~\ref{algo-candidate-generation}}). The main loop is over
the episodes in $\scrF_l$ (starting on line 4, {\em Algorithm~\ref{algo-candidate-generation}}). The
algorithm tries to combine each episode, $\scrF_l[i]$, with episodes in the same block as
$\scrF_l[i]$ that come after it (line 6, {\em Algorithm~\ref{algo-candidate-generation}}). In
the notation used earlier to describe the procedure, we can think of $\scrF_l[i]$ as $\alpha_1$ and
$\scrF_l[j]$ as $\alpha_2$. If $\scrF_l[i]$ and $\scrF_l[j]$ have identical event-types, we do not
combine them (line 7, {\em Algorithm~\ref{algo-candidate-generation}}). The {\tt
GetPotentialCandidates()} function, takes $\scrF_l[i]$ and $\scrF_l[j]$ as input and returns the
set, $\scrP$, of {\em potential} candidates corresponding to them (line 8, {\em
Algorithm~\ref{algo-candidate-generation}}). This function 
first generates the three potential candidates  by combining $\scrF_l[i]$ and $\scrF_l[j]$ as described in equations
(\ref{eq:Y0}),(\ref{eq:Y1}) and (\ref{eq:Y2}). For each of the three possibilities, it then does a transitive closure check to ascertain
their validity as partial orders\footnote{As explained before, one only needs to do a transitivity check on size-$3$ subsets of the form 
$\{x^{\alpha_1}_l, x^{\alpha_2}_l, x^{\alpha_1}_i\::\: 1\leq i\leq (l-1)\}$ separately on the three possibilities. Actually we can save time
in the transitivity check further. As explained in appendix, we need to generate only the $\scrY_0$ combination 
and perform some special checks on 
its nodes to decide the
valid partial orders to be generated in $\scrP$. }.  For each potential candidate, $\alpha\in\scrP$,
we construct its $l$-node (maximal)subepisodes (denoted as $\beta$ in the pseudocode) 
by dropping one node at-a-time from  $\alpha$ (lines 13-19, {\em
Algorithm~\ref{algo-candidate-generation}}). Note that there is no need to check the case of
dropping the last and last-but-one nodes of $\alpha$, since they would result in the subepisodes
$\scrF_l[i]$ and $\scrF_l[j]$, which are already known to be frequent. If all $l$-node maximal subepisodes
of $\alpha$ were found to be frequent, then $\alpha$ is added to $\scrC_{l+1}$, and its block
information suitably updated (lines 20-24, {\em Algorithm~\ref{algo-candidate-generation}}).

\subsection{Correctness of Candidate Generation}
In this section, we address two important questions regarding the candidate generation. The first question is
whether a given partial order is generated more than once in the algorithm. The second question is about whether every frequent
episode is generated by our candidate generation scheme. 

We now address the first question in detail.
It is easy to see from equation (\ref{eq:Y0}) to (\ref{eq:Y2}) that two partial orders generated from a given pair
$(\alpha_1,\alpha_2)$ of $l$-node episodes are all different. Hence we need to consider whether the same candidate is generated from two
different pairs of episodes. 
%Suppose two exactly same candidates come from the same pair $(\alpha_1,\alpha_2)$. This is not possible, as two candidates generated
%from the same pair $(\alpha_1,\alpha_2)$ certainly differ in the 
%order prescribed between
%$x_l^{\alpha_1}$ and $x_l^{\alpha_2}$(by construction).

Suppose an exactly same candidate is generated from different pairs $(\alpha_1,\alpha_2)$ and $(\alpha'_1,\alpha'_2)$. 
Call them $\scrY_r$ and $\scrY'_s$ where $r$
and $s$ vary from $0$ to $2$ depending on the type of combination of the episode pairs. First consider the case 
when both these candidates come
up as 
$\scrY_0$ and 
$\scrY'_0$. Note that $\scrY_0=(X^{\scrY_0},R^{\scrY_0})$ = 
$(X^{\alpha_1}\cup X^{\alpha_2}, R^{\alpha_1}\cup 
R^{\alpha_2})$ and
$\scrY'_0=(X^{\scrY'_0},R^{\scrY'_0})$ = $(X^{\alpha'_1}\cup X^{\alpha'_2}, R^{\alpha'_1}\cup R^{\alpha'_2})$. Since the candidates are same, $\scrY_0=\scrY'_0$. This implies (i)$X^{\scrY_0}=X^{\scrY'_0}$ and
(ii)$R^{\scrY_0}=R^{\scrY'_0}$. (i) implies  $X^{\alpha_1}\cup X^{\alpha_2}$ = $X^{\alpha'_1}\cup X^{\alpha'_2}$. Recall from the
conditions for forming candidates that  $X^{\alpha_1}\cup X^{\alpha_2}$ = $X^{\alpha_1}\cup \{x^{\alpha_2}_l$\} = $\{x^{\alpha_1}_1,\ldots
x^{\alpha_1}_l, x^{\alpha_2}_l\}$. Recall that $x^{\alpha_1}_i$ is the $i^{th}$ element of $X^{\alpha_1}$, $x^{\alpha_2}_l$ is the
$l^{th}$ element of $X^{\alpha_2}$ and $x_l^{\alpha_1}<x_l^{\alpha_2}$, all as per the lexicograhic ordering on $\scrE$. Hence, 
$x_i^{\alpha_1}$ 
is the $i^{th}$ element of $X^{\scrY_0}$ for $i=1,\ldots l$ and
$x_l^{\alpha_2}$ is its $(l+1)^{th}$ element. An analogous thing holds for $X^{\scrY'_0}$. Since $X^{\scrY_0}$ and $X^{\scrY'_0}$ are
same, their $i^{th}$ elements must also match. This means $x_i^{\alpha_1}=x_i^{\alpha'_1}$ for $i=1,\ldots l$ and $x_l^{\alpha_2}=x_l
^{\alpha'_2}$. This immediately implies $X^{\alpha_1}=X^{\alpha'_1}$. Also from the conditions of generating candidates we have
$x^{\alpha_1}_i=x^{\alpha_2}_i$ and $x^{\alpha'_1}_i=x^{\alpha'_2}_i$ for $i=1,\ldots (l-1)$. This together with
$x_i^{\alpha_1}=x_i^{\alpha'_1}$ for $i=1,\ldots l$ implies $x^{\alpha_2}_i=x^{\alpha'_2}_i$ for $i=1,\ldots (l-1)$. Finally combining
this with $x_l^{\alpha_2}=x_l
^{\alpha'_2}$, we have $X^{\alpha_2}=X^{\alpha'_2}$. Thus $X^{\scrY_0}=X^{\scrY_0'}$ $ \implies$ $X^{\alpha_1}=X^{\alpha_1'}$ and $X^{\alpha
_2}=X^{\alpha_2'}$. Since the pairs $(\alpha_1, \alpha_2)$ and $(\alpha_1',\alpha_2')$ are to be distinct, we need to have either
$(R^{\alpha_2}\neq R^{\alpha'_2})$ or $(R^{\alpha_1}\neq R^{\alpha'_1})$. We now show that this cannot be the case with
  $R^{\scrY_0}=R^{\scrY'_0}$. 
  %We will now show that (ii) which means $R^{\alpha_1}\cup R^{\alpha_2}$ = 
%$R^{\alpha'_1}\cup R^{\alpha'_2}$, implies (iii)$R^{\alpha_1} = R^{\alpha_2}$ and $R^{\alpha'_1} = R^{\alpha'_2}$. If we can show
%(iii),
%we arrive at a contradiction that the two pairs $(\alpha_1,\alpha_2)$ and $(\alpha'_1,\alpha'_2)$  are different. Hence we would have
%shown that every candidate is generated atmost once. We show 
%(iii) by contradiction.  
Suppose  either
$(R^{\alpha_2}\neq R^{\alpha'_2})$ or $(R^{\alpha_1}\neq R^{\alpha'_1})$. Without loss of generality assume, $(R^{\alpha_1}\neq 
R^{\alpha'_1})$. This
means (since $X^{\alpha_1} = X^{\alpha_2}$) there is an edge $(x,y)$ in $R^{\alpha_1}$ that is absent in $R^{\alpha'_1}$ (or the other way round). Again without loss of
generality assume there is an edge $(x,y)$ in $R^{\alpha_1}$ that is absent in $R^{\alpha'_1}$.  Thus, if $R^{\scrY_0}=R^{\scrY'_0}$,
we must have the edge $(x,y)$ in $R^{\alpha_2'}$. By the conditions for candidate
generation, $R^{\scrY'_0}$ can be viewed as the disjoint union of $R^{\alpha'_1}$ and $E^2$, where $E^2$ is the set of all edges in
$R^{\alpha'_2}$ involving $x^{\alpha'_2}_l$. This is because the restriction of
$R^{\alpha'_1}$ to the first $(l-1)$ nodes of $\alpha'_1$ is identical to the restriction of
$R^{\alpha'_2}$ to the first $(l-1)$ nodes of $\alpha'_2$. $(x,y)$ cannot belong to $E^2$ as neither $x$ nor $y$ can be
$x^{\alpha'_2}_l$ (because $x^{\alpha_2'}_l$ does not belong to $X^{\alpha'_1}$ which is same as $X^{\alpha_1}$, which contains both $x$ and
$y$).
Therefore the edge $(x,y) \in R^{\alpha_1}$ and hence in $R^{\scrY_0}$,  cannot appear 
in  $R^{\scrY'_0}$. This contradicts (ii) and hence $R^{\alpha_1} = R^{\alpha_2}$ and $R^{\alpha'_1} = R^{\alpha'_2}$. This means that
the pairs $(\alpha_1, \alpha_2)$ and $(\alpha_1',\alpha_2')$ that we started off with cannot be distinct.

On arguments similar to the $r=s=0$ case, we
can show that no $\scrY_r$ can be equal to any
$\scrY'_s$.
Hence we have shown that every candidate partial order is uniquely generated. We next show that every frequent episode would be in
the set of candidates output by {\em Algorithm \ref{CandGen}}.

We show this by induction on the size of the episode. At level one, the set of candidates contain all the one node episodes and
hence contains all the frequent one node episodes. Now suppose at level $l$, all frequent episodes of size $l$ are indeed generated. If 
an $(l+1)$-node 
episode $\alpha=(X,R)$ is frequent, then all its subepisodes are frequent. The maximal
$l$-node subepisodes $(X\backslash \{x_{l+1}\},R
\arrowvert_{X\backslash
 \{x_{l+1}}\})$ and $(X\backslash \{x_{l}\},R\arrowvert_{X\backslash \{x_{l}}\})$ in particular, are also frequent and hence generated
 at level $l$ (as per the induction hypothesis). Note that the $(l-1)$-node
 subepisodes obtained by dropping the last event-types of these two episodes are the same. Hence, the candidate generation method 
 combines these
2 frequent subepisodes in atmost 3 ways. Since $(X,R)$ is either a $\scrY_0$, $\scrY_1$ or $\scrY_2$ combination of these 2 episodes and
also a valid partial order,
{\em Algorithm \ref{CandGen}} generates it after the first step of candidate generation. The second step checks whether all 
its
remaining  maximal $l$-node subepisodes  are also frequent. This condition is true as per the induction hypothesis and $\alpha$ is therefore
generated in the list of candidates at level $l+1$. Thus we can see that our candidate generation algorithm outputs all valid
candidates without any repetition.
%

%parallel episode comparison 
%The top level algorithm
%for candidate generation of partial order episodes resembles the approach for parallel episodes described
%in \cite{MTV97}.
 %The candidate generation algorithm for parallel
%episodes \cite{MTV97} also employs a similar block structure(but based only on the event-types), where each (parallel) episode is represented by a lexicographically
%sorted array of event-types and the set, $\scrF_l$, of frequent $l$-node (parallel) episodes,
%is maintained as a lexicographically sorted list of (parallel) episodes. 
%We now explain how our candidate generation can be modified to mine in certain subclasses of partial orders.
%For our case of general
%episodes also, each episode is associated with a lexicographically sorted array of event-types. In addition, we have a binary matrix which
%stores the partial order information of these event types here. 
%(Recall %from {\em Definition~\ref{def:injective-episodes}} 

%junk
%The strategy is to form atmost three candidates from $\alpha_1$ and $\alpha_2$ as described in {\em Equations
%\ref{eq:Y0},\ref{eq:Y1} and \ref{eq:Y2}}.  The three possibilities would be $\scrY_0, \scrY_1$ 
 %and $\scrY_2$ as described earlier.
%One can show that the relations $\scrY_0$, $\scrY_1$ and $\scrY_2$ are always asymmetric irrespective of the $\alpha_1$ and
%$\alpha_2$ used to construct them. Interestingly, not all of the
%relations $\scrY_0$, $\scrY_1$ and $\scrY_2$ are transitively closed in general as illustrated by the examples already considered.

\subsection{Candidate Generation with structural constraints}
\label{subsec:cand-gen-structural-constraints}
The candidate generation scheme described above is very flexible. In particular, we can easily specialize it so that we generate only
parallel episodes or only serial episodes. For example, suppose that for every pair of combinable episodes we generate only the
$\scrY_0$ combination (and do not consider the $\scrY_1$ and $\scrY_2$ combinations). Since for all level one episodes, $X^{\alpha}$ is
a singleton and $R^{\alpha}$ is empty, if we do our $\scrY_0$ combination, then $R^{\alpha}$ will be empty for all level-$2$
candidates. Now, since we use our $\scrY_0$ combination throughout, it is easy to see that $R^{\alpha}$ would remain empty at all
levels and then we will be generating only parallel episodes. Similarly, it is easy to see that if we do only $\scrY_1$ and $\scrY_2$
combinations (an do not consider the $\scrY_0$ combination) at all levels then we would generate only serial episodes. Thus the method
we presented for mining general partial orders is easily specialized to a method for parallel episodes or serial episodes only. In
addition, we can also specialize it to mine for certain classes of partial orders as explained below. 

%Suppose, we form only $\scrY_{0}$ combinations of the
%two candidates to be combined, the discovery scheme now mines for all frequent parallel episodes. Performing $\scrY_0$ combinations
%alone from level one generates parallel episodes alone. This generates 
%every frequent $(l+1)$-node parallel episode because, any such episode is a $\scrY_0$ combination of two of its appropriate $l$-node 
%frequent 
%maximal 
%sub-episodes (which are also parallel episodes) . On the other
%extreme, if we only do a {\em valid}  $\scrY_1$ and $\scrY_2$ combination of candidates right from the first level, the discovery scheme 
%mines 
%for all
%frequent injective serial episodes.

Note that any class of partial orders where, for every partial order belonging to the class, all its maximal subepisodes also lie in the
same class, our candidate generation algorithm is easily specialized to such classes of partial orders. We refer to such a class as satisfying a {\em maximal 
subepisode}
property. For example, both the class of serial episodes and parallel episodes satisfy this property. To mine in a specific class of partial
orders, one just needs to do an additional check and
retain only those of the  potential candidates generated, which belong to the class of interest. For mining either in the space of serial or 
parallel orders, one need not perform this explicit check of whether the generated candidates belong to the concerned class. Instead, a more
efficient way  as described earlier can be adopted. 

We discuss a few interesting classes of partial orders satisfying the maximal subepisode property. The first of them is the set of all
partial orders, where length of the largest maximal path of each partial order (denoted as $L_{max}$) is bounded above by a user-defined 
threshold. Consider the episode $\alpha=(A\rA ((F)(B\rA (C\,D)\rA E)))$. It has three maximal paths namely $A\rA B\rA C\rA E$, $A\rA B
\rA D \rA E$ and $A\rA F$ and the length of its largest maximal path is $3$. For $L_{max}= 0$,
we get the set of all parallel episodes because any $N$-node parallel episode has $N$-maximal paths each of length $0$, and every
non-parallel episode has atleast one maximal path of length $1$. In general, for $L_{max}\leq k$, the corresponding class of partial orders
contains all parallel episodes, serial episodes of length less than $(k+1)$ and many more partial orders all of whose maximal paths have 
length
less than $(k+1)$. It is easy to see that for any partial order belonging to such a class, all its subepisodes too belong to the same
class. As $k$ is increased, the class of partial orders expands into the space of all partial orders from the parallel episode end. 
Another class of partial orders of interest could be one, where the number of maximal paths in each partial order (denoted as $N_{max}$)
is bounded above by a threshold. When $N_{max} \leq 1$, the class obtained is exactly equal to the set of serial episodes. For any partial
order
belonging to this class, only its maximal subepisodes are guaranteed to belong to the same class. For example, consider a serial episode
$A\rA B\rA C$. All its maximal sub-episodes are serial episodes. Its non-maximal subepisodes like $(A\,B)$ do not belong to the set of 
serial episodes. 

As
and when the candidates are generated, we calculate and check whether their $L_{max}$ or $N_{max}$ values satisfy the bound constraint. We
use the standard dynamic programming based algorithms to calculate $L_{max}$ or $N_{max}$ on the transitively reduced graph of each
generated candidate partial order. We could also work on a class of partial orders characterised by an upper bound on both $L_{max}$
and $N_{max}$, as such a class would also satisfy the maximal subepisode property. Mining with structural constraints can make the
discovery process more efficient as compared to mining in the class of all injective partial orders. We illustrate with simulation how one 
can mine for partial orders with an upper bound constraint on $L_{max}$ or $N_{max}$.

%Initially, $\scrP$
%is an empty set (line 1, {\em Algorithm~\ref{algo-GetPotentialCandidates}}). The first step in the
%algorithm is to construct $\scrY_0$ as a simple union of $\alpha_1$ and $\alpha_2$ (lines 2-9, {\em
%Algorithm~\ref{algo-GetPotentialCandidates}}). This is done by assigning all the distinct event-types
%in $\alpha_1$ and $\alpha_2$ to the array, $\scrY_0.g[]$ (lines 2-3, 5, {\em
%Algorithm~\ref{algo-GetPotentialCandidates}}). Similarly, the union of partial order
%relations is stored in the  binary matrix, $\scrY_0.e[][]$ (lines 2, 4, 6-9, {\em Algorithm~\ref{algo-GetPotentialCandidates}}).
%Next $\scrY_1$ and $\scrY_2$ are constructed by adding an edge from the last-but-one node to the last node, and vice-versa (lines
%10-11, {\em Algorithm~\ref{algo-GetPotentialCandidates}}). After $\scrY_0$, $\scrY_1$ and $\scrY_2$
%are constructed, the task is to verify if each of them is transitively closed (lines 12-20, {\em
%Algorithm~\ref{algo-GetPotentialCandidates}}). As mentioned earlier, this can be done efficiently,
%since we only need to check for transitive closure of edges involving the last and last-but-one
%node, along with any one other node. Every $\scrY_j$ that passes the transitivity closure check is
%a potential candidates and it is  added to the set, $\scrP$, that is
%returned by the algorithm (lines 21-22, {\em Algorithm~\ref{algo-GetPotentialCandidates}}).

\section{Discussion}
\label{sec:discussion-noninjective}
We wish to point out that the proposed counting and candidate generation algorithms for injective episodes can be extended to a class
of non-injective episodes, where nodes mapped to the same event lie along a chain in the associated partial order. It is interesting to note that the class of all
non-injective serial episodes is contained in this special class of non-injective partial order episodes. To keep the
representation of such $l$-node episodes unambiguos, the $g$-map is restricted (very similar to injective episodes) such that 
$g(v_1),g(v_2),\ldots
g(v_l)$ obey the lexicographic order (total) on $\scrE$. For example, suppose we have a $5$-node episode with $3$ of the nodes mapped
to $A$ and the remaining $2$ mapped to $B$. Then, $g(v_i)$ must be $A$ for $i=1,2,3$ and $B$ for $i=4,5$. Further, since the episodes
are such that the nodes mapped to the same event lie along a chain, we impose a special restriction on $<_\alpha$ to avoid further ambiguity.
Suppose $v_i,v_{i+1},\ldots v_{i+m}$ are mapped to the same event-type $E$. There are $(m+1)!$ total orders possible among these
nodes, each of which would represent the same episode pattern. To avoid this redundancy, 
we restrict $<_\alpha$ to be such that $v_i <_\alpha
v_{i+1} <_\alpha \ldots v_{i+m}$. 

Consider a non-injective episode $\alpha$ with $V_\alpha = \{v_1,v_2,v_3,v_4\}$, $<_\alpha=\{(v_1,v_3),(v_2,v_4)\}$. $g_\alpha()$ is
such that $g_\alpha(v_1)=g_\alpha(v_2)=A$, $g_\alpha(v_3)=B$ and $g_\alpha(v_4)=C$. To track an occurence of such an episode, we would
initially wait for $2$ $A$s. Once we see an $A$, we could either accept the $A$ associated with $v_1$ or $v_2$. Depending on what we
choose, we would now either wait for $\{A,B\}$ (if accepted $A$ is associated with $v_1$) OR $\{A,C\}$ (if accepted $A$ is associated with $v_2$). As per our
currrent counting stratedy, on
seeing $A$ there is more than one next state possible depending on the associated node  in $V_\alpha$. Hence, a non-deterministic finite
state automaton would be the right computational device to track occurences of $\alpha$. In general a non-deterministic finite state
automaton(NFA) would be computationally more expensive compared to a deterministic automaton. Interestingly, $\alpha$ doesn't belong to 
the class of non-injective episodes that
we are considering, i.e. the nodes $v_1$ and $v_2$ are not related even though they map to the same event $A$. We are trying to indicate that to count episodes
like $\alpha$, our strategy of counting leads to automata  which are non-deterministic in nature. Even though an NFA can be converted to an equivalent DFA the
number of states of this equivalent DFA can be huge. Hence, we have noted that counting is also not straight forward for episodes outside the class of non-injective
episodes considered here in addition to problems with representation. We now argue how deterministic finite state automata(DFA) can still be used to track occurences of 
this class of
non-injective episodes, even though in general for non-injective episodes, one requires NFAs OR hugher DFAs to track occurences.

The DFA construction procedure for injective episodes can be generalized to this class of non-injective episodes. Each state would
again be a tuple $(\scrQ^\alpha, \scrW^\alpha)$. $\scrQ^\alpha$ here would be a multiset(essentially a set having repeated elements).
Interestingly, one can verify that $\scrW^\alpha$ is always a proper set for every state in this construction. Suppose not, then
$\scrW^\alpha$ would have atleast $2$ repeated elements. All parents of their corresponding nodes are contained in the set of nodes
associated with that of $\scrQ^\alpha$ (from the constructive definition). Note that the two set of nodes from $V_\alpha$ associated with $\scrW^\alpha$ and
$\scrQ^\alpha$ are disjoint. This means the two nodes which map to repeated elements in $\scrW^\alpha$ are unrelated in $<_\alpha$. But as per
the class of non-injective episodes we are dealing with, two such nodes mapped to the same element must be related, which is a
contradiction. Hence, $\scrW^\alpha$ is
a proper set. Therefore, each transition
from a given state would be on seeing a unique event type. This ensures that the finite state automaton so constructed is
deterministic. Hence the counting algorithms proposed for injective episodes almost exactly go through for this class of injective
episodes.

We now elaborate on the candidate generation. We combine episodes $\alpha=(\{v_1,v_2,\ldots v_l\},<_{\alpha},g_\alpha)$
and $\beta=(\{v_1,v_2,\ldots v_l\},<_{\beta},g_\beta)$ if (i) $g_\alpha(v_i)=g_\beta(v_i)$ $\forall i=1,\dots(l-1)$, (ii)
$<_\alpha|_{\{v_1,v_2,\ldots v_{l-1}\}}$ $=\,<_\beta|_{\{v_1,v_2,\ldots v_{l-1}\}}$, (iii)$g_\alpha(v_l)=g_\beta(v_l)$ OR
$g_\alpha(v_l)$ precedes $g_\beta(v_l)$ as per the lexicographic ordering on $\scrE$. Let
$V_\gamma=\{v_1,\dots,v_l,v_{l+1}\}$. $<_\gamma$ is a relation on $V_\gamma$ defined as follows. $v_i <_\gamma v_j$ iff $v_i<_\alpha
v_j$ $\forall i,j=1,\dots l$. Also, $\forall i=1,2\dots (l-1)$, we have $v_i <_\gamma v_{l+1}$ iff $v_i<_\beta v_l$ and $v_{l+1} <_\gamma v_i$ 
iff $v_l<_\beta v_i$. $g_\gamma$, a map from $V_\gamma$ to $\scrE$ is such that $g_\gamma(v_i)=g_\alpha(v_i)$ $\forall i=1,\ldots l$
and $g_\gamma(v_{l+1})=g_\beta(v_l)$. 

The way we combine $\alpha$ and $\beta$ slightly varies with the two subconditions of (iii). Suppose $g_\alpha(v_l)$ precedes 
$g_\beta(v_l)$ as per the lexicographic ordering on $\scrE$. Consider the following relations
$<_{\gamma0}=<_\gamma$, $<_{\gamma1}=<_\gamma\cup (v_l,v_{l+1})$, $<_{\gamma2}=<_\gamma\cup (v_{l+1},v_l)$. An episode
$(V_\gamma,<_{\gamma i},g_\gamma)$ is generated iff $<_{\gamma i}$ is a partial order. Note that this is exactly similar to what we
have already been doing for injective episodes. The additional thing that needs to be done for this special class of non-injective
episdes is as follows. Suppose $g_\alpha(v_l)=g_\beta(v_l)$, then we
only ask if $<_{\gamma1}$ is a partial order. This is because of the following reason. We have $g_\gamma(v_l)\, =$ $g_\alpha(v_l)
\,=$$g_\beta(v_l)\, =$ $g_\gamma(v_{l+1})$. Hence, $g_\gamma$ maps $v_l$ and $v_{l+1}$ to the same event type. Recall that the
unambigous representation(for the special class of non-injective episodes)  demands that $v_l <_\gamma v_{l+1}$. Hence the only
permissible candidate would be $(V_\gamma,<_{\gamma 1},g_\gamma)$. So we generate this as a candidate if and only if $<_{\gamma 1}$ is
a partial order.

\section{Selection of Interesting Partial Order Episodes}
\label{sec:post-processing}
The frequent episode mining method would ultimately output all 
frequent episodes of upto some size. However, as we see 
in this section, frequency alone is not a sufficient indicator of 
interestingness in case of episodes with general partial orders. 

Consider an $l$-node episode, $\alpha = (X^{\alpha}, R^{\alpha})$. (That is $|X^{\alpha}|=l$). 
If $\alpha$ is frequent then all episodes $\alpha' = (X^{\alpha'}, R^{\alpha'})$ with 
$X^{\alpha'}=X^{\alpha}$ and $R^{\alpha'}\subset R^{\alpha}$ would also be frequent 
$l$-node episodes because every occurrence of $\alpha$ would constitute an occurrence 
of $\alpha'$. The point to note is that when we 
consider episodes with general partial orders, an episode of size $l$ can have subepisodes 
which are also of size $l$. Such a situation does not arise if the mining process is restricted 
to either serial or parallel episodes only. For example there is no 4-node serial episode 
that is a subepisode of $A\rightarrow B \rightarrow C \rightarrow D$. However, when considering 
general partial orders, given a $\alpha = (X^{\alpha}, R^{\alpha})$ there can be, in general, 
exponentially many episodes $\alpha' = (X^{\alpha'}, R^{\alpha'})$ with $X^{\alpha'}=X^{\alpha}$ 
and $R^{\alpha'}\subset R^{\alpha}$. For example, $(A(B \rightarrow C \rightarrow D))$, 
$(B(A \rightarrow C \rightarrow D))$, $(C(A \rightarrow B \rightarrow D))$, $(D(A \rightarrow B \rightarrow C))$,
$(A B)(C \rightarrow D)$, $(A B)\rA C \rA D$, $(ABC)\rA D$, $A \rightarrow (B C) \rightarrow D$ etc. are all such subepisodes 
of $A\rightarrow B \rightarrow C \rightarrow D$. Thus, there is an inherent combinatorial 
explosion in frequent episodes of a given size when we are considering general partial 
orders and, hence, frequency alone may not be a sufficient indicator of `interestingness'. In this section, we propose a new measure,
called {\em bidirectional evidence} of an episode which can be used in conjunction with frequency of an episode to make the mining
process more efficient and meaningful.  

\subsection{Bidirectional evidence}
\label{sec:bidirectional-evidence}

A simple minded strategy to tackle the explosion of frequent episodes could be to use a notion similar to that of  maximal frequent patterns 
that has been used in other datamining contexts such as item sets or sequential patterns.

\begin{definition}
An $\ell$-node episode $\alpha' = (X^{\alpha'}, R^{\alpha'})$ is said to be {\em less specific} 
than $\ell$-node episode  $\alpha = (X^{\alpha}, R^{\alpha})$ if $X^{\alpha'}=X^{\alpha}$ and 
$R^{\alpha'}\subset R^{\alpha}$. Given a set of $\ell$-node episodes, an episode is a most 
specific episode if it is not less specific than any other episode in the set. (Note that, 
in general, there can be many most specific episodes in a given set of episodes).  
\end{definition}

Now, after the mining process (that is, after finding all frequent episodes of size $l$, for a given $l$), we can output only the most specific episodes of the set of frequent episodes. 
This prunes out many partial orders (episodes) 
which are 
presumed uninteresting because a more specific partial order (episode) is frequent and interesting. 
%Also, given this reduced set of frequent episodes, we can generate all the episodes which 
%are less specific and hence  the full set of all frequent episodes. Thus we can think of 
%the specificity filter as a lossless compression of the set of frequent episodes. 
This specificity-based filter is not wholly satisfactory though it reduces the number of 
frequent episodes (of a given size) that are output. Suppose the data actually contains 
the partial order (episode) $(A B) \rightarrow C$. Suppose there are 200 occurrences of this 
 episode of which 110  
are occurrences of $A \rightarrow B \rightarrow C$ while 90 are those of 
$B \rightarrow A \rightarrow C$. Depending on the frequency threshold, suppose one or 
both of these serial episodes are also frequent. 
%Then the specificity filter would output 
%the serial episode(s) and suppress $(A B) \rightarrow C$. 
The parallel episode $(A B C)$, being 
less specific, would also be frequent (and would have a frequency greater than $200$ ). 
The specificity based filter would always suppress the parallel episode $(ABC)$ and importantly also suppress the episode $(A\,B)\rA C$
in preference to any of the serial episodes whenver they are frequent. Thus what we ouput depends very critically on the frequency
threshold. In addition, if is also not satisfactory that whether or not we suppress $(A\,B)\rA C$ depends only on the counts of these
episodes. Instead we can ask is there any evidence in the data to decide which of these partial orders is a better fit. If the data
indeed contains only the partial order $(A\,B)\rA C$ then it would be the case that in most of the occurences of the parallel episode
$(A\,B\,C)$, $C$ follows both $A$ and $B$. We would also see that  in occurences of $(A\,B)\rA C$, $A$ follows $B$ roughly as often as
it precedes $B$. 
%Suppose that in most of the occurrences of $(A B C)$ (counted by the algorithm) $C$ followed $A$ and $B$. 
Now the fact that we have seen $A$ following $B$ roughly as often as 
$A$ preceeding $B$ and that we have rarely seen $C$ not following both $A$ and $B$ should 
mean that the partial order  $(A B) \rightarrow C$ is a better representation of the 
dependencies in data as compared to the serial episode or the parallel episode. 
%For a frequency threshold of $100$, the specificity 
%filter would select the serial episode $A \rA B \rA C$ as the interesting one while ranking based on 
%frequency alone would give higher weightage to the parallel episode (because it would 
%most probably have a few more occurrences). 
Thus, in addition to frequency, it would be 
nice to evaluate interestingness of partial orders based on whether there is evidence in the 
data for not constraining the order of occurence of some pairs of event types. 
That is, we can demand that  
in the occurrences of the episode (as counted by the algorithm) any two event types, 
$i,j \in X^{\alpha}$, such that $i$ and $j$ are not related under $R^{\alpha}$ should 
occur in either order `sufficiently often'. We will now formalize this notion.

 %A serial extension of a partially ordered set $(X^\alpha, R^\alpha)$ 
%is a totally 
%ordered set  $(X^\alpha,R')$ such
%that $R^\alpha\subseteq R'$. 
%A totally ordered set $(X^\alpha,R')$ is said to be compatible with $\alpha=(X^\alpha,R^\alpha)$ if it is a 
%serial extension of $\alpha$. 
%We can say that a specific partial order pattern is interesting (or has enough evidence in data) 
 %if each of its serial extensions contribute substantially to its frequency. Actually, we 
%do not need all serial extensions to contribute. It is enough if the partial order pattern 
%is the only one compatible with all the serial extensions that contribute substantially. 
%However, while counting the frequency of a partial order, it is computationally difficult 
%to keep track of the relative contributions of each of its (relevant) serial extensions. 
%(Note that when a partial order is a candidate episode for counting it is not 
 %necessary that each (or even any) of its serial extensions are candidates). 

Given an episode $\alpha$ let $\scrG^\alpha = \{(i,j) \: : \: i,j \in X^{\alpha}, i\neq j,
(i,j),(j,i)\notin R^\alpha\}$. Let $f^{\alpha}$ denote the number of occurrences (i.e.,  
frequency)  of $\alpha$ counted by our algorithm and let $f_{ij}^{\alpha}$ denote the number of these occurrences 
where $i$ precedes $j$. Let $p^{\alpha}_{ij} = f^{\alpha}_{ij} / f^{\alpha}$. 
%It is easy to see that if all (relevant) serial extensions of $\alpha$ contribute non-zero 
%amounts towards frequency of $\alpha$ then we would have 
 %$f_{ij}^\alpha,p_{ij}^\alpha>0$ for all $(i,j) \in \scrG^\alpha$ and conversely. 
 To rate 
the interestingness of the partial order episode $\alpha$ we define a measure that tries to 
capture the relative magnitudes of $p_{ij}^{\alpha}$ and $p_{ji}^{\alpha}$. Let 
\begin{equation}
H_{ij}^\alpha = -p_{ij}^\alpha log(p_{ij}^\alpha)\,-\,(1-p_{ij}^\alpha) log(1-p_{ij}^\alpha)
\end{equation}
 Since, in each occurrence either $i$ preceeds $j$ or $j$ preceeds $i$, we have 
 $p^{\alpha}_{ij}= 1 - p^{\alpha}_{ji}$ and hence $H_{ij}^\alpha$ is symmetric in $i,j$. Note that $H_{ij}^\alpha$ is the entropy of
 the distribution $[p^{\alpha}_{ij}, (1 - p^{\alpha}_{ij})]$. We refrain from using the term entropy for $H_{ij}^\alpha$, as 
 $p^{\alpha}_{ij} = f^{\alpha}_{ij} / f^{\alpha}$ is tied to the specific subset of occurences counted by our algorithm 
 %and hence it
 %is not clear whether  $p^{\alpha}_{ij}$ is a true estimate of the relevant probability. 
  
The {\em bidirectional evidence} of an episode $\alpha$ denoted by $H(\alpha)$ is defined as follows.
\begin{equation}
H(\alpha) = \min_{(i,j)\in\scrG^\alpha}\,H_{ij}^\alpha
\label{eq:bidirectional-evidence}
\end{equation}

We use $H(\alpha)$ as an additional interestingness measure for $\alpha$. Essentially, if 
$H(\alpha)$ is above some threshold, then there is sufficient evidence that all pairs of 
event types in $\alpha$ that are not constrained by the partial order $R^{\alpha}$ appear 
in either order sufficiently often. We say that an episode $\alpha$ is interesting 
if (i) the frequency is above a threshold, and (ii) $H(\alpha)$ is above a threshold. 
%We call this the post-processing filter based on 
%bidirectional evidence.  

We now explain how $H(\alpha)$ can be computed during our frequency 
counting process. For each
episode, we maintain an $l\times l$ matrix  $\alpha.H$ whose 
$(i,j)^{th}$ element 
would 
contain $f_{ij}^\alpha$ by the end of counting.  For each candidate episode $\alpha$, the matrix $\alpha.H$ is initialized to $0$ just 
before counting. For each
automaton that is initialized, we initialize a separate $l\times l$ matrix of zeros stored with the automaton.
Whenever an automata 
makes state transitions on an event-type $j$, for all $i$ such that event-type
$i$ is already seen, we increment the $(i,j)$  entry in this matrix. The matrix associated with an automaton that reaches its
final state, is added to $\alpha.H$ and results in increment of relevant $f_{ij}^\alpha$ entries. Thus, at the end of the counting,
$\alpha.H$ gives the
$f_{ij}^\alpha$ information.
%At the end of counting, $\alpha.H[i][j]$ would contain $f^{\alpha}_{ij}$.

\subsection{Mining with an additional $H(\alpha)$ threshold}
One can use $H_{\alpha}$ as a postprocessing filter. That is, after the mining process we only  output those $\alpha$ (of a given
size) where $H_{\alpha}$ is a above a threshold. While this may reduce the number of frequent episodes output, it will not make the
mining process efficient. A better way would be to use a threshold on $H(\alpha)$ at each size (or level) in our 
apriori style level-wise counting procedure. This can substantially contribute towards 
the efficiency of mining for general partial orders. 
However, unlike in the case of frequency 
threshold, it is not quite clear whether $H(\alpha)$ also posseses the so called 
anti-monotonicity property. The main difficulty is that $H(\alpha)$ is tied to a 
specific set of occurrences counted by the algorithm. However, if an 
episode $\alpha$ has a bidirectional evidence $H(\alpha) = e$, in a given set of occurences, 
 then one can see that any maximal subepisode of $\alpha$
(obtained by the restriction of $R^\alpha$ onto a subset of $X^\alpha$) 
also has a bidirectional evidence of atleast $e$ in the same set of
occurences. Hence atleast in cases where the embedded pattern's subpepisodes most often occur with the embedded pattern, the bidirectional
evidence of all its maximal subepisodes will be atleast that of the embedded pattern. Since our candidate generation is based on the
existence of all maximal subepisodes at the lower levels, the embedded pattern $\beta$ most often comes up after mining, in the simulations. 
Further,
the bidirectional evidence of all the non-maximal subepisodes of the embedded pattern will be very low (almost zero). This is because of the
following. Any non-maximal subepisode $\gamma$ will not have some edge $(i,j)$ present in the embedded pattern, inspite of the nodes $i$ 
and $j$ 
being present in
$\gamma$. If most occurences of $\gamma$ are also those of $\beta$, $i$ precedes $j$ in almost all occurences of $\gamma$ and hence
$H(\gamma)$ is negligible. Hence almost all non-maximal subepisodes of $\beta$ will have negligible bidirectional evidence inspite of being
frequent. Therefore, we weed out almost all the non-maximal sub-episodes of $\beta$ due to the $H(\beta)$ threshold being incorporated
levelwise. These non-maximal sub-episodes if not weeded out, would otherwise contribute to the generation of many more patterns at various
levels. In particular, it would result in the generation of all the less specific patterns of the embedded pattern as pointed in the
beginning of this section, which doesn't happen now. We show through simulation that mining with $H_{\alpha}$ threshold at each
level is indeed very effective.

\section{Simulation Results}
\label{sec:simulation-results}

\subsection{Synthetic Data Generation}
\label{sec:synthetic-data-generation}

%------------------------------- TABLE 3  ------------------------------done------------------------------

Synthetic data is generated by embedding occurrences of partial orders (episodes) in varying levels
of noise. Input to the data generator is a set of episodes that we want to embed in the data.
For each episode to be embedded, we generate an {\em episode event stream} just containing non-overlapped occurrences of the partial-order
episode. We next generate a separate noise stream involving all event-types. We merge the various episode streams and the
noise stream (that is, string together all events in all the streams in a time-ordered 
 fashion) to generate the final data stream consisting of $T$ time ticks. The data generation process has three user-specified 
parameters: $\eta, p, \rho$, whose roles are explained below.

Each of the episode data streams are generated as follows. To embed each occurrence of an episode, we choose, 
at random, one of its {\em serial extensions}\footnote{A serial extension of a partially ordered set $(X^\alpha, R^\alpha)$ is a totally ordered set  $(X^\alpha,R')$ such that $R^\alpha\subseteq R'$.} 
and then generate the occurrence by having a sequence of event types as needed, with the difference in times 
of occurrence of successive events being geometric with parameter $\eta$. The time between successive 
occurrences of the episode is geometric with parameter $p$. 

We generate the noise stream as follows. Let $\scrE_1$ denote the set of event types that appear 
in any of the embedded episodes. Any event-type not in $\scrE_1$ is referred to as a noise event-type. For each noise event-type, we 
generate a stream of just its occurrences, with time between successive events  geometric with 
parameter $\rho$. Similarly, for each  event-type in $\scrE_1$, we generate a stream of just its occurrences, with time between 
successive events  geometric with 
parameter $\rho/5$. This is done to introduce some random occurrences of the event-types associated with the embedded partial orders. 
All these streams are merged to form a single noise stream. Noise stream is generated in this way so that there may be multiple events 
(constituting noise) at
the same time instant. The noise data stream is merged with all the 
episode 
data streams to obtain the final data stream.

\subsection{Effectiveness of Partial Order Mining}
\label{sec:effectiveness-expts}
We first show that our algorithm is effective in unearthing the embedded partial orders 
in the data stream and also that our new
 measure of interestingness, namely, bidirectional evidence, is very useful in improving  
the efficiency of the mining process. 

We generated a data stream of about $50,000$ events (from a set of $60$ event types) with  $10,000$ time ticks, in which  are embedded the partial orders $\alpha_1=(A\rA (B\,C)\rA 
(D\,E)\rA F)$ and $\alpha_2=(G\rA((H\rA(J\,K))(I\rA L))$ both of which are $6$-node episodes. 
Table~\ref{tab:firstlevel} shows the results obtained with our mining algorithm. 
% with appropriate thresholds on frequency ($f_{th}$),bidirectional evidence ($H_{th}$)
%and expiry time ($T_X$). 
We show the number of candidates ({\em \#Cand}) and the
number of frequent episodes ({\em \#Freq}) at different levels. (Recall that at level $k$, the algorithm finds all frequent episodes
of size $k$). The table  shows the results for the cases: (i) when we only use a threshold on frequency  
($f_{th}$ only), (ii) when we use a threshold on frequency for mining but use a threshold on $H(\alpha) $  as a
post processing filter at each level ($H_{th}$ for post processing) and (iii) when we use a threshold on frequency as well as on 
$H(\alpha)$ at each level ($f_{th}$ and $H_{th}$). (Other
parameters  such as noise levels, thresholds, expiry time etc. are given   in the table caption). 

%\begin{table*}
%\small
%\centering
%\caption{Frequent Episode Output and Run time of the algorithm with and without bidirectional evidence.
%(Patterns: $\alpha_1$ and $\alpha_2$, $n = 10000$, $\eta = 0.3$, $\rho= 0.055 $, $p= 0.068 $, $M= 60$, $T= $, $f_{th} = 350$, $T_{X}=15$, 
%$H_{th}= 0 $)}
%\begin{tabular}{|c|c|c|c|c|c|c|} \hline
%Level & \multicolumn{2}{|c|}{Frequency threshold only} & \multicolumn{2}{|c|}{$H(\alpha)$ for post processing} &
%\multicolumn{2}{|c|}{Threshold on both frequency and $H(\alpha)$} \\ \cline{2-7}
 %& \#Cand & \#Freq & \#Cand & \#Freq & \#Cand & \#Freq \\ \hline 
%1& & & & & & \\ \hline
%2& & & & & & \\ \hline
%3& & & & & & \\ \hline
%4& & & & & & \\ \hline
%5& & & & & & \\ \hline
%6& & & & & & \\ \hline
%Run Time & \multicolumn{2}{|c|}{} & \multicolumn{2}{|c|}{} & \multicolumn{2}{|c|}{} \\ \hline
%\end{tabular}
%\label{tab:firstlevel}
%\end{table*}

\begin{table}[t]
\footnotesize
\centering
\caption{Frequent Episode Output  of the algorithm with and without bidirectional evidence.
(Patterns: $\alpha_1$ and $\alpha_2$, $\eta = 0.7$, $\rho= 0.055 $, $p= 0.068 $, $M= 60$, $T=10000 $, $f_{th} = 350$, $T_X=15$, 
$H_{th}= 0.4 $)}
\begin{tabular}{|c|c|c|c|c|c|c|} \hline
Level & \multicolumn{2}{|c|}{$f_{th}$ only} & \multicolumn{2}{|c|}{$H_{th}$ for post-filtering } &
\multicolumn{2}{|c|}{$f_{th}$ and $H_{th}$} \\ \cline{2-7}
 & \#Cand & \#Freq & \#Cand & \#Freq & \#Cand & \#Freq \\ \hline 
1& 60 	& 60     & 60 	& 60    & 60 	& 60   \\ \hline
2& 5310 & 565    & 5310 & 565   & 5310 & 565   \\ \hline
3& 3810	& 435    & 3810	& 331   & 3810	& 331  \\ \hline
4& 1358 & 760    & 1358 & 129   & 623  & 125   \\ \hline
5& 1861 & 1855   & 1861 & 37    & 36	& 32   \\ \hline
6& 2993 & 2993   & 2993 & 6     & 6 & 6        \\ \hline
7& 0    & 0      & 0    & 0     &0        &0   \\ \hline

Run Time & \multicolumn{2}{|c|}{134s} & \multicolumn{2}{|c|}{142s} & \multicolumn{2}{|c|}{52s} \\ \hline
\end{tabular}
\label{tab:firstlevel}
\end{table}

The two embedded patterns are reported as frequent in all the three cases. 
However, with only a frequency threshold, a lot of
uninteresting patterns (like the subepisodes of the embedded patterns) are also reported frequent. 
When we use a $H(\alpha)$ threshold based post processing filter (case (ii)), the number of candidates 
naturally remains the same, but the frequent episodes output comes down drastically as can be seen 
from the table. However, the run-time actually increases marginally because of the overhead of 
calculating $H(\alpha)$. When we use a threshold on both frequency as well as $H(\alpha)$,  the 
efficiency improves considerably as can be seen from the reduction in number of candidates as well as run-time. 
As can be seen from the table, whether we use threshold on $H(\alpha)$ 
 only for post processing the outputs or also for reducing the 
candidates at each level, we get essentially the same output at all levels; at level 6, the two embedded 
patterns along with some superepisodes are the only ones output. 
We note that even when we use thresholds on both $H(\alpha)$ as well as frequency, we simply refer to
the output as `frequent episodes.'

%We now discuss in greater detail how a threshold on bidirectional evidence at each level while mining 
%is very efficient by looking at the kind of episodes that are declared frequent in each case. 
 %Table~\ref{tab:normal}, Table~\ref{tab:bidirectional-evidence-filter} and 
%Table~\ref{tab:bidirectional-evidence-levelwise}
%give a detailed account of the different kinds of frequent episodes obtained at different levels under the three cases
%discussed earlier. 

\begin{table*}
\small
\centering
\caption{Details of frequent episodes obtained when we use only a frequency threshold. 
(Patterns: $\alpha_1$ and $\alpha_2$, $\eta = 0.7$, $\rho= 0.055 $, $p= 0.068 $, $M= 60$, $T= 10000$, $f_{th} = 350$, $T_{X}=15$. 
)}
\begin{tabular}{|c|c|c|c|c|c|c|c|c|c|c|c|c|} \hline
\multicolumn{3}{|c|}{}	& \multicolumn{4}{c|}{Subepisodes} & \multicolumn{6}{c|}{Non-subepisodes} \\ \hline
Level & \#Cand &\#Freq & \multicolumn{2}{c|}{\#Max} & \multicolumn{2}{c|}{\#Non-max} &\#Noise &\#Mix &\multicolumn{2}{c|}{\#Super}
&\multicolumn{2}{c|}{\#Others} \\ \cline{4-7} \cline{10-13}
 & & & $\alpha_1$ &  $\alpha_2$ &  $\alpha_1$ & $\alpha_2$ &  &  & $\alpha_1$ & $\alpha_2$ & $\alpha_1$ & $\alpha_2$ \\ \hline
 
1& 60 	& 60	&6  	&6 	&0 	&0	&48 	&0	&0 	&0	&0 	&0	\\ \hline	
2& 5310 & 565 	&15 	&15	&8 	&13	&474	&36	&4 	&0	&0	&0 \\ \hline
3& 3810	& 435 	&20 	&20	&49 	&96 	&10 	&214 	&13 	&0	&13	&0 \\ \hline
4& 1358 & 760 	&15 	&15	&142 	&411 	&0 	&52 	&19 	&0	&106	&0 \\ \hline
5& 1861 & 1855 	&6  	&6	&228 	&1268 	&0 	&0 	&12 	&0	&335	&0 \\ \hline
6& 2993 & 2993 	&1 	&1	&174 	&2385 	&0 	&0 	&3 	&0	&429	&0 \\ \hline
\end{tabular}
\label{tab:normal}
\end{table*}

\begin{table*}
\small
\centering
\caption{Details of frequent episodes obtained when we use bidirectional evidence as a post-filter.
(Patterns: $\alpha_1$ and $\alpha_2$, $\eta = 0.7$, $\rho= 0.055 $, $p= 0.068 $, $M= 60$, $T=10000 $, $f_{th} = 350$, $T_{X}=15$, 
$H_{th}= 0.4 $)}
\begin{tabular}{|c|c|c|c|c|c|c|c|c|c|c|c|c|} \hline
\multicolumn{3}{|c|}{}	& \multicolumn{4}{c|}{Subepisodes} & \multicolumn{6}{c|}{Non-subepisodes} \\ \hline
Level & \#Cand &\#Freq & \multicolumn{2}{c|}{\#Max} & \multicolumn{2}{c|}{\#Non-max} &\#Noise &\#Mix &\multicolumn{2}{c|}{\#Super}
&\multicolumn{2}{c|}{\#Others} \\ \cline{4-7} \cline{10-13}
 & & & $\alpha_1$ &  $\alpha_2$ &  $\alpha_1$ & $\alpha_2$ &  &  & $\alpha_1$ & $\alpha_2$ & $\alpha_1$ & $\alpha_2$ \\ \hline
1& 60 	& 60	&6  	&6 	&0 	&0	&48 	&0	&0 	&0	&0 	&0	\\ \hline	
2& 5310 & 565 	&15 	&15	&8 	&13	&474	&36	&4 	&0	&0	&0\\ \hline
3& 3810	& 331 	&20 	&20	&27 	&23 	&10 	&214 	&13 	&0	&4	&0 \\ \hline
4& 1358 & 129 	&15 	&15	&14 	&15 	&0 	&41 	&19 	&0	&10	&0 \\ \hline
5& 1861 & 37 	&6  	&6	&1 	&6 	&0 	&0 	&12 	&0	&6	&0 \\ \hline
6& 2993 & 6 	&1 	&1	&0 	&1 	&0 	&0 	&3 	&0	&0	&0 \\ \hline
\end{tabular}
\label{tab:bidirectional-evidence-filter}
\end{table*}

\begin{table*}
\small
\centering
\caption{Details of frequent episodes obtained when we use Bidirectional Evidence threshold at each level.
(Patterns: $\alpha_1$ and $\alpha_2$, $\eta = 0.7$, $\rho= 0.055 $, $p= 0.068 $, $M= 60$, $T=10000 $, $f_{th} = 350$, $T_{X}=15$, 
$H_{th}= 0.4$)}
\begin{tabular}{|c|c|c|c|c|c|c|c|c|c|c|c|c|} \hline
\multicolumn{3}{|c|}{}	& \multicolumn{4}{c|}{Subepisodes} & \multicolumn{6}{c|}{Non-subepisodes} \\ \hline
Level & \#Cand &\#Freq & \multicolumn{2}{c|}{\#Max} & \multicolumn{2}{c|}{\#Non-max} &\#Noise &\#Mix &\multicolumn{2}{c|}{\#Super}
&\multicolumn{2}{c|}{\#Others} \\ \cline{4-7} \cline{10-13}
 & & & $\alpha_1$ &  $\alpha_2$ &  $\alpha_1$ & $\alpha_2$ &  &  & $\alpha_1$ & $\alpha_2$ & $\alpha_1$ & $\alpha_2$ \\ \hline

1& 60 	& 60	&6  	&6 	&0 	&0	&48 	&0	&0 	&0	&0 	&0	\\ \hline	
2& 5310 & 565 	&15 	&15	&8 	&13	&474	&36	&4 	&0	&0	&0\\ \hline
3& 3810	& 331 	&20 	&20	&27 	&23 	&10 	&214 	&13 	&0	&4	&0 \\ \hline
4& 623  & 125 	&15 	&15	&13 	&15 	&0 	&41 	&19 	&0	&7	&0 \\ \hline
5& 36	& 32	&6  	&6	&1 	&6 	&0 	&0 	&12 	&0	&1	&0 \\ \hline
6& 6 & 6 	&1 	&1	&0 	&1 	&0 	&0 	&3 	&0	&0	&0 \\ \hline

\end{tabular}
\label{tab:bidirectional-evidence-levelwise}
\end{table*}

%this is para from 20-08
Columns {\em \#Cand} and {\em \#Freq}  indicate the
number of candidate and frequent episodes obtained at each level respectively. The remaining columns of this table explains the various 
different kind of
frequent episodes obtained at different levels. 
The columns under {\em Subepisodes} category   indicate the number of frequent subepisodes of the embedded patterns at 
each
level. The columns under {\em Non-subepisodes} category describe the  various frequent episodes which are not subepisodes of 
any of the embedded patterns. Column 
{\em \#Max} 
indicates the number of maximal subepisodes of each embedded pattern at each level. Column {\em \#Non-max} indicates the number of 
non-maximal
subepisodes of both embedded patterns at each level.  Any episode which has an associated noise event-type ($\notin$ $\scrE_1$, the
set of all event-types associated with the embedded partial orders) is referred 
to 
as a {\em noise} episode. $(A\rA Z)$ is a noise episode for example. The number of such 
frequent
noise episodes at each level is given in column {\em \#Noise}. The information of episodes all whose associated event-types are contained in
$\scrE_1$ and necessarily involving event-types from atleast two of the embedded patterns, is tabulated in column {\em \#Mixed}. The current
event-stream, of course is generated by embedding only two patterns. An episode like $(A\rA
B\rA G\rA H)$ is a {\em mixed} episode.   Consider an
episode $\alpha=(X,R)$ either under the {\em super} or {\em others} category (columns {\em \#Super} or {\em \#Others} respectively). All 
event-types from 
$X$ necessarily come from 
one of the embedded patterns (say $\alpha_1$). Consider the maximal subepisode $\alpha_1'$ of $\alpha_1$ obtained by its restriction on
$X$. If $\alpha$ is a super-episode of $\alpha_1'$, then it belongs to the super category. For example,
$(A\rA B\rA C)$ is a super-episode of the maximal subepisode $(A\rA(B\,C))$(of $\alpha_1$). Similarly, $(H\rA J\rA K)$ is a 
super-episode of the maximal subepisode $(H\rA(J\,K)$(of $\alpha_2$). If $\alpha$ is neither a super nor sub-episode of $\alpha_1'$,
then it belongs to others category. For example, consider the maximal subepisode $\alpha_1'=(B\,C)\rA
D$(of $\alpha_1$). $\alpha=(C\rA(B\,D)$ would belong to the others category. {\em \#Init} column in 
Table~\ref{tab:bidirectional-evidence-filter} and Table~\ref{tab:bidirectional-evidence-levelwise}
indicates the number of episodes which are both frequent and have a high enough
$H(\alpha)$.

From Table~\ref{tab:normal} we see that using only a threshold on frequency leads to a total of 
2993 episodes of size 6 being reported as frequent. Of these, two are the embedded patterns (under the 
maximal subepisodes category), 174 and 2385 are non-maximal subepisodes of $\alpha_1$ and $\alpha_2$ 
respectively, 3 are super-episodes of $\alpha_1$ and 429 are spurious episodes that do not contain 
any `noise' event type. The results in Table~\ref{tab:bidirectional-evidence-levelwise} show that when we use a threshold 
on both frequency and $H(\alpha)$, only 6 episodes of size 6 are reported as frequent: the two embedded 
patterns, one non-maximal subepisode of $\alpha_2$ and three super-episodes of $\alpha_1$.
Thus, when we use only a threshold on frequency, most of the episodes reported as frequent are the non-maximal 
subepisodes which can never be eliminated based on their frequencies because they occur at least as 
frequently as the embedded patterns. This is the inherent combinatorial explosion in partial
order mining that we pointed out in Sec.~\ref{sec:bidirectional-evidence}. 
Bidirectional evidence is effective in eliminating these and reporting only the actual partial
orders embedded in the data.  
This is because  patterns grouped under {\em Non-maximal subepisodes} and {\em Others} category would have
a pair of event-types $(i,j)$ which are not related in these episodes, 
but are related in one of the embedded patterns. Since, most of the occurrences of these episodes 
come from the embedded pattern, it is easy to verify from {\em Eq.~\ref{eq:bidirectional-evidence}}  
that almost all these patterns have a very low bidirectional evidence. This effect is seen
at all levels in the tables. From Tables~\ref{tab:bidirectional-evidence-filter} and 
Table~\ref{tab:bidirectional-evidence-levelwise}, we see that the frequent episodes output are essentially 
the same whether we use a post-processing or a level-wise threshold on $H(\alpha)$.
All these results show that using a level-wise threshold on $H(\alpha)$ provides substantial improvement
in efficiency while not missing any important patterns in the set of frequent episodes output.

Also the $H(\alpha)$ based threshold helps us in mining larger sized patterns. For example,  when this algorithm was run (with only a frequency threshold) on a
data stream with an $8$-node episode embedded in it, even after a run-time of
about $300$ seconds, the algorithm was still counting the candidates at level $7$. This is mainly due to the inherent combinatorial
explosion in partial order mining. Most of the  patterns reported in the non-max and others category  at lower levels contribute 
to the 
generation
of a huge number of uninteresting frequent patterns at higher levels, inturn leading to a huge number of candidate patterns at higher levels. Hence, counting at level $7$ was taking a lot of
time. Mining with a $H(\alpha)$ threshold, we could discover the $8$-node embedded episode in a reasonable amount of time.

\subsection{Flexibility in candidate generation}

\begin{table}[t]
\footnotesize
\centering
\caption{Frequent Episodes obtained when the algorithm is run in serial, parallel  and   general mode.
(Patterns: $2$ serial, $2$ parallel, $\alpha_1$ and $\alpha_2$,  $\eta = 0.7$, $\rho= 0.055 $, $p= 0.068 $, $M= 100$, $T=10000 $, 
$f_{th} = 375$, $T_{X}=12$, 
$H_{th}=  0.4$)}
\begin{tabular}{|c|c|c|c|c|c|c|} \hline
Level & \multicolumn{2}{|c|}{Serial mode} & \multicolumn{2}{|c|}{Parallel mode} &
\multicolumn{2}{|c|}{General mode} \\ \cline{2-7}
 & \#Cand & \#Freq & \#Cand & \#Freq & \#Cand & \#Freq \\ \hline 
1&100 &100 &100 &100 &100 &100 \\ \hline
2&9900 &54 &4950 &555 &14850 &609 \\ \hline
3&58 &58 &4830 &71 &6422 &225 \\ \hline
4&34 &34 &37 &33 &184 &156 \\ \hline
5&12 &12 &12 &12 &60 &60 \\ \hline
6&2 &2 &2 &2 &10 &8 \\ \hline
Run Time & \multicolumn{2}{|c|}{58 s} & \multicolumn{2}{|c|}{1 min 28 s} & \multicolumn{2}{|c|}{3m 07 s} \\ \hline
\end{tabular}
\label{tab:serial-parallel-general}
\end{table}

% note have to change the caption and the column headers. 
\begin{table}[t]
\footnotesize
	\centering
	\caption{Results obtained when mining with thresholds on $L_{max}$ and $N_{max}$. 
	$\rho = 0.045 , p = 0.055, \eta = 0.7, M = 100, T = 10000 $, $f_{th} = 300, H_{th} = 0.35, T_{X} = 15$ }
	
	\begin{tabular}{|c|c|c|c|c|} \hline
	$L_{max}$ 	& $N_{max}$ & \#Satisfying(fig.~\ref{fig:embedded-partial-orders})  & \#freq     &Run-time \\	\hline	
    	0 	& 10		& 1 & 1 	& 6 m 29 s			\\ 	\hline
    	2	& 10		& 2 & 3		& 9 m 48 s	\\  \hline
	    5 	& 4		    & 3	& 5		& 9 m 45 s	\\ 	\hline
	    6 	& 2		    & 1 & 2		& 2 m 57 s 	\\ 	\hline
	    7 	& 1		    & 1	& 1		& 53 s			\\ 	\hline
	    7  	& 6		    & 5	& 10	& 9 m 55 s	\\ 	\hline
	    7 	& 18		& 8	& 13	& 10 m 0 s 	\\ 	\hline
	    3 	& 3		    & 0	& 0 	& 9 m 27 s 	\\ 	\hline	
	\end{tabular}
\label{tab:LmaxNmax}
\end{table}

%\begin{table}[t]
%\footnotesize
	%\centering
	%\caption{Run-Time and Memory usage as $L_{max}$ is increased. 
	%$\rho = 0.045 , p = 0.057, \eta = 0.7, M = 100, T = 10000 $, $f_{th} = 320, H_{th} = 0.35$ }
	%
	%\begin{tabular}{|c|c|c|} \hline
	%%$\rho$	&\multicolumn{2}{c|}{Rank}	&\multicolumn{2}{c|}{$f_{rel}$} \\ \cline{2-5}
	%%		&$H_{th}=0.4$	&$H_{th} = 0.9$		&$H_{th}=0.4$	&$H_{th}=0.9$		\\ \hline
	%$L_{max}$ 	&Run-time		&Memory usage \\	\hline	
    	%0 	& 35 s	& 	\\ 	\hline
	%1 	& 1m 50s	& 	\\ 	\hline
	%2 	& 2m 10s	& 	\\ 	\hline
	%3 	& 2m 20s	& 	\\ 	\hline
	%4 	& 2m 20s	& 	\\ 	\hline
	%5 	& 2m 20s	& 	\\ 	\hline
	%6 	& 2m 20s	& 	\\ 	\hline
	%7 	& 2m 20s	& 	\\ 	\hline	
	%\end{tabular}
%\label{tab:Lmax}
%\end{table}

%\begin{table}[t]
	%\footnotesize
	%\centering
	%\caption{Run-Time and Memory usage as $N_{max}$ is increased.
	%$\rho = 0.045 , p = 0.057, \eta = 0.7, M = 100, T = 10000 $, $f_{th} = 320, H_{th} = 0.35$ }
	%\begin{tabular}{|c|c|c|} \hline
	%%$\rho$	&\multicolumn{2}{c|}{Rank}	&\multicolumn{2}{c|}{$f_{rel}$} \\ \cline{2-5}
	%%		&$H_{th}=0.4$	&$H_{th} = 0.9$		&$H_{th}=0.4$	&$H_{th}=0.9$		\\ \hline
	%$N_{max}$ 	&Run-time		&Memory usage \\	\hline	
    %1 	& 1 m	& 	\\ 	\hline
	%2 	& 1 m 45s	& 	\\ 	\hline
	%4 	& 2 m 20s	& 	\\ 	\hline
	%8 	& 2 m 20s	& 	\\ 	\hline
	%18 	& 2 m 20s	& 	\\ 	\hline
	%\end{tabular}
%\label{tab:Nmax}
%\end{table}
% the following are the Nmax values for the different episodes
% [1 2 ... 8] : 8
% [11 12 ..]  : 8
% [21 22 ..]  : 18
% [31 32 ..]  : 4
% [41 42 ..]  : 4
% [51 52 ..]  : 4
% [61 62...]  : 2
% [71 72 ..]  : 1

As described in Section~\ref{subsec:cand-gen-structural-constraints}, the same algorithm (with minor modifications in the candidate
generation) can be used to mine either serial episodes,
parallel episodes or any sub-class of partial orders satisfying
the maximal-subepisode property. To illustrate this, we generated a data stream of about $50,000$ events  
where, in addition to the episodes $\alpha_1$ and $\alpha_2$ defined in
Sec.~\ref{sec:effectiveness-expts}, we 
embedded two more serial episodes and two more parallel episodes.
We ran our algorithm on this data in the serial episode, parallel episode and the general 
modes. When run in the serial episode  mode and the parallel episode mode, we recovered the two
serial and the two parallel episodes respectively. In the general mode, all six
embedded partial orders (along with two other episodes which were superepisodes of the embedded  
partial orders) were obtained. Table~\ref{tab:serial-parallel-general} shows these results.
%number of candidates, frequent
%episodes and the run-times for the three cases.

\begin{figure*}
\begin{center}
%\pgfdeclarelayer{background}
%\pgfsetlayers{background,main}
\begin{tikzpicture}
[ nodeevent/.style={circle,font=\small, inner sep=0pt, minimum size=3mm},
  curved/.style={to path={.. controls +(225:8) and +(210:8) .. (\tikztotarget) \tikztonodes}} ]
\begin{scope}
\node[nodeevent]	 		(A0)				{A};
\node[nodeevent]	 		(B0)	[below=0.3 of A0]		{B};
\node[nodeevent]	 		(C0)	[below=0.3 of B0]		{C};
\node[nodeevent]	 		(D0)	[below=0.3 of C0]		{D};
\node[nodeevent]	 		(E0)	[below=0.3 of D0]		{E};
\node[nodeevent]	 		(F0)	[below=0.3 of E0]		{F};
\node[nodeevent]	 		(G0)	[below=0.3 of F0]		{G};
\node[nodeevent]	 		(H0)	[below=0.3 of G0]		{H};
%\node[font=\scriptsize, below=0.3cm of G0]	(caption0)		[below=of H0]		{(i)$L_{max}=0$ and $N_{max}=8$};
\path (H0) ++(-0,-0.6)  node[font=\scriptsize, text width=3cm, text centered] (caption0) {(i)$L_{max}=0$, $N_{max}=8$ (Parallel
Episode)};

\end{scope}

\begin{scope}[xshift=2.5cm]

\node[nodeevent]	 		(A1)				        {A};
\node[nodeevent]	 		(B1)	[below=0.5cm of A1]		{B};
\node[nodeevent]	 		(C1)	[below=0.5cm of B1]		{C};
\node[nodeevent]	 		(D1)	[below=0.5cm of C1]		{D};
\node[nodeevent]	 		(E1)	[right=1cm of A1]		{E};
\node[nodeevent]	 		(F1)	[right=1cm of B1]		{F};
\node[nodeevent]	 		(G1)	[right=1cm of C1]		{G};
\node[nodeevent]	 		(H1)	[right=1cm of D1]		{H};
\path (H1) ++(-0.5,-0.5)  node[font=\scriptsize] (caption1) {(ii)$L_{max}=1$, $N_{max}=8$};
%\node[font=\footnotesize]	 			(caption0)	[below=of H1]		{(a)$L_{max}=1$, $N_{max}=8$};

\draw[->] (A1) to (E1);
\draw[->] (B1) to (F1);
\draw[->] (C1) to (G1);
\draw[->] (D1) to (H1);
\draw[->] (B1) to (E1);
\draw[->] (C1) to (F1);
\draw[->] (D1) to (G1);
%\draw[->] (A1) to [curved] (H1);
\draw[->] (A1) to (H1);
\end{scope}

\begin{scope}[xshift=5.75cm, yshift=-0.5cm]

\path (0,0) node[nodeevent] (A2) {A}
+(30:1) node[nodeevent] (C2) {C}
++(-30:1) node[nodeevent] (D2) {D}
++(210:1) node[nodeevent] (B2) {B}
++(-30:1) node[nodeevent] (E2) {E};
\path (C2) ++(1,0) node[nodeevent] (F2) {F};
\path (D2) ++(1,0) node[nodeevent] (G2) {G};
\path (E2) ++(1,0) node[nodeevent] (H2) {H};
\path (E2) ++(0,-0.5)  node[font=\scriptsize] (caption2) {(iii)$L_{max}=1$, $N_{max}=18$};
%\node[font=\scriptsize, below=0.5cm of E2]  (caption2) {(iii)$L_{max}=2$, $N_{max}=18$};

%\node[nodeevent]	 		(C1)				{C};
%\node[nodeevent]	 		(D1)	[below=of C1]		{D};
%\node[nodeevent]	 		(E1)	[below=of D1]		{E};
%\node[nodeevent]	 		(A1)	[left=of C1]		{A};
%\node[nodeevent]	 		(B1)	[left=of E1]		{B};
%\node[nodeevent]	 		(F1)	[right=of C1]		{F};
%\node[nodeevent]	 		(G1)	[right=of D1]		{G};
%\node[nodeevent]	 		(H1)	[right=of E1]		{H};
%\node[]	 		(caption0)		[below=of E1]		{(a)$L_{max}=2$ and $N_{max}=27$};
%
\draw[->] (A2) to (C2);
\draw[->] (A2) to (D2);
\draw[->] (A2) to (E2);
\draw[->] (B2) to (C2);
\draw[->] (B2) to (D2);
\draw[->] (B2) to (E2);
\draw[->] (C2) to (F2);
\draw[->] (D2) to (F2);
\draw[->] (E2) to (F2);
\draw[->] (C2) to (G2);
\draw[->] (D2) to (G2);
\draw[->] (E2) to (G2);
\draw[->] (C2) to (H2);
\draw[->] (D2) to (H2);
\draw[->] (E2) to (H2);
\end{scope}

\begin{scope}[xshift=8.5cm, yshift=-0.8cm]

\node[nodeevent]	 		(A3)				{A};
\path (A3) +(20:1) node[nodeevent ] (B3) {B}
+(-20:1) node[nodeevent ] (C3) {C};
\path (B3) ++(1,0) node[nodeevent ] (E3) {E} ++(0,0.5) node[nodeevent ] (D3) {D};
\path (C3) ++(1,0) node[nodeevent ] (F3) {F} ++(0,-0.5) node[nodeevent ] (G3) {G};
\path (F3) ++(20:1) node[nodeevent ] (H3) {H};
\path (G3) ++(-0.5,-0.5) node[font=\scriptsize] (caption3) {(iv)$L_{max}=3$, $N_{max}=4$};
%\node[nodeevent]	 		(B1)	[above right=of A1]		{B};
%\node[nodeevent]	 		(C1)	[below right=of A1]		{C};
%\node[nodeevent]	 		(D1)	[right=of B1]		{D};
%\node[nodeevent]	 		(E1)	[above=of D1]		{E};
%\node[nodeevent]	 		(F1)	[right=of C1]		{F};
%\node[nodeevent]	 		(G1)	[below=of F1]		{G};
%\node[nodeevent]	 		(H1)	[below right=of D1]		{H};
%\node[]	 		(caption0)		[below=of G1]		{(a)$L_{max}=3$ and $N_{max}=16$};
%
\draw[->] (A3) to (B3);
\draw[->] (A3) to (C3);
\draw[->] (B3) to (D3);
\draw[->] (B3) to (E3);
\draw[->] (C3) to (F3);
\draw[->] (C3) to (G3);
\draw[->] (D3) to (H3);
\draw[->] (E3) to (H3);
\draw[->] (F3) to (H3);
\draw[->] (G3) to (H3);
\end{scope}

\begin{scope}[xshift=12cm, yshift=-0.8cm]% yshift=-4.5cm]
\path (0,0) node[nodeevent] (A4) {A} 
++(1,0) node[nodeevent ] (B4) {B}
++(1,0) node[nodeevent ] (C4) {C}
++(1,0) node[nodeevent ] (D4) {D};

\path (D4) +(45:1) node[nodeevent ] (E4) {E}
+(15:1) node[nodeevent ] (F4) {F}
+(-15:1) node[nodeevent ] (G4) {G}
+(-45:1) node[nodeevent ] (H4) {H};
\path (C4) ++(0,-1.5) node[font=\scriptsize ] (caption4) {(v)$L_{max}=4$, $N_{max}=4$};
\draw[->] (A4) to (B4);
\draw[->] (B4) to (C4);
\draw[->] (C4) to (D4);
\draw[->] (D4) to (E4);
\draw[->] (D4) to (F4);
\draw[->] (D4) to (G4);
\draw[->] (D4) to (H4);
\end{scope}

\begin{scope}[xshift=2.5cm,  yshift=-4.75cm]
\path (0,0) node[nodeevent] (A4) {A} 
++(1,0) node[nodeevent ] (B4) {B}
++(1,0) node[nodeevent ] (C4) {C}
++(1,0) node[nodeevent ] (E4) {E}
+(150:1) node[nodeevent ] (D4) {D}
++(1,0) node[nodeevent ] (F4) {F}
+(30:1) node[nodeevent ] (G4) {G}
+(-30:1) node[nodeevent ] (H4) {H};

\path (C4) ++(0.5,-1) node[font=\scriptsize ] (caption4) {(vi)$L_{max}=5$, $N_{max}=4$};
\draw[->] (A4) to (B4);
\draw[->] (B4) to (C4);
\draw[->] (C4) to (E4);
\draw[->] (D4) to (E4);
\draw[->] (E4) to (F4);
\draw[->] (F4) to (G4);
\draw[->] (F4) to (H4);
\end{scope}

\begin{scope}[xshift=9.5cm,  yshift=-3.25cm]
\path (0,0) node[nodeevent] (A5) {A} 
+(-20:1) node[nodeevent ] (C5) {C}
++(1,0) node[nodeevent ] (B5) {B}
++(1,0) node[nodeevent ] (D5) {D}
++(1,0) node[nodeevent ] (E5) {E}
++(1,0) node[nodeevent ] (F5) {F}
++(1,0) node[nodeevent ] (G5) {G}
++(1,0) node[nodeevent ] (H5) {H};

\path (E5) ++(0,-0.75) node[font=\scriptsize ] (caption5) {(vii)$L_{max}=7$, $N_{max}=2$};
\draw[->] (A5) to (B5);
\draw[->] (A5) to (C5);
\draw[->] (B5) to (D5);
\draw[->] (D5) to (E5);
\draw[->] (E5) to (F5);
\draw[->] (F5) to (G5);
\draw[->] (G5) to (H5);
\end{scope}

\begin{scope}[xshift=9cm,  yshift=-5cm]
\path (0,0) node[nodeevent] (A5) {A} 
++(1,0) node[nodeevent ] (B5) {B}
++(1,0) node[nodeevent ] (C5) {C}
++(1,0) node[nodeevent ] (D5) {D}
++(1,0) node[nodeevent ] (E5) {E}
++(1,0) node[nodeevent ] (F5) {F}
++(1,0) node[nodeevent ] (G5) {G}
++(1,0) node[nodeevent ] (H5) {H};

\path (D5) ++(0.5,-0.75) node[font=\scriptsize, text width=3cm, text badly centered] (caption5) {(viii)$L_{max}=8$, $N_{max}=1$ (Serial
Episode)};
\draw[->] (A5) to (B5);
\draw[->] (B5) to (C5);
\draw[->] (C5) to (D5);
\draw[->] (D5) to (E5);
\draw[->] (E5) to (F5);
\draw[->] (F5) to (G5);
\draw[->] (G5) to (H5);
\end{scope}

\end{tikzpicture}
\end{center}
\caption{Partial Order Episodes used for embedding in the data streams.}
\label{fig:embedded-partial-orders}
\end{figure*}
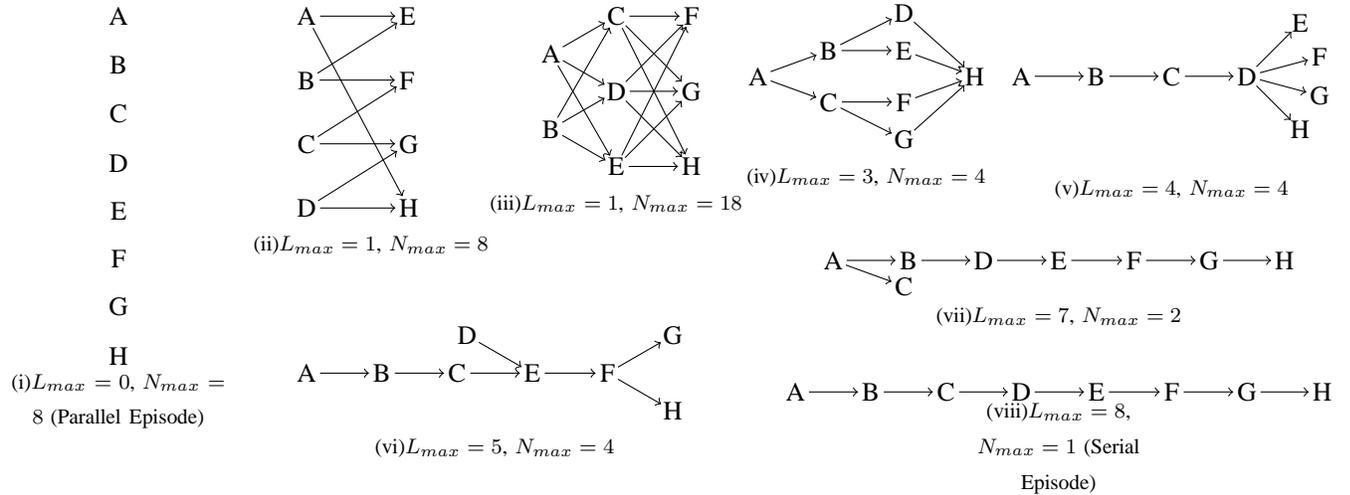

Next, we generated synthetic data  by embedding all the $8$ partial orders of
Figure~\ref{fig:embedded-partial-orders}. 
Recall that $L_{max}$ for a partial order is the length of its largest maximal path. Similarly,
$N_{max}$ for a partial order is the number of maximal paths in it. We present results obtained by 
mining in this data under different thresholds on $L_{max}$ and $N_{max}$ (Table~{\ref{tab:LmaxNmax}}).   
%The threshold on $L_{max}$ was progressively increased from $0$ to $7$ and in all cases the embedded partial orders which
%satisfied the threshold constraint on $L_{max}$ were discovered and the ones which didn't were pruned by the algorithm. A similar thing was
%observed with variation of the threshold on $N_{max}$. 
%Tables~\ref{tab:Lmax} and~\ref{tab:Nmax} describe the run-times and
%memory-usage when the discovery algorithm is run with upper bounds on $L_{max}$ and $N_{max}$ respectively. 
%Tables~\ref{tab:LmaxNmax} describes the run-times when the discovery algorithm is run with an upper bound constraint on both $L_{max}$ and 
%$N_{max}$ simultaneously. 
The column titled {\em 'Satisfying (fig.~\ref{fig:embedded-partial-orders})'} refers to the number of partial orders in 
Figure~\ref{fig:embedded-partial-orders}
which satisfy the corresponding $L_{max}$ and $N_{max}$ constraints. 
We get all the embedded patterns that satisfy the $L_{max}$, $N_{max}$ constraint as
frequent episodes along with a few extra episodes (as seen under \#freq).
From the table we see that at lower thresholds on either $L_{max}$ OR $N_{max}$, 
the algorithm runs faster. At higher thresholds, the run-times were almost the same as 
those for mining all partial orders. This is because most of the computational burden is due to large number of candidates
at levels $2$ and $3$, and the candidates at these lower levels are not reduced if the bounds on $L_{max}$ and $N_{max}$ are 
high.

\subsection{Scaling and other properties of the algorithm}

\begin{figure}
	\centering
	\begin{tikzpicture}[scale=0.7]
		\begin{semilogyaxis}[
			xlabel=$f_{th}$,
			ylabel=Number of $8$-node frequent episodes, grid=both]
		%\addplot[mark=x] plot file {./Raajayfigures/figures/frequencies-txt/freqGraph-xy-reduced-2};
		\addplot[mark=o] plot file {./freqGraph-xy-reduced-5};
		%\legend{$N_{emb}=2$, $N_{emb}=5$}
		\end{semilogyaxis}
	\end{tikzpicture}
\caption{Variation in number of frequent episodes as a function of frequency threshold, No. of embedded episodes, 
$N_{emb}=5$,  $\rho = 0.055 , p = 0.068, \eta = 0.7, M = 100, T = 10000 , 
H_{th} = 0.75, T_X=15$.}
\label{fig:frequency-threshold-variation}
\end{figure}
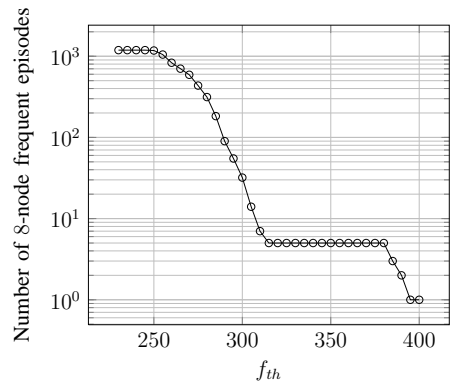
%(i)(iii)(v)(vi)(viii)
\begin{figure}
	\centering
	\begin{tikzpicture}[scale=0.7]
		\begin{semilogyaxis}[
			xlabel=$H_{th}$,
			ylabel=Number of $8$-node frequent episodes, grid=both]
		\addplot[mark=x, mark size=2] plot file {./entGraph-xy-5nodes-reduced};
		%\addplot plot file {./Raajayfigures/figures/frequencies-txt/freqGraph-xy-5};
		%\legend{$N_{emb}=2$, $N_{emb}=5$}
		\end{semilogyaxis}
	\end{tikzpicture}
\caption{Variation in number of frequent episodes as a function of $H(\alpha)$ threshold.  $f_{th} = 360$, 
rest same as previous fig.}
%Embedded episodes
%(fig.~\ref{fig:embedded-partial-orders}):(i)(iii)(v)(vi)(viii), $\rho = 0.055 , p = 0.068, \eta = 0.3, M = 100, T = 10000 , 
%, T_X=15$.}
\label{fig:H-threshold-variation}
\end{figure}
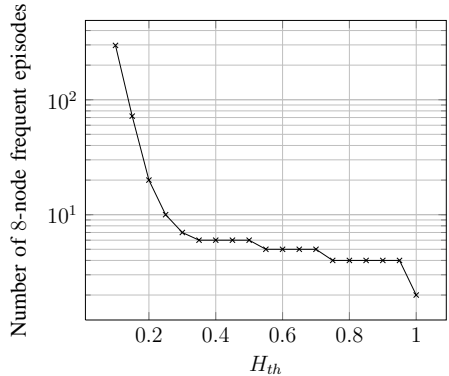

\begin{table}[t]
    \centering
    \caption{Run-time as noise level is increased by varying $\rho$. Patterns Embedded: (iii) \& (vi) from fig.\ref{fig:embedded-partial-orders}. $p = 0.055, 
    \eta = 0.7, M = 100, T = 10000 $, $f_{th} = 300, T_X=15, H_{th} = 0.35$. }
    \begin{tabular}{|c|c|c|} \hline
    $\rho$ &  Noise level($L_{ns}$)	&Run-time		 \\	\hline
    0.005    & 0.43  & 3 s \\ \hline
    0.02    & 0.75 & 6 s \\ \hline
    0.03    & 0.82  & 30 s \\ \hline
    0.045   & 0.87  & 1 m 45 s \\ \hline
    0.05    & 0.885 & 6 m 1 s \\ \hline
    \end{tabular}
\label{tab:noise}
\end{table}

\begin{table}[t]
	\centering
	\caption{Run-Time  as the data length is increased. $f_{th}/T= 0.03,\rho=0.045$, rest same as table~\ref{tab:noise}.}
	\begin{tabular}{|c|c|c|} \hline
	%		&$H_{th}=0.4$	&$H_{th} = 0.9$		&$H_{th}=0.4$	&$H_{th}=0.9$		\\ \hline
	 $T$ 	&Data Length($n$) &Run-time	\\		\hline	
	 5000	&22,500		& 52 s	\\ 	\hline
	 10000	&45,000		& 1 m 45 s \\  	\hline
	 15000	&67,500		& 2 m 36 s	 \\  		\hline
	 20000	&90,000		& 3 m 25 s	 \\  		\hline
	\end{tabular}
\label{tab:data-length}
\end{table}
\begin{table}[t]
	\centering
	\caption{Run-Time  as the number of embedded patterns is increased. $\rho = 0.045$, rest same as table~\ref{tab:noise}.}	
	\begin{tabular}{|c|c|c|} \hline
	No. of patterns		&Embedded Patterns (Fig.~\ref{fig:embedded-partial-orders}) &Run-Time\\	\hline	
	%1 	&  	\\    & 	\hline         
	2 	& (iii),(vi)    & 1 m 45 s	\\ 	\hline
	%3 	& 	      & 	\\ 	\hline
	%4 	& 	      & 	\\ 	\hline
	5 	& (i),(iii),(v),(vi),(viii)     & 4 m 18 s	\\ 	\hline
	%6 	& 	      & 	\\ 	\hline
	%7 	& 	      & 	\\ 	\hline
	8 	& (i)-(viii)  & 11 m 10 s	\\ 	\hline
	\end{tabular}
\label{tab:number-embedded-patterns}
\end{table}

Our mining algorithm is robust to choice of frequency and $H(\alpha)$ thresholds as illustrated in 
figures~\ref{fig:frequency-threshold-variation} \& \ref{fig:H-threshold-variation}. Once the threshold is high enough to eliminate 
noisy/spurious episodes, the number of episodes output is close to constant over a wide range of
threshold choices. 

The algorithm also scales well with number of embedded patterns, data length and noise level.
In Tables~\ref{tab:number-embedded-patterns}, \ref{tab:data-length} \& \ref{tab:noise}, the data is generated from a set of $100$ event types, with different $8$-node
episodes embedded from fig.~\ref{fig:embedded-partial-orders}. The run-times given are average values obtained over $10$ different runs. 

Let $N_{ns}$ and $N_{sig}$ be the expected number of noise 
and signal events respectively in the data stream using our 
simulation model. By noise events here, we refer to the events in the noise stream. Similarly, by signal events, we refer to all the events coming from the various 
episode
streams. In a data stream with $N_{emb}$ embedded episodes (each of size $l$), one can verify that $N_{ns}=(|\scrE -
\scrE_1|\rho + \scrE_1 \rho/5)T$ and $N_{sig}$=$ \frac{TlN_{emb}}{ (l-1)/\eta + 1/p }$. We define the noise level ($L_{ns}$) as
fraction  of the
expected number of noise events, i.e. $L_{ns}=\frac{N_{ns}}{N_{ns}+N_{sig}}$.
Table~\ref{tab:noise} describes increase in run-times  with noise level $L_{ns}$,
which is the ratio of the expected number of noise events to the expected total number of events as per our simulation model. 
 We see that for low $\rho$ (say 0.02) 
  the running time is very less. THis is because at the one node level
 only the signal events are frequent, as a result of which the number of candidates in successive levels
 are less. As $\rho$ increases the number of candidates at 2 and 3 node level increases. Thus running times go up.
 %For $\rho$ = 0.04 the frequency of noise and signal events are roughly the same (infact the noise is on the higher side).
 For $\rho$ = 0.045, the number of candidates in the 3 node level goes up to the order of 30,000, because many 2-node episodes are frequent. Consequently, the
 running times are very high for noise levels at about $0.88$.

Similarly, Table~\ref{tab:data-length}  describes the run-time  variations with  data
length. We observe that the run-times increase almost linearly with data length. As the data length is increased, the ratio of
$f_{th}/T$ is kept constant. 
Table~\ref{tab:number-embedded-patterns} shows the run-time  variations with the number of embedded partial orders. 
We see an increase in the run-times because of increased number of candidates as the number of patterns is increased.
\section{Conclusions}

\label{sec:conclusions}
In this paper we presented a method for discovering frequent episodes with general partial orders.
Episode discovery from event streams is a very useful data mining technique though all the
currently available methods can discover only serial or parallel episodes. Here, we presented a
finite automata based algorithm for counting non-overlapped occurrences of injective episodes
with general partial orders. (Along the way, we note some interesting properties of the finite state automaton used to track the occurences).
The method is efficient and can take care of expiry-time constraints.
 The candidate generation algorithm presented here is very flexible and can be used to focus the
discovery process on many interesting subclasses of partial orders. In particular, our method can be
easily specialized to mine only serial or only parallel episodes. Thus, the algorithm presented
here can be used as a single method to discover serial episodes or parallel episodes or episodes
with general partial orders. Another important contribution of this paper is a new measure of
interestingness for partial order episodes, namely, bidirectional evidence. We showed that there is
an inherent combinatorial explosion in the number of frequent episodes when one considers general partial orders.
%This is because, for episodes with general partial orders, an $l$-node episode can have many
%$l$-node subepisodes.
Our bidirectional evidence is very useful in discovering the most appropriate
partial orders from the data stream. The effectiveness of the data mining method is demonstrated
through extensive simulations.

In this paper we have considered injective episodes and a special subclass of non-injective partial order episodes (which includes all injective partial order episodes). 
We note that this subclass includes the set
of all non-injective serial episodes. In that sense, our algorithms truely generalize existing serial and parallel episode discovery algorithms.  Extending the ideas 
presented here to the class of all partial order episodes
is an important problem. Another potential direction is the development of a statistical significance
test for general partial order patterns in event streams. We will address these in our future work.

\bibliographystyle{abbrv}
\bibliography{srivats}

\section{Appendix}
\label{sec:app}

Finally, we describe the more efficient {\tt GetPotentialCandidates()} function (listed as {\em
Algorithm~\ref{algo-GetPotentialCandidates}}). The input to {\em Algorithm~\ref{algo-GetPotentialCandidates}} is a
pair of episodes, $\alpha_1$ and $\alpha_2$, both of size $l$, and both appearing in the same block of
the set, $\scrF_l$, of frequent $l$-node episodes. Recall that $\alpha_1$ and $\alpha_2$ are
identical in their first $(l-1)$ nodes (in respect of both the associated event-types as well as the partial order among these
event-types).  This common $(l-1)$-size partially ordered set is denoted as $\scrX$. The output of {\em Algorithm~\ref
{algo-GetPotentialCandidates}} is the set, $\scrP$,
of potential candidates that can be constructed from $\alpha_1$ and $\alpha_2$. 
The function {\tt
GetPotentialCandidates()} constructs a $\scrY_0$ combination of $\alpha_1$ and $\alpha_2$ as per (eqn.\ref{eq:Y0}) using the function {\tt
SimpleJoin()} 
%(listed as {\em Algorithm~\ref{algo-SimpleJoin}}) 
and  retains only those
combinations of $\alpha_1$ and $\alpha_2$ (of the three possible) which satisfy transitivity. 
%For an $l$-node
%episode, a brute force transitivity check would be $\scrO(l^3)$. Here, due to the structural restriction on $\alpha_1$ and $\alpha_2$,
%we just 
%have to do the 
%transitivity checks which necessarily involve the event types $x^{\alpha_1}_l$ and $x^{\alpha_2}_l$). Interestingly, 
%this can be done in $\scrO(l)$ time. 
%One could do
%these transitivity checks separately on the three combinations of $\alpha_1$ and $\alpha_2$. 
The {\tt GetPotentialCandidates()} function  decides the valid combinations based on some special checks on the kind of nodes in $\scrX$. 

For purposes of easier illustration, we classify the nodes in $\scrX$ based on its relation with $x_l^{\alpha_1}$ 
and $x_l^{\alpha_2}$.
We would have $9$ such type of nodes. A node $z\in \scrX$ is of the following types. \newline
$(1)$ -  $(x_l^{\alpha_1},z)$ and $(z,x_l^{\alpha_2})$ belong to $R$. \newline
$(1')$ - $(x_l^{\alpha_2},z)$ and $(z,x_l^{\alpha_1})$ belong to $R$. \newline
$(2)$  - $(x_l^{\alpha_1},z) \in R$, no edge between $z$ and $x_l^{\alpha_2}$. \newline
$(2')$ - $(z,x_l^{\alpha_1})\in R$, no edge between $z$ and $x_l^{\alpha_2}$. \newline
$(3)$ - $(x_l^{\alpha_2},z) \in R$, no edge between $z$ and $x_l^{\alpha_1}$. \newline
$(3')$  - $(z,x_l^{\alpha_2})\in R$, no edge between $z$ and $x_l^{\alpha_1}$. \newline
$(4)$ - $(z,x_l^{\alpha_2})$ and $(z,x_l^{\alpha_1})$ belong to $R$.\newline
$(4')$ - $(x_l^{\alpha_1},z)$ and $(x_l^{\alpha_1},z)$ belong to $R$.\newline
$(4'')$ - neither connected to $x_l^{\alpha_1}$ nor $x_l^{\alpha_2}$.\newline

We describe the {\tt
GetPotentialCandidates()} function with these nodes in mind. If a node of type
(1) exists (condition as per line $3$), then $\scrY_1$  is the only generated candidate (as per lines $7,9,10$). Similarly, if a node of type(1') exists (condition as per
line $4$), then $\scrY_2$ is the only
generated candidate (as per lines $8-10$). Suppose neither nodes of the type $(1)$ nor $(1')$ exist, then $\scrY_0$ is a sure candidate. Further, $\scrY_1$ 
is
generated iff nodes of type (2') and (3) do not exist in $\scrX$. Similarly, 
 $\scrY_2$ 
is
generated iff nodes of type (2) and (3') dont exist in $\scrX$. One can verify this from lines $14$ onwards in {\tt
GetPotentialCandidates()}. To show its correctness, we first make an important observation state as a lemma.

\begin{lemma}\label{node prop}
In $\scrY_0$, if a  node of type (1) exists, there cannot exist nodes of type (1'), (2') and (3). Similarly, if a node of type (1') exists, 
there cannot
exist nodes of type (1),(2) and (3'). This also holds for $\scrY_1$ and $\scrY_2$.
\end{lemma}

\begin{proof}
Given that a node $z_0$ of type (1) exists in $\scrX$, we will show that none of above 3 type of nodes can exist by contradiction.
Suppose a node $z_1$ of type (1') exists, then $(z_1,x_l^{\alpha_1}) \in R \implies (z_1,x_l^{\alpha_1}) \in R^{\alpha_1}$. Since $z_0$ is of 
type (1),
$(x_l^{\alpha_1},z_0) \in R \implies (x_l^{\alpha_1},z_0) \in R^{\alpha_1}$. By the transitivity of $R^{\alpha_1}$, it follows that $(z_1,z_0) 
\in R_1$. Also, since
$z_0$ is of type (1),
we have $(z_0,x_l^{\alpha_2}) \in R \implies (z_0,x_l^{\alpha_2}) \in R^{\alpha_2}$. Likewise, $z_1$ being of type (1') also says $(x_l^
{\alpha_2} , z_0) \in R
\implies     (x_l^{\alpha_2},z_1) \in R^{\alpha_2}$. Hence the transitivity of $R_2$ tells us that $(z_0,z_1) \in R_2$. So we now have the same
pair of nodes being connected in opposite ways in $R^{\alpha_1}$ and $R^{\alpha_2}$. This contradicts condition (2) of comvining $R^{\alpha_1}$ 
and $R^{\alpha_2}$ 
that they share the  same partial order on $x_1x_2\dots x_{l-1}$.

Suppose a node $z_1$ of type (2') exists, then $(z_1,x_l^{\alpha_1}) \in R \implies (z_1,x_l^{\alpha_1}) \in R^{\alpha_1}$. 
Also $(x_l^{\alpha_1},z_0)\in R^{\alpha_1}$.
Transitivity of $R^{\alpha_1}$ tells us $(z_1,z_0) \in R_1$. Since both $z_0$ and $z_1$ both belong to $x_1x_2 \dots x_{l-1}$, $(z_1,z_0)
\in R_2$. We also have $(z_0,x_l^{\alpha_2}) \in R \implies (z_0,x_l^{\alpha_2}) \in R^{\alpha_2}$. Transitivity in $R_2$ now 
implies $(z_1,x_l^{\alpha_2})\in R^{\alpha_2}$
and hence is in $R$. But this  edge must be absent as $z_1$ is of type(2'). A similar contradiction arises for a node of type
(3).

The second statement of the theorem has proofs analogous to that of the first statement.

\end{proof}

We will now show that this efficient procedure generates the correct relations.
\begin{theorem}
The generated realtions (among the three combinations $\scrY_0$, $\scrY_1$ and $\scrY_2$ possible) as per {\em
Algorithm\ref{algo-GetPotentialCandidates}},  are 
all transitively closed and the ones not generated violate
transitivity. 
\end{theorem}

\begin{proof}
Let us list out the six
possibilities that need to be checked, for proving transitivity  (because $\alpha_1$ and $\alpha_2$ share the same subepisode on dropping their last
nodes respectively, as already discussed in the candidate generation section). Let $z'$ denote an element belonging to $\scrX$. \newline
(a)$(z',x_l^{\alpha_1}),(x_l^{\alpha_1}, x_l^{\alpha_2})\in \scrY_i \implies (z',x_l^{\alpha_2})\in \scrY_i$\newline
(b)$(z',x_l^{\alpha_2}),(x_l^{\alpha_2}, x_l^{\alpha_1})\in \scrY_i \implies (z',x_l^{\alpha_1})\in \scrY_i$\newline
(c)$(x_l^{\alpha_1},z'),(z', x_l^{\alpha_2})\in \scrY_i \implies (x_l^{\alpha_1},x_l^{\alpha_2})\in \scrY_i$\newline
(d)$(x_l^{\alpha_2},z'),(z', x_l^{\alpha_1})\in \scrY_i \implies (x_l^{\alpha_2},x_l^{\alpha_1})\in \scrY_i$\newline
(e)$(x_l^{\alpha_1}, x_l^{\alpha_2}),(x_l^{\alpha_2},z')\in \scrY_i \implies (x_l^{\alpha_1},z')\in \scrY_i$\newline
(f)$(x_l^{\alpha_2}, x_l^{\alpha_1}),(x_l^{\alpha_1},z')\in \scrY_i \implies (x_l^{\alpha_2},z')\in \scrY_i$

We do these six transitivity checks on a case by case basis as adopted by the procedure.

Case(i) A node $z$ of type (1) exists in $\scrX$ : Here, we need to show that $\scrY_0$, $\scrY_2$ are not transitively closed and
$\scrY_1$ is indeed closed. 

Since $z$ is of type (1), $(x_l^{\alpha_1},z)$ and $(z,x_l^{\alpha_2})$ are present in both $\scrY_0$ and $\scrY_2$. But transitivity demands 
the
edge $(x_l^{\alpha_1},x_l^{\alpha_2})$ which is absent in both $\scrY_0$ and $\scrY_2$. Hence both of them are not closed.

To prove the transitive closedness of $Y_1$, let us perform the six checks listed above. If hypothesis of (a) is true, and suppose
$(z',x_l^{\alpha_2})\notin \scrY_1)$, then either there exists an edge $(x_l^{\alpha_2},z')\in \scrY_1$ or  there exists no edge between 
$z'$ and
$x_l^{\alpha_2}$. In the first case $z'$ must be of type (1') which cannot exist from lemma \ref{node prop}. In the second case $z'$
must be of type (2') which also cannot exist from lemma \ref{node prop}. This proves (a). Hypothesis of (b) and (f)
cannot be true in $\scrY_1$ because of the reverse edge $(x_l^{\alpha_1}, x_l^{\alpha_2})$. (c) is obviously true in $\scrY_1$. The hypothesis of (d) 
indicates the existence of a type
(1') node in $\scrY_1$ which is not possible from lemma \ref{node prop}. Correctness of (e) is similar to that of (a). If
hypothesis of (e) is true, and suppose
$(x_l^{\alpha_1},z')\notin \scrY_1)$, then either there exists an edge $(z',x_l^{\alpha_1})\in \scrY_1$ or  there exists no edge between $z'$ and
$x_l^{\alpha_1}$. In the first case $z'$ must be of type (1') which cannot exist from lemma \ref{node prop}. In the second case $z'$
must be of type (3) which also cannot exist from lemma \ref{node prop}. This proves (e).

Case(ii) A node of type (1') exists in $\scrX$: This is analogous to case (i).

Case(iii) Neither a node of type(1) nor type(1') exists : First we need to show that $\scrY_0$ is closed always here. Hypothesis
of (a), (b), (e) and (f) are never true as they involve a direct edge between $x_l^{\alpha_1}$ and $x_l^{\alpha_2}$ which is not present in
$\scrY_0$. Hypothesis of (c) and (d) demand the existence of nodes of type (1) and (1') respectively which dont arise this
scenario. This shows the transitivity of $\scrY_0$ in this case.

Further, we show that $\scrY_1$ is closed iff no nodes of type (2') and (3)  exist in $\scrX$. \newline
$(\Rightarrow)$ Let us prove the contrapositive of the forward statement. If a node $z'$ of type (2') exists, then we have 
$(z',x_l^{\alpha_1})$, $(x_l^{\alpha_1},x_l^{\alpha_2})\in \scrY_1$ but there is no edge between $z'$ and
$x_l^{\alpha_2}$. This violates transitivity of $\scrY_1$. similarly, if a node $z'$ of type (3) exists, then we have
$(x_l^{\alpha_1},x_l^{\alpha_2}),(x_l^{\alpha_2},z')\in \scrY_1$, but there is no edge between $z'$ and $x_l^{\alpha_1}$. This violates transitivity of $\scrY_1$.

$(\Leftarrow)$ Suppose no nodes of type (2') and (3) exist, we will show the closedness of $\scrY_1$. If hypothesis of (a) is true, 
and suppose
$(z',x_l^{\alpha_2})\notin \scrY_1)$, then either there exists an edge $(x_l^{\alpha_2},z')\in \scrY_1$ or  there exists no edge between $z'$ and
$x_l^{\alpha_2}$. In the first case $z'$ must be of type (1') which cannot exist here in case(iii). In the second case $z'$
must be of type (2') which also cannot exist from the hypothesis. This proves (a). The hypothesis of (b) and (e) are not
satisfied here as they involve the edge $(x_l^{\alpha_1},x_l^{\alpha_2})$. The hypothesis of (c) and (d) demands the existence of nodes of type(1)
and (1') which cannot exist here in case (iii).If
hypothesis of (e) is true, and suppose
$(x_l^{\alpha_1},z')\notin \scrY_1)$, then either there exists an edge $(z',x_l^{\alpha_1})\in \scrY_1$ or  there exists no edge between $z'$ and
$x_l^{\alpha_1}$. In the first case $z'$ must be of type (1') which cannot exist here in case(iii). In the second case $z'$
must be of type (3) which also cannot exist from the hypothesis. This proves (e).

Further, we show that $\scrY_2$ is closed iff no nodes of type (2) and (3')  exist in $\scrX$. The proof of this is analogous to
that of $\scrY_1$.
\end{proof}

\begin{algorithm}
\footnotesize
\caption{GetPotentialCandidates($\alpha_1$, $\alpha_2$)}
\label{algo-GetPotentialCandidates}
\linesnumbered
\SetKw{KwAnd}{and}
\SetKw{KwOr}{or}
\SetKwFunction{SimpleJoin}{SimpleJoin}

\KwIn{Patterns, $\alpha_1$ and $\alpha_2$, both of size $l$}
\KwOut{$\scrP$, candidate possibilities from $\alpha_1$ and $\alpha_2$}
\SetKwData{TRUE}{TRUE}
\SetKwData{FALSE}{FALSE}

Initialize $flg,flg1,flg2 \leftarrow 0$ and $\scrP \leftarrow \phi$\;
\For{$(i\leftarrow 1$; $i \leq l-1$ \KwAnd $flg =0$; $i \leftarrow i +1 )$}{
	\lIf{$(\alpha_1.e[l][i] = 1$ \KwAnd $\alpha_2.e[i][l] =1)$}{$flg \leftarrow 1$}\;
	\lIf{$(\alpha_2.e[l][i] = 1$ \KwAnd $\alpha_1.e[i][l] =1)$}{$flg \leftarrow 2$}\;
}
\If{$flg \neq 0$}{
	$\gamma_1	\leftarrow $ \SimpleJoin{$\alpha_1$, $\alpha_2$}\;
	\lIf{$flg = 1$}{$\gamma_1.e[l][l+l] \leftarrow 1$}\;
	\lElse{$\gamma_1.e[l+1][l] \leftarrow 1$}\;
	Add $\gamma_1$ to $\scrP$\;	
	\Return{$\scrP$}\;
}
\Else{
	$\gamma_1 \leftarrow$ \SimpleJoin{$\alpha_1$, $\alpha_2$}\;
	Add $\gamma_1$ to $\scrP$\;
	\For{$i\leftarrow 1$ \KwTo $l-1$}{
		\If{($\alpha_1.e[l][i] = 1$ \KwAnd $\alpha_2.e[l][i] = 0$) \KwOr \\ ($\alpha_1.e[i][l] = 0$ \KwAnd $\alpha_2.e[i][l] = 1$)}{
			%\lIf{$flg \neq 2$}{$flg \leftarrow 1$}
			%\lElse{\Return{$\scrP$}}\;
			$flg1=1$\;
		}
		\If{($\alpha_1.e[l][i] = 0$ \KwAnd $\alpha_2.e[l][i] = 1$) \KwOr \\ ($\alpha_1.e[i][l] = 1$ \KwAnd $\alpha_2.e[i][l] = 0$)}{
			%\lIf{$flg \neq 1$}{$flg \leftarrow 2$}
			%\lElse{\Return{$\scrP$}}\;
			$flg2=1$\;
		}
	}	
	\If{$flg1 = 0$ \KwAnd $flg2 = 0$}{
		$\gamma_2	\leftarrow$ \SimpleJoin{$\alpha_1$, $\alpha_2$}\;
		$\gamma_2.e[l][l+1] \leftarrow 1$\;
		Add $\gamma_2$ to $\scrP$\;
		
		$\gamma_3	\leftarrow$ \SimpleJoin{$\alpha_1$, $\alpha_2$}\;
		$\gamma_3.e[l+1][l] \leftarrow 1$\;
		Add $\gamma_3$ to $\scrP$\;	
	}
	\If{$flg1 = 1$ \KwAnd $flg2 = 0$}{
		$\gamma_2 \leftarrow$ \SimpleJoin{$\alpha_1$, $\alpha_2$}\;
		$\gamma_2.e[l][l+1] \leftarrow 1$\;
		Add $\gamma_2$ to $\scrP$\;	
	}
	\If{$flg1 = 0$ \KwAnd $flg2 = 1$}{
		$\gamma_3 \leftarrow$ \SimpleJoin{$\alpha_1$, $\alpha_2$}\;
		$\gamma_3.e[l+1][l] \leftarrow 1$\;
		Add $\gamma_3$ to $\scrP$\;
	}
	\Return{$\scrP$}\;
}

\end{algorithm}

\end{document}